\documentclass[final,12pt]{clear2024} 

\title[Towards the Reusability and Compositionality of Causal Representations]{Towards the Reusability and Compositionality of Causal Representations}
\usepackage{times}

\usepackage{booktabs}
\usepackage{caption}
\usepackage{subcaption}
\usepackage{tikz}
\usepackage{multirow}
\usepackage{xspace}
\usepackage{paralist}
\usepackage{floatrow}
\usepackage{xcolor}
\usepackage{soul}
\usepackage{multirow}
\usepackage{colortbl}
\usepackage{cleveref}
\usepackage{arydshln}
\newlength{\twosubht}
\newsavebox{\twosubbox}

\newcommand{\OurApproach}{DECAF}
\newcommand{\ie}{i.e.,\xspace}
\newcommand{\eg}{e.g.,\xspace}
\renewcommand{\cite}{\citep}

\usepackage{amsmath,amssymb,amsfonts,bm,bbm,stmaryrd}

\def\eqref#1{equation~\ref{#1}}

\def\1{\bm{1}}

\DeclareMathAlphabet{\mathsfit}{\encodingdefault}{\sfdefault}{m}{sl}
\SetMathAlphabet{\mathsfit}{bold}{\encodingdefault}{\sfdefault}{bx}{n}

\newcommand{\range}[2]{\llbracket#1..#2\rrbracket}

\newcommand{\pa}[1]{\text{pa}(#1)} %

\newfloatcommand{capbtabbox}{table}[][\FBwidth]
\DeclareCaptionLabelFormat{andfigure}{#1~#2  \&  \figurename~\thefigure}

\clearauthor{%
 \Name{Davide Talon} \Email{talon.davide@gmail.com}\\
 \addr PAVIS - Istituto Italiano di Tecnologia (IIT), University of  Genova
 \AND
 \Name{Phillip Lippe} \Email{p.lippe@uva.nl}\\
 \addr QUVA Lab - University of Amsterdam%
 \AND
 \Name{Stuart James} \Email{stuart.a.james@durham.ac.uk}\\
 \addr PAVIS - Istituto Italiano di Tecnologia (IIT), Durham University%
 \AND
 \Name{Alessio {Del Bue}} \Email{alessio.delbue@iit.it}\\
 \addr PAVIS - Istituto Italiano di Tecnologia (IIT)%
 \AND
 \Name{Sara Magliacane} \Email{s.magliacane@uva.nl}\\
 \addr AMLab - University of Amsterdam, MIT-IBM Watson AI Lab
}

\begin{document}

\maketitle

\begin{abstract}%
Causal Representation Learning (CRL) aims at identifying high-level causal factors and their relationships from high-dimensional observations, e.g., images. While most CRL works focus on learning causal representations in a single environment, in this work we instead propose a first step towards learning causal representations from temporal sequences of images that can be adapted in a new environment, or composed across multiple related environments. In particular, we introduce \OurApproach{}, a framework that detects which causal factors can be reused and which need to be adapted from previously learned causal representations. Our approach is based on the availability of intervention targets, that indicate which variables are perturbed at each time step.
Experiments on three benchmark datasets show that integrating our framework with four state-of-the-art CRL approaches leads to accurate representations in a new environment with only a few samples.
\end{abstract}


\section{Introduction}

Causal Representation Learning (CRL) ~\citep{scholkopf2021toward, lachapelle2022disentanglement, lippe2022citris, yao2021learning} aims at identifying high-level causal factors and their relationships from underlying low-level observations, \eg images. 
While learning structured and disentangled representations has proved effective for interpretability, efficiency and fairness of deep learning models \citep{higgins2017beta, van2019disentangled, locatello2019fairness}, most methods assume independent factors of variation. This assumption is often not met in real-world applications, which hinders the generalization capabilities of these methods \cite{dittadi2020transfer,trauble2021disentangled, dittadi2021generalization, roth2022disentanglement}.
CRL generalizes the disentanglement setting by considering potential causal relations between the latent causal variables. Recent works rely on auxiliary variables ~\cite{khemakhem2020variational, lippe2023biscuit}, non-stationarity~\cite{yao2021learning, yao2022temporally}, sparsity~\cite{lachapelle2022disentanglement, lachapelle2022synergies}, intervention targets ~\cite{lippe2022intervention, lippe2022citris, lippe2022icitris} and counterfactuals \cite{von2021self, brehmer2022weakly} to identify the causal factors.
Causal representations retain the \emph{modular} nature of the associated causal generative model:
an external change, \ie an intervention, on a specific target variable will not affect the \emph{causal mechanism}, \ie the conditional distribution of any other variable given its parents \cite{pearl2009causality}.

While most CRL works focus on learning causal representations in a single environment, in this work we instead propose a first step towards learning causal representations from temporal sequences of images that can be adapted in new environments, or composed across multiple related environments. We are motivated by leveraging the implicit modularity of causal representations, as well as many real-world applications in which we want an agent to leverage its previous knowledge and adapt to changes in the environment with the least interactions possible.

In particular, we consider the TempoRal Intervened Sequences (TRIS) setting \cite{lippe2022citris}. In this setting we observe temporal sequences of high-dimensional observations of an underlying causal system, and at each time step any of the causal variables might be intervened. We also assume that we have labels for which variables were intervened at each time step, represented as a binary \emph{intervention target} vector.
We leverage this information in \OurApproach{} (DEtect Changes and Adapt Factors), a framework that detects which causal factors can be reused and which need to be adapted from previously learned causal representations. DECAF can be combined with any CRL approach that works in TRIS.

\begin{figure}[t!]
\centering
  \includegraphics[width=0.9\linewidth]{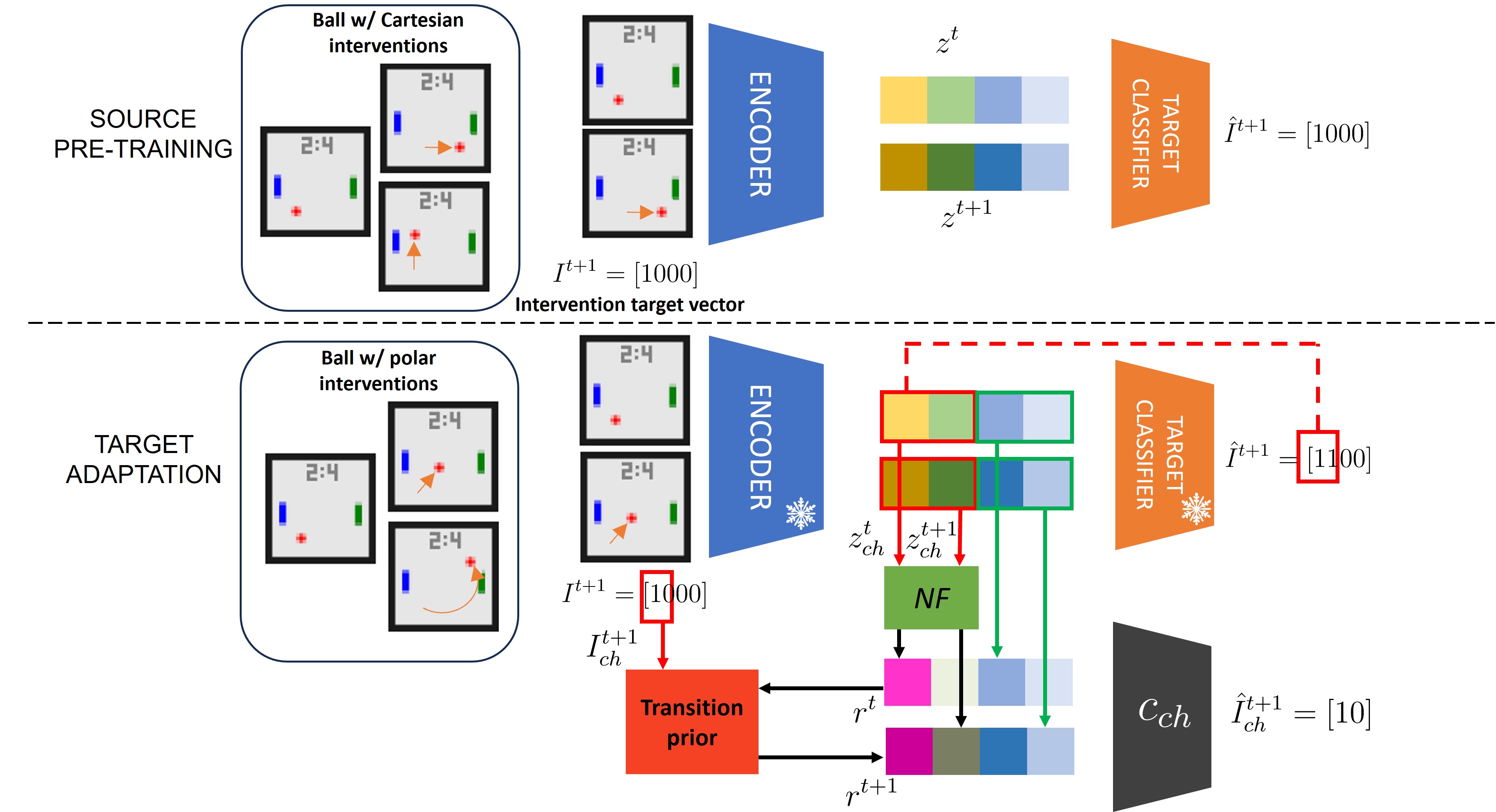}
\caption{Overview of our approach for the \emph{adaptation task} in Pong, where the \emph{source} environment on which we learn the initial causal representation models the position of the ball as Cartesian coordinates, while the \emph{target} environment uses polar coordinates for the ball position.
}
\label{fig:concept}
\end{figure}
To motivate our approach, we show an application of our framework for the \emph{adaptation task} in Pong in Figure~\ref{fig:concept}. In the \emph{source} environment, we exploit the available intervention targets $I^t$ at each timestep $t$ to learn the causal representation of the system, including the position of the two paddles and the ball. In this environment the position of the ball is measured in Cartesian coordinates $x$ and $y$. Instead, in the \emph{target} environment, the dynamics of the ball are modelled in polar coordinates, radius $r$ and angle $\theta$. Hence also the available interventions in this environment are changing either the radius or the angle of a ball.  In this setting, \OurApproach{} first learns a causal representation learned in the source domain using a standard CRL approach with an \emph{encoder}. It also trains a \emph{target classifier} to predict the intervened variables $I^{t+1}$ at time $t+1$ from the predicted latent states $z^t$ at time $t$ and $z^{t+1}$ at time $t+1$. When applying the target classifier to the new environment, \OurApproach{} exploits the discrepancies in the predicted and the intervened targets to detect which of the causal factors need to be adapted. Only these factors are then adapted by training a normalizing flow (NF) with a \emph{transition prior} and an \emph{auxiliary target classifier} that enforces that each newly learned latent variable models at most one intervention target. The other causal factors can be directly used in the new environment. As we show in the experiments, we can use a similar approach also in \emph{compositional} settings in which we can combine representations from multiple source environments. 

The contribution of this work is three-fold: (i) we formalize a generative model for the changes across environments for which we can \emph{adapt} or \emph{compose} causal representations, (ii) we propose \OurApproach{} (DEtect Changes and Adapt Factors), a novel framework that detects changes, adapts and composes causal representations, (iii) we validate the benefits of repurposing learned causal representations on 
three existing CRL benchmarks, for which we develop several adaptation and composition tasks.

\section{Background}
\label{sec:bckg}
We assume our data follow the TempoRal Intervened Sequence (TRIS) setting \citep{lippe2022citris}. In this setting we assume that there is an underlying unobserved causal system, and at each time step there can be an intervention on a set of causal variables. We only observe a time series of high-dimensional observations of it and the labels describing which variables have been intervened on, the \emph{intervention targets}. Here we summarize the assumptions, and refer to \citet{lippe2022citris} for details.

\paragraph{Latent causal process.} We assume the latent causal process can be described by a Dynamic Bayesian Network (DBN) \cite{DBN, murphy2002dynamic} over a set of $K$ multidimensional causal variables $(C_1, \dots, C_K)$ that generates the data at hand. At each time step, we only  allow that a variable $C_i^t$ can be potentially a parent of a variable $C_j^{t+1}$ for $i,j \in \range{1}{K}$, i.e. the DBN is first-order Markov and has no instantaneous effects, and the causal relations are stationary, \ie the causal parents repeat across all timesteps.
In other words, each causal variable follows the structural causal equation $C_i^t = f_i(\pa{C_i^t}, \epsilon_i)$ for $i=\range{1}{K}$, where $\pa$ are the parents, which are a subset of the variables in the previous time step, and $\epsilon_i$ is its exogenous noise.
We assume the noises $\epsilon_i$ for $i =\range{1}{K}$ to be mutually independent. Causal factors can be multivariate, \ie $C_i \in \mathcal{D}_i^{M_i}$ with $M_i \geq 1$ where $\mathcal{D}_i$ is $\mathbb{R}$ for continuous variables and $\mathbb{Z}$ for discrete ones. Hence, the causal factor space is defined as $\mathcal{C} = \mathcal{D}_1^{M_1}\times\mathcal{D}_2^{M_2}\times...\times\mathcal{D}_K^{M_K}$.
We denote as $C^t=(C^t_1, \dots, C^t_K)$ the causal factors at time step $t$.

\paragraph{Interventions.} We assume that the causal system can be subjected to an intervention at each time step and that if it happens, we know the intervention targets. In particular, a binary vector $I^t \in \{0,1\}^{K}$ indicates that a variable $C^t_i$ is intervened upon iff $I^t_i=1$. Intervention values are unobserved.
Interventions can be \emph{soft}~\cite{eberhardt2007causation}, e.g. inducing a change in the mechanism of the intervened variables without necessarily making the target, or \emph{hard}, e.g. do-interventions $\text{do}(C_i=c_i)$~\cite{pearl2009causality}. Multiple variables can be intervened simultaneously. We model potential dependencies between intervention targets with an unobserved regime variable $R^t$ \cite{mooij2020joint}. We assume faithfulness of the distribution, hence there are no further independences than those given by the causal graph.

\paragraph{Observation function.} At each time step $t$, we observe a high-dimensional observation of the latent causal factors.
Let $f:\mathcal{C} \times \mathcal{U} \rightarrow \mathcal{X}$ be the invertible observation function from the space of factors $\mathcal{C}$ and noises $\mathcal{U}$ to the observation space $\mathcal{X}$. We define the high-dimensional observation $X^t = f(C^t_1,C^t_2,...,C^t_K,U^t)$, where $U^t \in \mathcal{U}$.

\paragraph{Adaptation of CRL approaches to TRIS.} Since the TRIS setting was originally developed for CITRIS~\cite{lippe2022citris}, we can use it as is in this setting.
We also adapt three other state-of-the-art CRL methods to work in the TRIS setting.
iVAE~\cite{khemakhem2020variational} assumes that the causal variables are conditionally independent given some auxiliary information. In TRIS, this information can be provided by $\{C^t, I^{t+1} \}$. LEAP~\cite{yao2021learning} leverages nonstationarity that is captured by a categorical auxiliary variable $u$, which can be represented with the intervention target vector $I^{t+1}$. 
Given actions with unknown targets, DMSVAE~\cite{lachapelle2022disentanglement} identifies the causal factors when the underlying causal graph has a sparse structure. In TRIS, we consider the information target vector as the action itself. 

Since we have multidimensional causal variables, we also need to learn a mapping from a latent space $\mathcal{Z} \subseteq \mathbb{R}^M$ with $M \geq K + 1$ to the causal space $\mathcal{C}$. We call this mapping the  assignment function $\psi: \range{1}{M}\to\range{0}{K}$. We denote the latent variables assigned to a causal variable $C^t_i$ as $z^{t}_{\psi_i}$ for $i \in \range{1}{K}$, while we denote with $z^{t}_{\psi_0}$ the latent variables that are not assigned to any causal variable. CITRIS  learns $\psi$ as part of its training, but iVAE, LEAP and DMSVAE do not and their identifiability is up to permutation and element-wise transformation.
To compare them, we then use supervision to match the latent space learned by iVAE, LEAP and DMSVAE with the ground truth causal variables.

\paragraph{Remark.}
While knowledge about the intervention targets might not be always possible, we stress that there are enough real-world settings in which we might have this information available. As real-world examples, consider an experiment in which we want to learn the causal relations between different genes from imaging data, and our experiments consist of gene knockouts of specific genes. In this setting, we typically have access to intervention targets. Other applications include experiment or intervention design~\citep{eberhardt2007causation, hyttinen2013experiment, shanmugam2015learning} in which we decide which variables we might want to intervene on to identify the graph or CRL~\citep{lippe2023biscuit}, especially in RL environments.

\section{A simple generative model of environments for adaptation and composition}

In this section we propose a simple generative model for changes across environments, for which our framework will be able to \emph{adapt} and \emph{compose} causal representations. 

\subsection{Adaption of causal representations.}
For simplicity, we assume that we have two environments, the source $S$ and the target $T$. We assume there is an underlying latent causal process with \emph{underlying causal variables} $C^t$ that is the same for both environments. In the source, we consider a set of \emph{source causal variables} $C_S^t$, which are an invertible function of the underlying causal variables $C^t$. Similarly, we consider a set of \emph{target causal variables} $C_T^t$, which are an invertible function of $C^t$. In general, we will assume that some of underlying causal variables $C^t_{sh}$ are \emph{shared} across the environments and with the underlying causal model, while others $C^t_{ch}$ can change across the environments and w.r.t. the underlying causal model.

More formally, we will assume that the underlying causal variables $C^t$ with size $K$ can be partitioned in $C^t_{ch}$ with size $K_{ch}$ and $C^t_{sh}$ with size $K_{sh}$. The source causal variables $C^t_S$ can be then defined as  $C^t_S= (h_S(C^t_{ch}), C^t_{sh})$, where $h_S$ is an invertible function. Similarly, the target causal variables are defined as $C^t_T = (h_T(C^t_{ch}), C^t_{sh})$ for an invertible $h_T$.
We denote with $K_S$ the number of source causal variables and with $K_T$ the number of target causal variables. The number of causal variables may change between source and target, as well as with respect to the underlying causal variables. Hence, we allow for refinement or coarsening of variables. However, the invertible functions $h_S, h_T$ imply that the joint dimensionality of the causal variables is always constant.

\subsection{Composition of causal representations.}
We can extend the same notation to the case of composition, in which there are multiple source environments and a single target environment. We again assume that there is an underlying causal model with variables $C^t$. Let $C^t_{S_i}$ be the source causal variables of one of the $L$ sources, and define $C^t_{sh_i}$ as the shared causal variables between the $S_i$-th source and the target environment.
We assume that the target causal variables $C^t_T$ are a composition of source causal variables that have been independently learned on the source environments. 

More formally:
\begin{align}
C^t_T = (h_{T}(C^t_{ch_T}), C^t_{sh}, C^t_{sh_1}, \dots, C^t_{sh_L}),
\end{align}
where $C^t_{sh}$ are the target causal variables shared with the underlying causal graph and $C^t_{ch_T}$ are the causal variables that are changed in the target environment with respect to the underlying causal variables through the invertible function $h_T$.
If the shared causal variables $C_{sh_i}$ are not disjoint, then the intersections will still be identical, and we can remove the duplicates. 
\section{Detection, Adaptation and Composition of Factors}
\label{sec:detection}

\begin{figure}[t!]

\centering
  \includegraphics[width=0.8\linewidth]{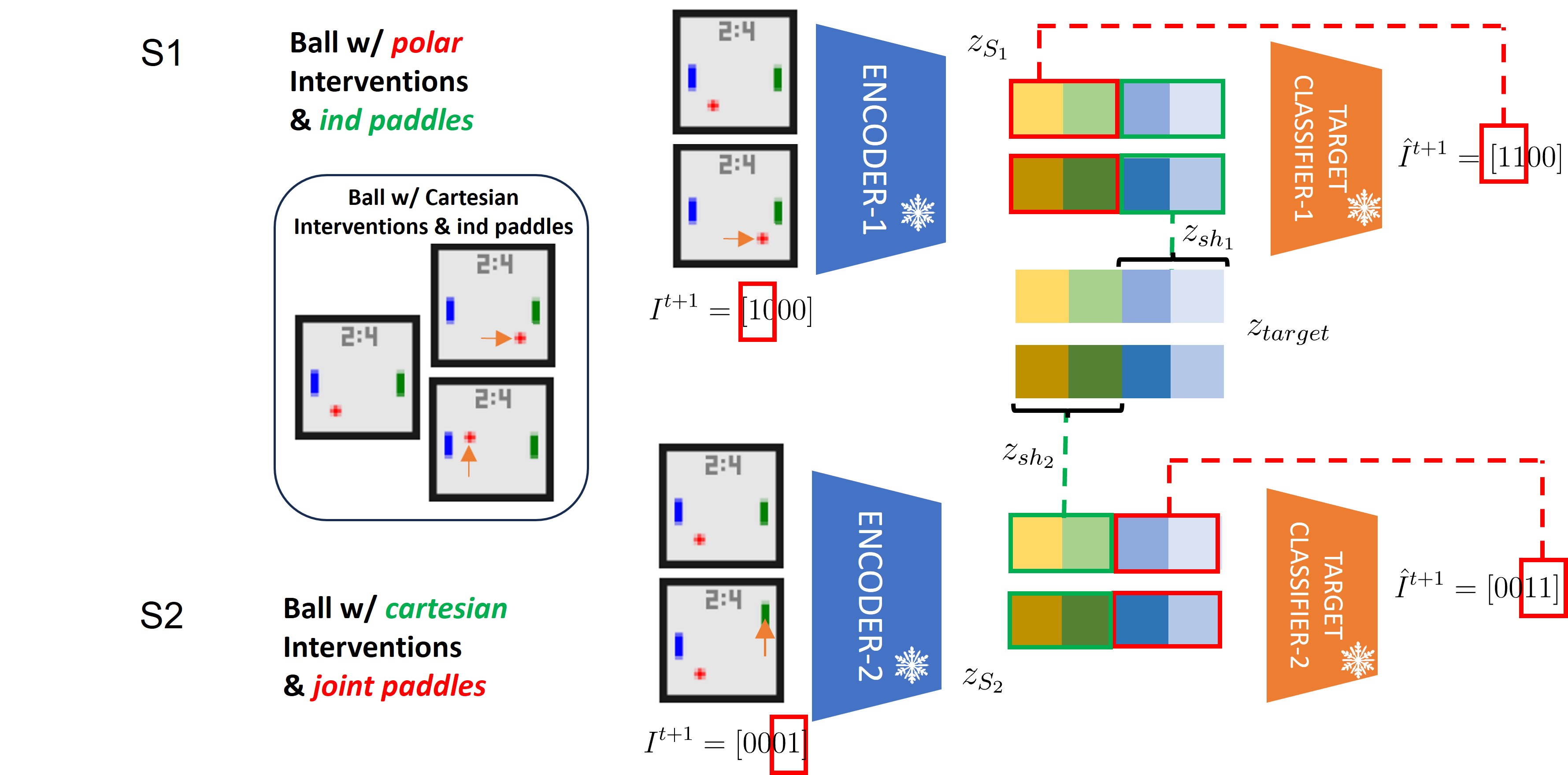}
\caption{Overview of our approach for the \emph{composition task} in Pong, where the first source environment models the data with polar ball position and independent paddles and the second environment employs Cartesian ball coordinates but entangled paddles. The \emph{target} environment uses Cartesian coordinates for the ball position and has independent paddles.
}
\label{fig:concept-comp}
\end{figure}

Here we describe our framework \OurApproach{} and show how it adapts or composes causal representations in environments that follow our generative model. We first introduce how we detect the changed causal variables, based on the discrepancies in predicting the intervention targets. We then describe how we adapt the changed factors with a normalizing flow and how we compose causal representations.

\paragraph{Changing variable detection.} 
Using a CRL approach adapted to the TRIS setting, as described in Section~\ref{sec:bckg},  we can learn a causal representation on the source data.
We also learn a \emph{target classifier} \cite{lippe2022citris} that predicts the next step intervention targets $I_i^{t+1}$ from the current latent state $z^t$ and the next step latent state assigned to the causal variable $C_i$, which we denote as $z^{t+1}_{\psi_i}$.
Intuitively, when we run the target classifier in the target environment, we expect that its accuracy would drop for the causal variables that have changed from the source to the target. 
In particular, for $k \in \range{1}{K}$, we define  $X_{S,I_k=1} := \{X_{S}^{t} \mid I^{t}_k = 1, t\in\range{1}{T}\}$ as the set of observations on the source environment $S$ in which $C_k$ has been intervened upon. Similarly let $X_{T,I_k=1}:= \{X_{T}^{t} \mid I^{t}_k = 1, t\in\range{1}{T}\}$ be the set of observations on the target environment $T$ in which $C_k$ has been intervened upon.

We define $\texttt{FPR}_{S,i}^k(j)$ and $\texttt{FNR}_{S,i}^k(j)$ as the False Positive Rate and False Negative Rate for intervention predictions of the classifier on the \emph{source environment} on samples $X_{S,I_k=1}$, when predicting the intervention target $I_j$ from the current time step $z^t$ and the subset of latents assigned to the variable $z_{\psi_i}$ at time steps $t+1$.
Similarly, we define the False Positive Rate and the False Negative Rate for intervention predictions on the \emph{target environment} as $\texttt{FPR}_{T,i}^k(j)$ and $\texttt{FNR}_{T,i}^k(j)$.
We detect the changing causal factors $\hat{C}_{ch}$ by considering differences in false positive rates or false negative rates greater than threshold $\tau$:
\begin{equation}
\hat{C}_{ch}=\{j \mid \exists \; i, k \in \range{1}{K} \; s.t. \; 
|\texttt{FPR}_{T,i}^k (j) - \texttt{FPR}_{S,i}^k (j)|
> \tau \lor |\texttt{FNR}_{T,i}^k (j) - \texttt{FNR}_{S,i}^k (j)| > \tau \}.
\end{equation}

As the target classifier generally predicts an intervention when the dynamics differ from the learnt observational ones, it tends to over-predict interventions in unseen environments. Thus, we found that using \texttt{FPR} to consistently outperforms using \texttt{FNR} and apply it throughout our experiments.

\paragraph{Adaptation.}
Once we have identified the changing causal variables $C_{ch}$, we adapt their representation $z_{ch} \in \mathbb{R}^{M_{ch}}$ by a Normalizing Flow (NF)~\cite{rezende2015variational}. The Normalizing Flow maps $z_{ch}$ to a new representation $r \in \mathbb{R}^{M_{ch}}$ with the same dimensionality, while guaranteeing invertibility between the representations.
Similarly to CITRIS \cite{lippe2022citris}, we train this flow with a transition prior $p_\phi$ parameterized by $\phi$ and condition each latent on exactly one intervention target $I_{ch}$ of the changing variables:
\begin{equation}
p_\phi(r^{t+1} \mid r^t, I^{t+1}_{ch}) = \prod_{C_{ch_i} \in C_{ch}}p_\phi(r^{t+1}_{{\psi_{ch}}_i}\mid r^t, I^{t+1}_{{ch}_i}),
\end{equation}
where $\psi_{ch_i}$ is the learnt assignment of the components of $z_{ch}$ to the causal variable $C_{ch_i}$, $\psi_{ch}: \range{1}{M_{ch}}\to\range{1}{K_{ch}}$.
The model directly learns an invertible map from source to target representation by maximizing the log-likelihood of the target samples:
\begin{equation}
\mathcal{L}^{\phi,\omega}_{\text{MLE}} = \log p_{z_{ch}}(z_{ch}) = \log p_\phi(\texttt{NF}_{\omega}(z_{ch})) + \log \left|\det \frac{d\, \texttt{NF}_{\omega}({z_{ch})}}{d\, z_{ch}}\right|,
\end{equation}
where $\texttt{NF}_{\omega}$ represents the normalizing flow with parameters $\omega$, and the original representation $z_{ch}$ is kept frozen.
During inference, we construct the final representation by replacing the changed causal variables $z_{ch}$ with the adapted representation $r=\texttt{NF}_{\omega}(z_{ch})$.

\paragraph{Composing causal factors.}
\label{sec:method-composition}
Besides adapting causal representations, we can also try to compose the representations that we have identified across a set of source environments, to form the causal representation of a new target environment, see Figure~\ref{fig:concept-comp} for an illustration. 
More formally, consider the representation of $L$ source environments $z_{S_l}, l=\range{1}{L}$.
First, we detect the causal variables $C_{sh_l}$ that are shared between each source representation $C_{S_l}$ and the target using the changing variable detection described previously.
In a second phase, we then concatenate the latent representation of all identified shared variables, i.e. $z_{\text{target}}=\{z_{sh_l} | l \in \range{1}{L}\}$.
With that, we construct a representation that identifies the causal variables in the target environment if all causal variables can be found in the provided source environments.

\section{Experiments}
We evaluate \OurApproach{} on three benchmark datasets and compare it to baselines for adaptation of causal representations. We apply \OurApproach{} to four different CRL approaches that has been adapted for the TRIS setting as in Section~\ref{sec:bckg}: CITRISVAE~\cite{lippe2022citris}, LEAP~\cite{yao2021learning}, DMSVAE~\cite{lachapelle2022disentanglement} and iVAE~\cite{khemakhem2020variational}. 
We denote the combination with \OurApproach{} with a suffix \texttt{*-DECAF}. Source models are trained on large data (250K samples), further details are presented in Appendix~\ref{app:experimental-details}.

\subsection{Experimental Setup}
\smallskip\textbf{Voronoi benchmark.} We consider the non-instantenous version of the Voronoi benchmark \cite{lippe2022icitris} rendering colored Voronoi tiles whose colors are a mixed version of the ground truth generating factors. The underlying causal representation model is synthetically generated: starting from a random DAG, each variable is evaluated as sample from a Gaussian centered on the output of the mechanism randomly initialized neural network. Finally, the variables are mixed by a random normalizing flow and depicted as colors of a fixed-structure Voronoi diagram. We experiment with the 6 variables version of the dataset where all variables undergo perfect interventions. To allow for the change, we generate a version of the dataset where the 3 changed variables are fed to a randomly initialized NF. This simulates a coordinate system change for these three variables, with interventions being applied in the new system. We denote with \texttt{REG} and \texttt{CH} the regular and changed versions of the dataset, respectively. In another version of the dataset, we enable for joint interventions on a group of 2 variables, while making sure there is no overlap between changed and coarse variables. We refer to the coarse version of the dataset as $\texttt{j}$ and with \texttt{i} its independently intervened counterpart.

\smallskip
\noindent\textbf{InterventionalPong.} We generate sequential data starting from InterventionalPong~\citep{lippe2022citris}, based on the known Atari game Pong \citep{bellemare2013arcade}. Six high-level causal variables underlie the generated data: \texttt{ball-pos-x}, \texttt{ball-pos-y}, \texttt{paddle-left-y}, \texttt{paddle}-\texttt{right-y}, \texttt{score-left}, \texttt{score-right}. The game dynamics follow two paddles playing one versus the other with the aim to score, i.e., let the ball go over the opponent's line of movement. Interventions are available for all causal variables, the scores are considered as a coarse variable. We generate multiple versions of the dataset, depending on different parameterizations of the interventions. Specifically, we consider a setting where the ball position is modelled in Cartesian  (\texttt{CA}) coordinates and a polar (\texttt{PO}) version where the ball moves in a polar coordinate system whose origin is the centre of the playground. Further, we consider coarse cases where a group of causal variables is always jointly intervened on and, hence, cannot be disentangled. In particular, we focus on the granularity of interventions associated with the paddles that could be independently (\texttt{PA}) or jointly (\texttt{jPA}) intervened.

\smallskip
\noindent\textbf{Temporal Causal3DIdent.} We consider the common benchmark of Temporal Causal3DIdent from \cite{von2021self} based on the temporal version in~\cite{lippe2022citris}. Samples visualize a rubber 3D object in the centre of a rendering scene. The dynamics are based on trigonometric functions. Observations follow 10 causal factors: \texttt{pos-x}, \texttt{pos-y}, \texttt{pos-z}, \texttt{rot-$\alpha$}, \texttt{rot-$\beta$}, \texttt{rot-spotlight}, \texttt{hue-obj}, \texttt{hue-spotlight}, \texttt{hue-background}, \texttt{obj-shape}. All causal factors are subject to interventions. We adapt the dataset to support a different parameterization of the object position and different intervention granularities. Precisely, we generate a version of the dataset with rotated z-axis of 30 degrees. As a consequence, the xy coordinate system is rotated by 30 degrees anticlockwise. We indicate the rotated version as \texttt{ROT} while the non-rotated version as \texttt{CA}. We take into account different levels of coarsening for the hue variables and denote as \texttt{jHUE} (\texttt{HUE}) the version of the dataset where hue variables are jointly (independently) intervened.

\begin{figure}[t!]
\centering

\sbox\twosubbox{%
  \resizebox{\dimexpr.97\textwidth-1em}{!}{%
    \includegraphics[height=5cm]{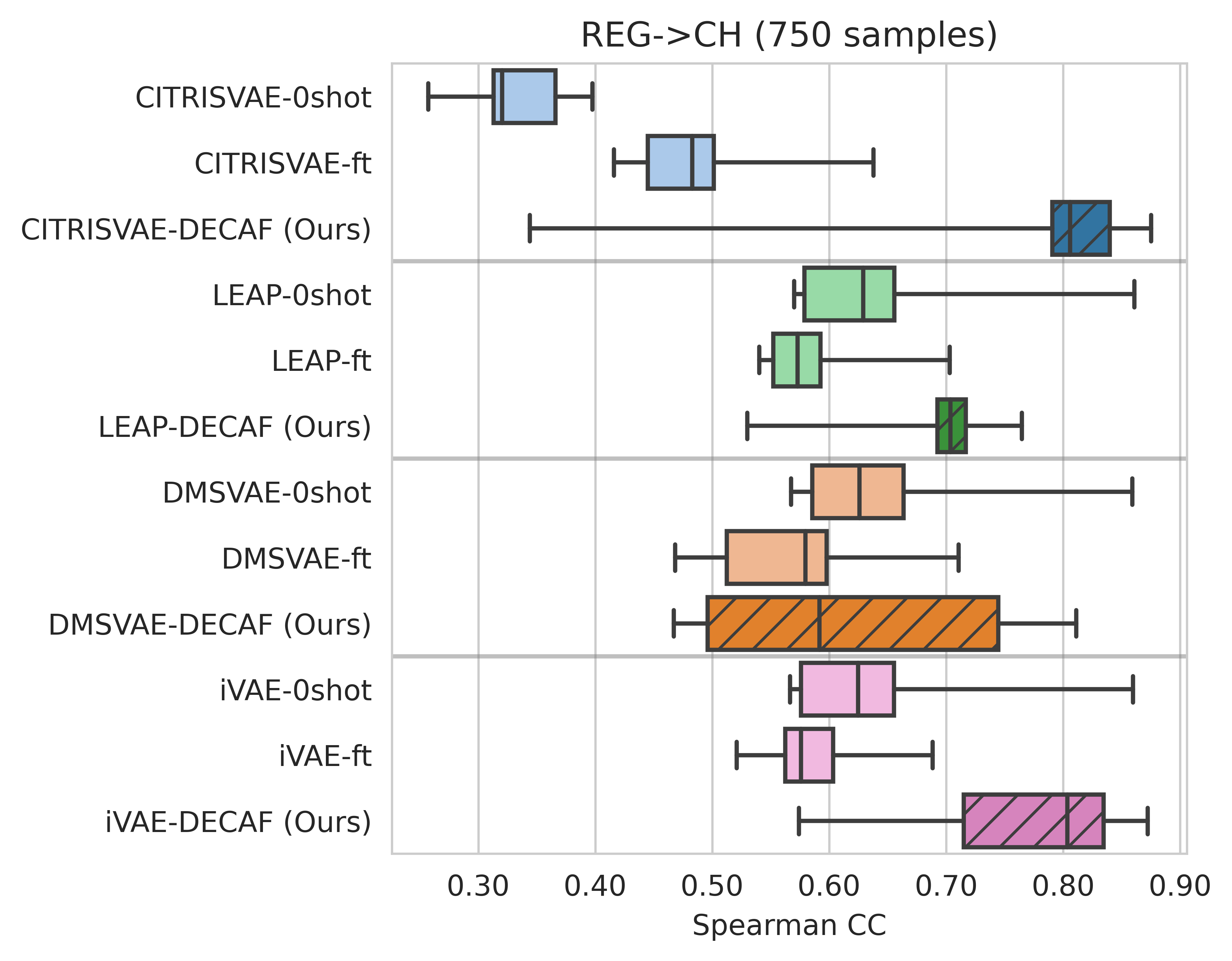}%
    \includegraphics[height=5cm]{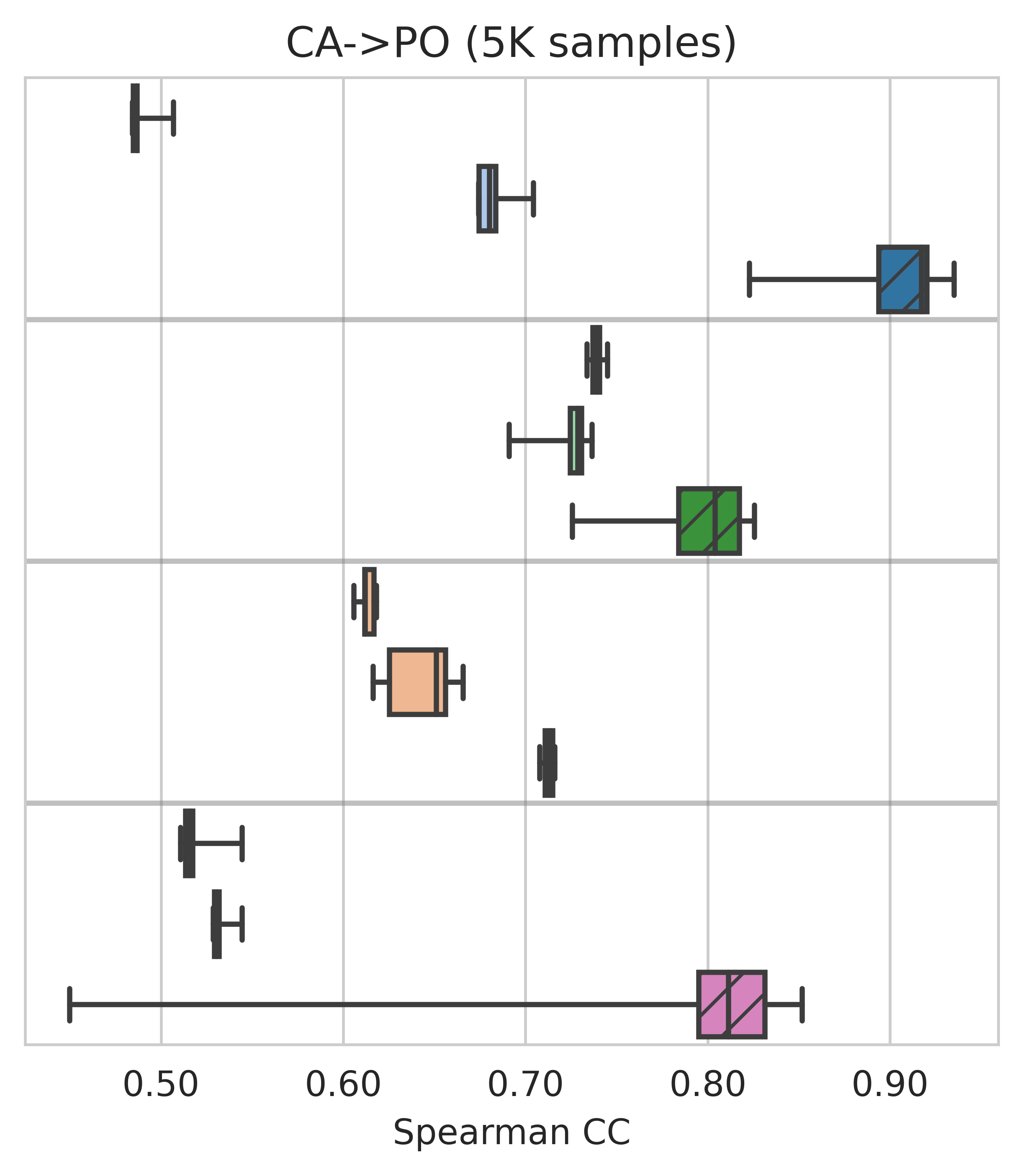}%
    \includegraphics[height=5cm]{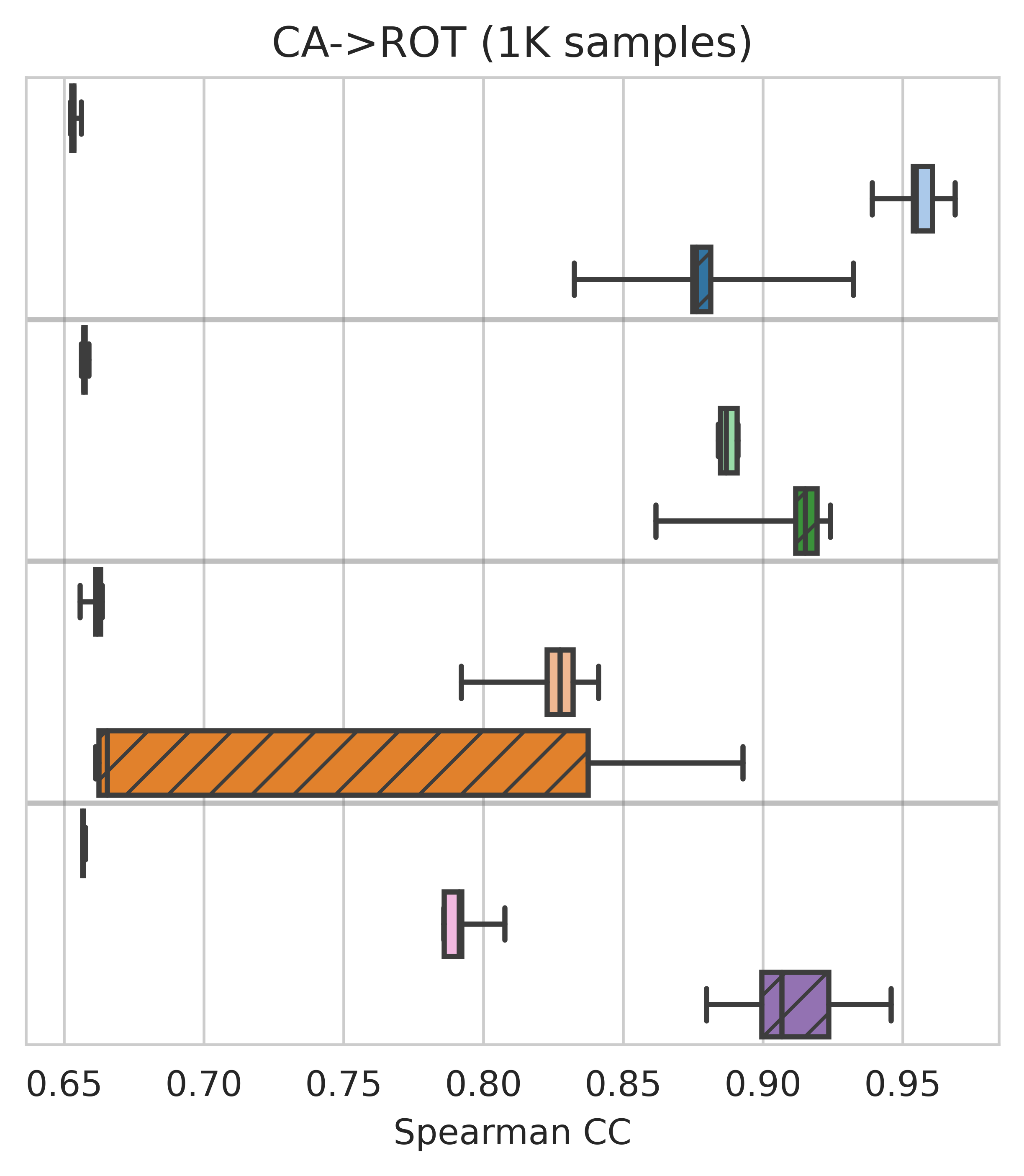}%
  }%
}
\setlength{\twosubht}{\ht\twosubbox}


\centering

\subcaptionbox{Voronoi Benchmark\label{fig:voronoi-re-reg2sh}}{%
  \includegraphics[height=\twosubht]{clear/results/results-voronoi-re-reg2sh.png}%
}
\subcaptionbox{InterventionalPong\label{fig:pong-re-ca2po}}{%
  \includegraphics[height=\twosubht]{clear/results/results-pong-re-ca2po.png}%
}
\subcaptionbox{Temporal Causal3DIdent\label{fig:c3d-re-ca2rot}}{%
  \includegraphics[height=\twosubht]{clear/results/results-c3d-re-ca2rot.png}%
}
\caption{Spearman CC (higher best, $\uparrow$) of inferred latents to the ground truth changed variables when adapting the representations. CRL approaches are color-coded, the proposed method has a darker color.}
\end{figure}
\smallskip
\noindent\textbf{Baselines.} For adapting causal representations from a source to a target environment, we compare \OurApproach{} to two simple adaptation baselines for reference, for each of the CRL methods: (1) \texttt{0shot}, where the model trained on the source environment is frozen and directly evaluated on the target data, and (2) \texttt{ft}, where the source model is fine-tuned on the target data using the same causal representation learning approach as in the source.

\smallskip
\noindent\textbf{Evaluation metrics.} We evaluate the approaches based on the correlation between inferred latents and the ground truth causal factors, as estimated using the $R^2$ coefficient of determination~\cite{wright1921correlation} and Spearman's rank correlation~\cite{spearman1904proof}. 
For methods that only identify the causal variables up to permutations, we follow previous works \cite{lippe2022citris, lachapelle2022disentanglement, lippe2023biscuit} by assigning latents to the ground truth causal variable with the highest correlation.
This results in a correlation matrix where the diagonal shows the correlation between matched learned and ground truth causal variables (higher is better, best 1.0), and off-diagonal elements the correlation to other variables (lower is better, best 0.0).
We propose a summary metric similar to the F1 score that combines the \emph{average diagonal correlation} \texttt{diag} and the average max off-diagonal 
correlation \texttt{off\_diag} through a harmonic mean. \texttt{diag} is intuitively similar to recall, while $(1-\texttt{off\_diag})$ is similar to precision. We define then the \emph{Combined Correlation} (CC) as:

\begin{equation}
CC = 2\ \frac{\texttt{diag}\cdot(1-\texttt{off\_diag})}{\texttt{diag} + (1 - \texttt{off\_diag})}.
\end{equation}
A model that perfectly identifies all causal variables achieves a score of $CC=1$, while it decreases for models that have low correlation between its identified latents and the ground truth causal variables (low \texttt{diag}), or large cross correlation across variables (high \texttt{off\_diag}). Full results are reported in Appendix~\ref{app:full-results}.

\subsection{Adaptation of causal representations}

\begin{table}
\begin{minipage}{0.55\linewidth}
		\centering
		\resizebox{\textwidth}{!}{%
\begin{tabular}{llcccc}
        \toprule
        Approach & Adaptation & $R^2$ diag $\uparrow$ & $R^2$ off-diag $\downarrow$ & Spearman diag $\uparrow$ & Spearman off-diag $\downarrow$\\ 
        \midrule

CITRISVAE & 0shot & 0.60 \scriptsize{$\pm$ 0.01} & 0.60 \scriptsize{$\pm$ 0.01} & 0.53 \scriptsize{$\pm$ 0.01} & 0.55 \scriptsize{$\pm$ 0.01}\\ 
& ft & 0.77 \scriptsize{$\pm$ 0.01} & 0.34 \scriptsize{$\pm$ 0.02} & 0.69 \scriptsize{$\pm$ 0.01} & 0.33 \scriptsize{$\pm$ 0.02}\\ 
& DECAF (Ours) & \textbf{0.93} \scriptsize{$\pm$ 0.03} & \textbf{0.09} \scriptsize{$\pm$ 0.04} & \textbf{0.94} \scriptsize{$\pm$ 0.03} & \textbf{0.14} \scriptsize{$\pm$ 0.06}\\ 
\midrule 
LEAP & 0shot & \textbf{0.85} \scriptsize{$\pm$ 0.01} & 0.24 \scriptsize{$\pm$ 0.01} & \textbf{0.87} \scriptsize{$\pm$ 0.01} & 0.36 \scriptsize{$\pm$ 0.01}\\ 
& ft & 0.64 \scriptsize{$\pm$ 0.02} & \textbf{0.16} \scriptsize{$\pm$ 0.02} & 0.72 \scriptsize{$\pm$ 0.02} & 0.28 \scriptsize{$\pm$ 0.02}\\ 
& DECAF (Ours) & 0.84 \scriptsize{$\pm$ 0.04} & 0.18 \scriptsize{$\pm$ 0.07} & 0.86 \scriptsize{$\pm$ 0.06} & \textbf{0.26} \scriptsize{$\pm$ 0.03}\\ 
\midrule 
DMSVAE & 0shot & 0.50 \scriptsize{$\pm$ 0.01} & 0.25 \scriptsize{$\pm$ 0.01} & 0.57 \scriptsize{$\pm$ 0.00} & 0.33 \scriptsize{$\pm$ 0.01}\\ 
& ft & 0.53 \scriptsize{$\pm$ 0.04} & 0.18 \scriptsize{$\pm$ 0.03} & 0.59 \scriptsize{$\pm$ 0.04} & 0.30 \scriptsize{$\pm$ 0.01}\\ 
& DECAF (Ours) & \textbf{0.61} \scriptsize{$\pm$ 0.01} & \textbf{0.14} \scriptsize{$\pm$ 0.01} & \textbf{0.65} \scriptsize{$\pm$ 0.01} & \textbf{0.21} \scriptsize{$\pm$ 0.01}\\ 
\midrule 
iVAE & 0shot & 0.59 \scriptsize{$\pm$ 0.04} & 0.53 \scriptsize{$\pm$ 0.01} & 0.53 \scriptsize{$\pm$ 0.03} & 0.49 \scriptsize{$\pm$ 0.02}\\ 
& ft & 0.58 \scriptsize{$\pm$ 0.03} & 0.51 \scriptsize{$\pm$ 0.01} & 0.55 \scriptsize{$\pm$ 0.02} & 0.48 \scriptsize{$\pm$ 0.03}\\ 
& DECAF (Ours) & \textbf{0.71} \scriptsize{$\pm$ 0.17} & \textbf{0.20} \scriptsize{$\pm$ 0.19} & \textbf{0.77} \scriptsize{$\pm$ 0.19} & \textbf{0.27} \scriptsize{$\pm$ 0.15}\\
        \bottomrule
    \end{tabular}
    }
	\end{minipage}\hfill
	\begin{minipage}{0.45\linewidth}
  \centering

\includegraphics[width=0.8\textwidth]{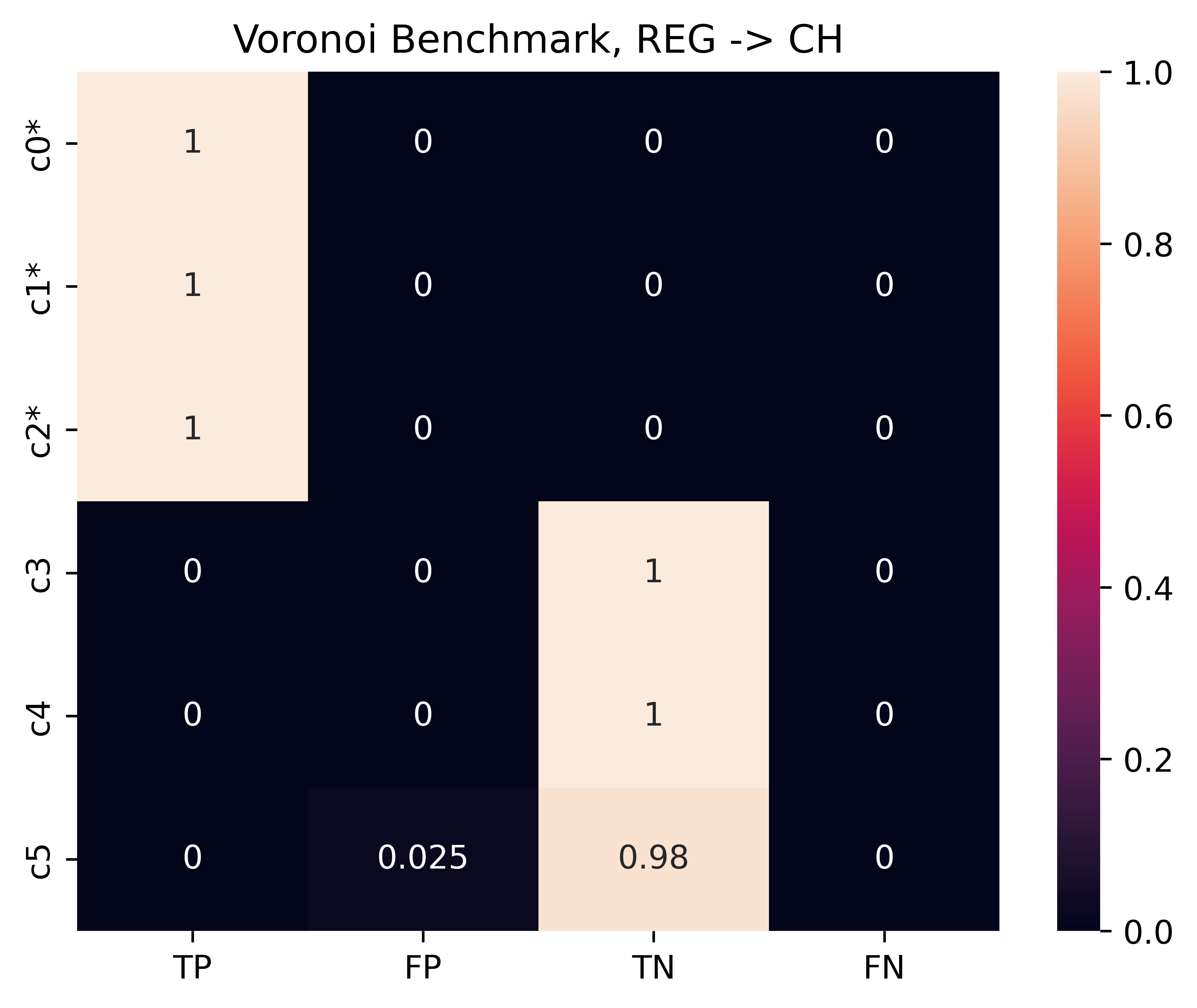}
\addtocounter{figure}{1}
\hypertarget{fig:detection-acc}{}
\end{minipage}
 \captionsetup{labelformat=andfigure}
    \caption{\textbf{Left:} Diag and off-diag metrics for changing factors when adapting CA $\rightarrow$ PO in InterventionalPong dataset. \textbf{Right:} detection confusion matrix for changed causal factors for all CRL approaches in Voronoi Benchmark when moving from REG $\rightarrow$ CH and viceversa, changing factors are indicated with *.\label{tab:results-pong-adapt-ca2po-metrics}}
\end{table}

\noindent\textbf{Voronoi Benchmark.}
We conduct experiments on the change \texttt{REG} $\rightarrow$ \texttt{CH} in the Voronoi Benchmark. Results on 750 data points from the target dataset over five seeds are presented in Figure~\ref{fig:voronoi-re-reg2sh}. Both the baselines and the \OurApproach{} approaches show high dependency on the source-to-target variation, as evidenced by the performance variance. We find that the 0-shot evaluation outperforms the fine-tuning approach, possibly due to the challenging detection of stochastic interventions in low-data regimes. As reported in Appendix~\ref{app:full-results}, fine-tuning with a larger number of target samples benefits adaptation.
Yet, all \OurApproach{} approaches achieve a high CC score and outperform the baselines for CITRIS, LEAP and iVAE, showing its benefit and efficiency of adapting its source representation.
We investigate the detection of changed causal factors on the synthetic changes offered by the Voronoi Benchmark and consider the detection on the change \texttt{REG} $\rightarrow$ \texttt{CH} and the reversed direction. In Figure~\hyperlink{fig:detection-acc}{4}
, we report the aggregated confusion matrix of the variable change detection for all causal representation approaches. \OurApproach{} accurately predicts the changed causal factors in both the directions of the change. The detector always detects changed factors denoted with (*). We observe one failure in recognizing an invariant factor only in one transfer over forty, where one factor is incorrectly predicted as changing. We refer to Appendix~\ref{app:full-results} for the detection accuracy grouped by datasets and causal representation approaches. 

\begin{figure}[t!]
\centering
\sbox\twosubbox{%
  \resizebox{\dimexpr.75\textwidth-1em}{!}{%
    \includegraphics[height=5cm]{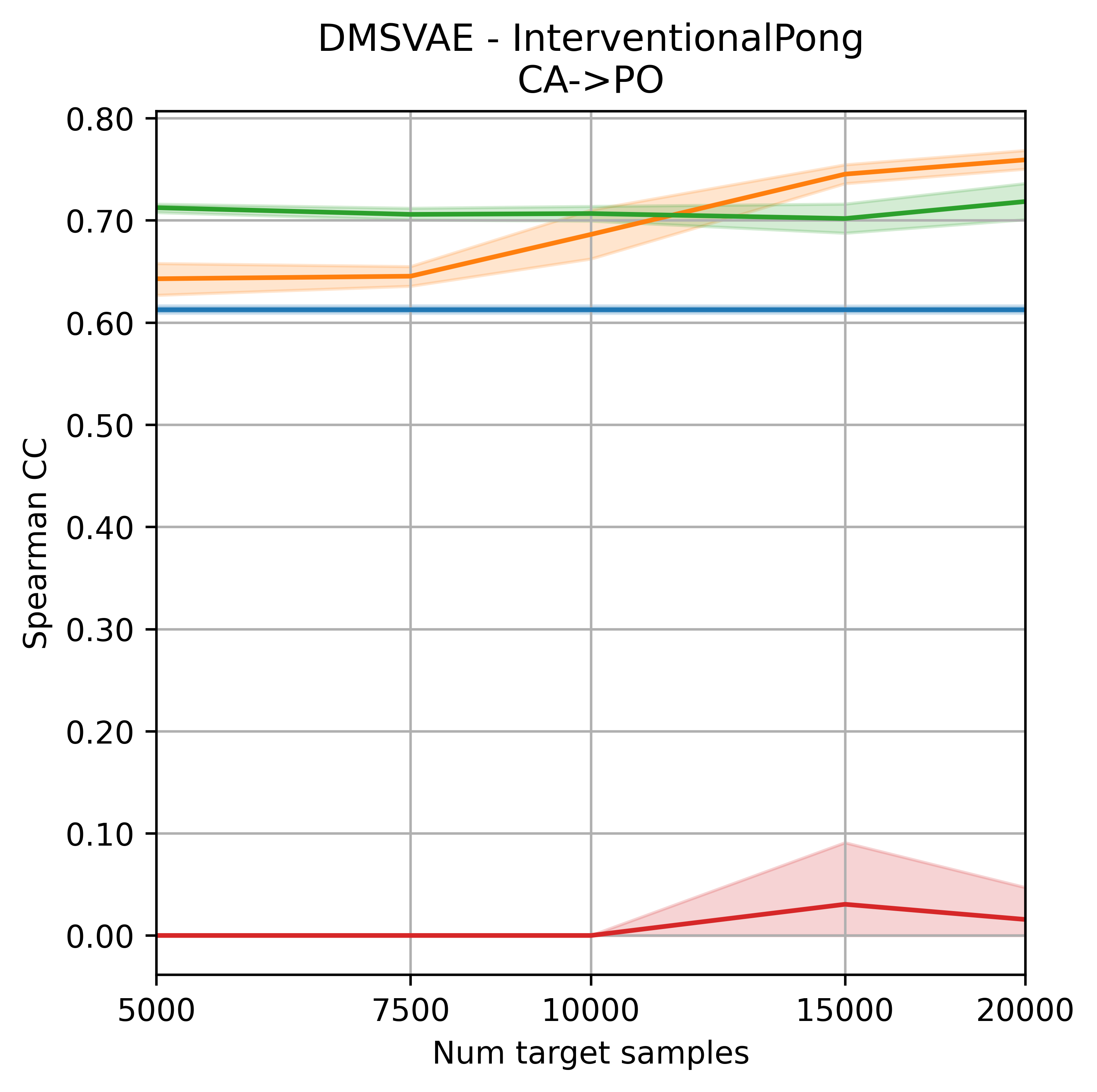}%
    \includegraphics[height=5cm]{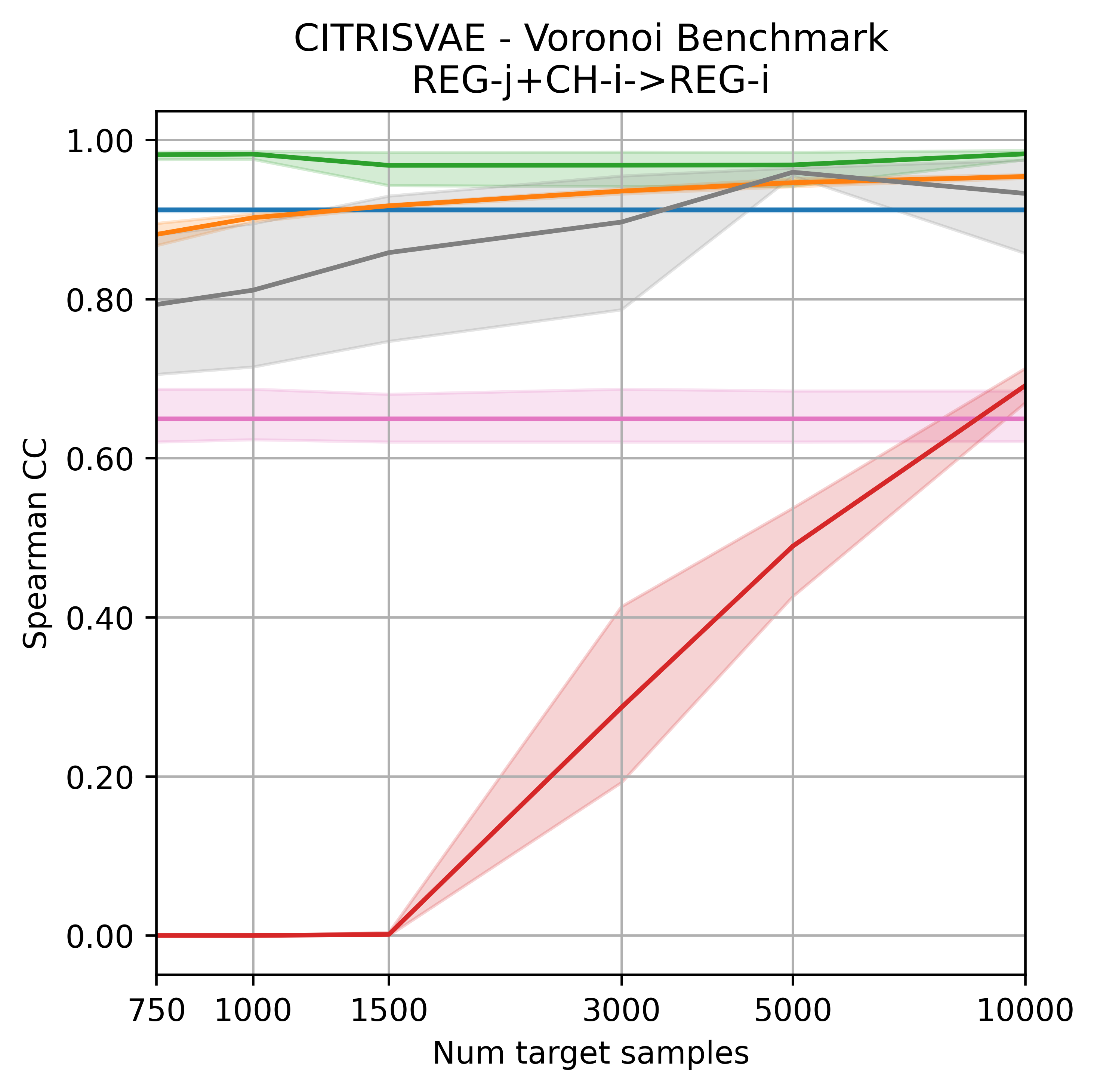}%
  }%
}
\setlength{\twosubht}{\ht\twosubbox}

\centering
\setlength{\twosubht}{\ht\twosubbox}

\subcaptionbox{DMSVAE - InterventionalPong\label{fig:pong-ada-increasing-ca}}{%
  \includegraphics[height=\twosubht]{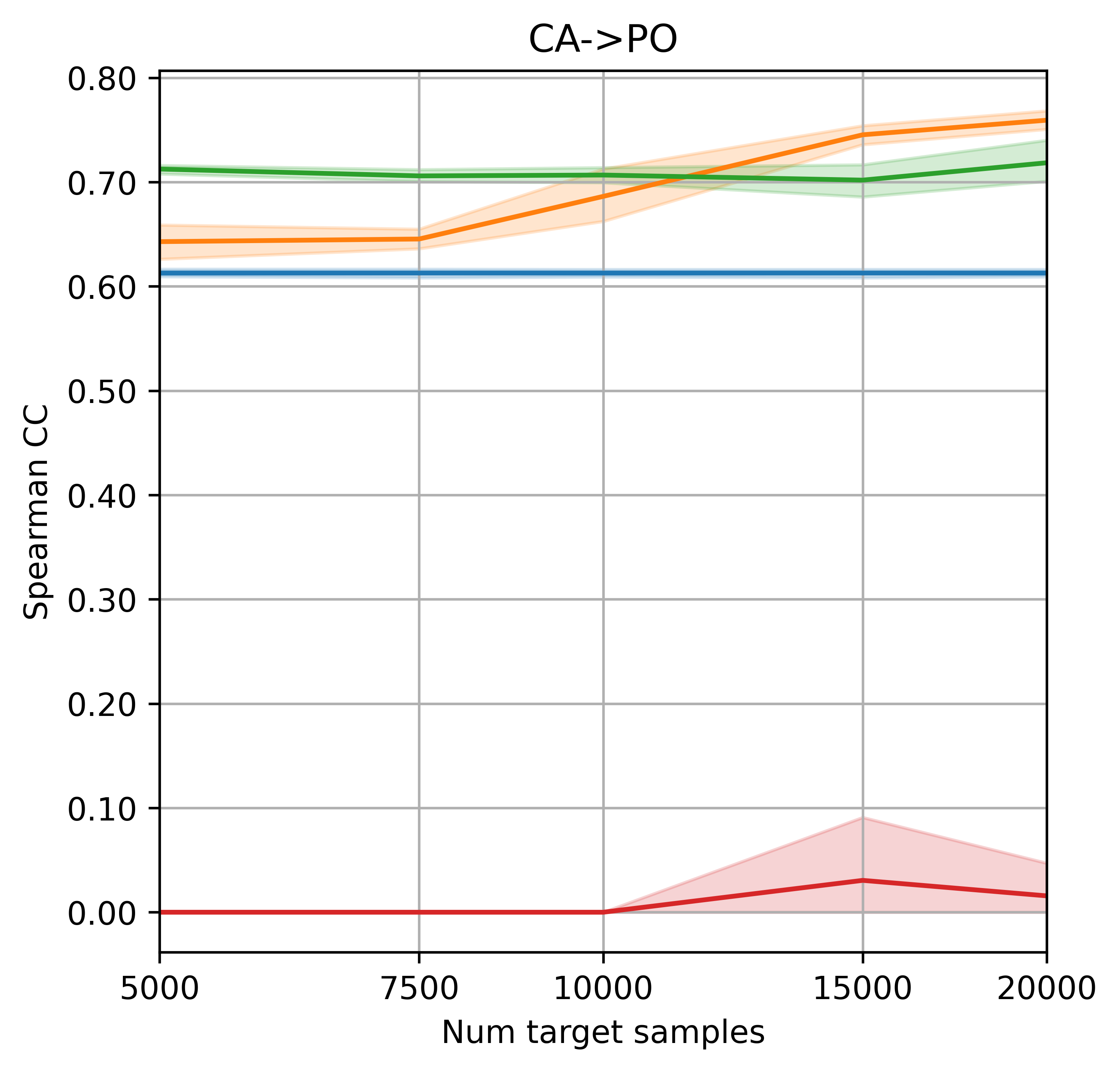}%
}
\subcaptionbox{CITRISVAE - Voronoi Benchmark\label{fig:voronoi-comp-increasing-ca}}{%
  \includegraphics[height=\twosubht]{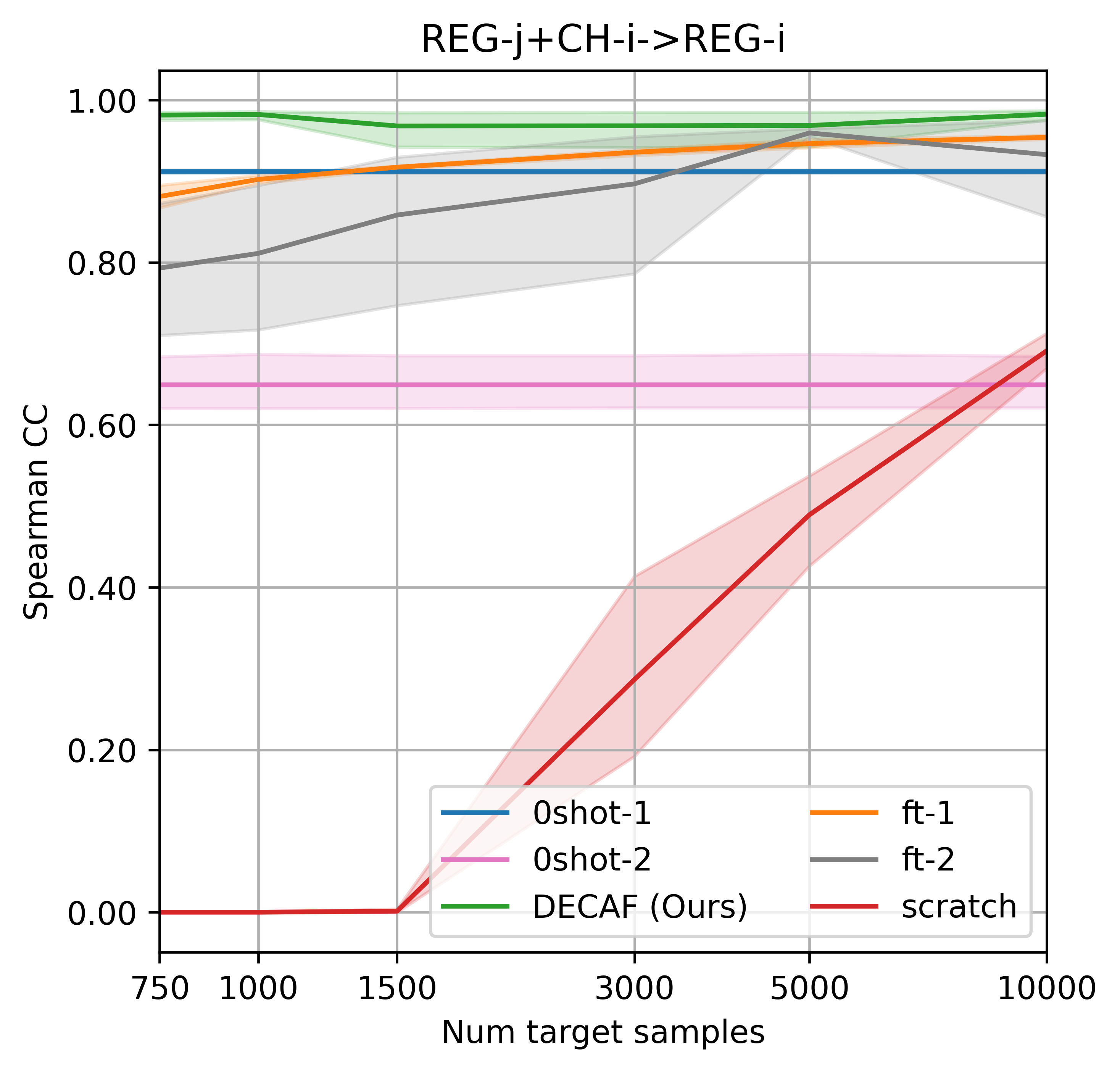}%
}
\caption{Spearman CC (higher best, $\uparrow$) when adaptating and composing with increasing number of samples. Solid lines describe the mean while shaded areas the standard deviation over 5 runs. \textbf{(a)} Correlation of changed factors for DMSVAE approach when adapting from CA $\rightarrow$ PO in InterventionalPong. \textbf{(b)} Correlation of all factors when composing REG-j+CH-i$\rightarrow$REG-i in Voronoi Benchmark.}
\end{figure}

\smallskip
\noindent\textbf{InterventionalPong.}
In the InterventionalPong dataset, we change the ball coordinate system from Cartesian in the source domain to polar in the target ($\texttt{CA} \rightarrow \texttt{PO}$), providing 5K samples in the target environment. For all considered CRL methods, the combination with \OurApproach{} outperforms the adaptation baselines as seen in Figure~\ref{fig:pong-re-ca2po}, although notable performance drops are observed for both iVAE and LEAP methods, where for one seed, the classifier fails to separate intervention targets. As reported in Table~\ref{tab:results-pong-adapt-ca2po-metrics}, the correlation metrics \texttt{diag} and \texttt{off\_diag} show that the approach achieves improved performance for both the metrics. We observe the largest benefit for CITRISVAE that gains 25\% and 19\% in the Spearman \texttt{diag} and \texttt{off\_diag}, respectively. Though data efficient, the learnt latent-to-factors assignment matrix in CITRISVAE limits its adaptation to new environments. \OurApproach{} aids separation of causal factors, especially for iVAE where the $R^2$ \texttt{off\_diag} improves of 31\%.
With more samples, \texttt{ft} can catch up to \OurApproach{}, as seen in Figure~\ref{fig:pong-ada-increasing-ca}. Yet, the low performance of training from scratch shows the importance of adaptive representations.

\smallskip
\noindent\textbf{Temporal Causal3DIdent.}
Finally, in Temporal Causal3DIdent, we investigate the adaptation of the object position variables to a ROTated (ROT) x-y coordinate system, $\texttt{CA} \rightarrow \texttt{ROT}$ with 1K total samples, see Figure~\ref{fig:c3d-re-ca2rot} averaged over 5 seeds. As can be noted, \OurApproach{} is still competitive in the more visually complex scenario. While fine-tuning proves effective for CITRISVAE, in the other settings \OurApproach{} improves over the baselines, with a large gain of about 10\% in iVAE. The DMSVAE approach exhibits high variance, with three runs detecting only one of the two changed variables. Notably, the high zero-shot performance shows how the two environments are well aligned one to the other, motivating the incorrectly detected changes and ease of adaptation for the \texttt{ft} baseline.

\subsection{Composition of causal representations}
\begin{figure}[]
\centering
\sbox\twosubbox{%
  \resizebox{\dimexpr.99\textwidth-1em}{!}{%
    \includegraphics[height=5cm]{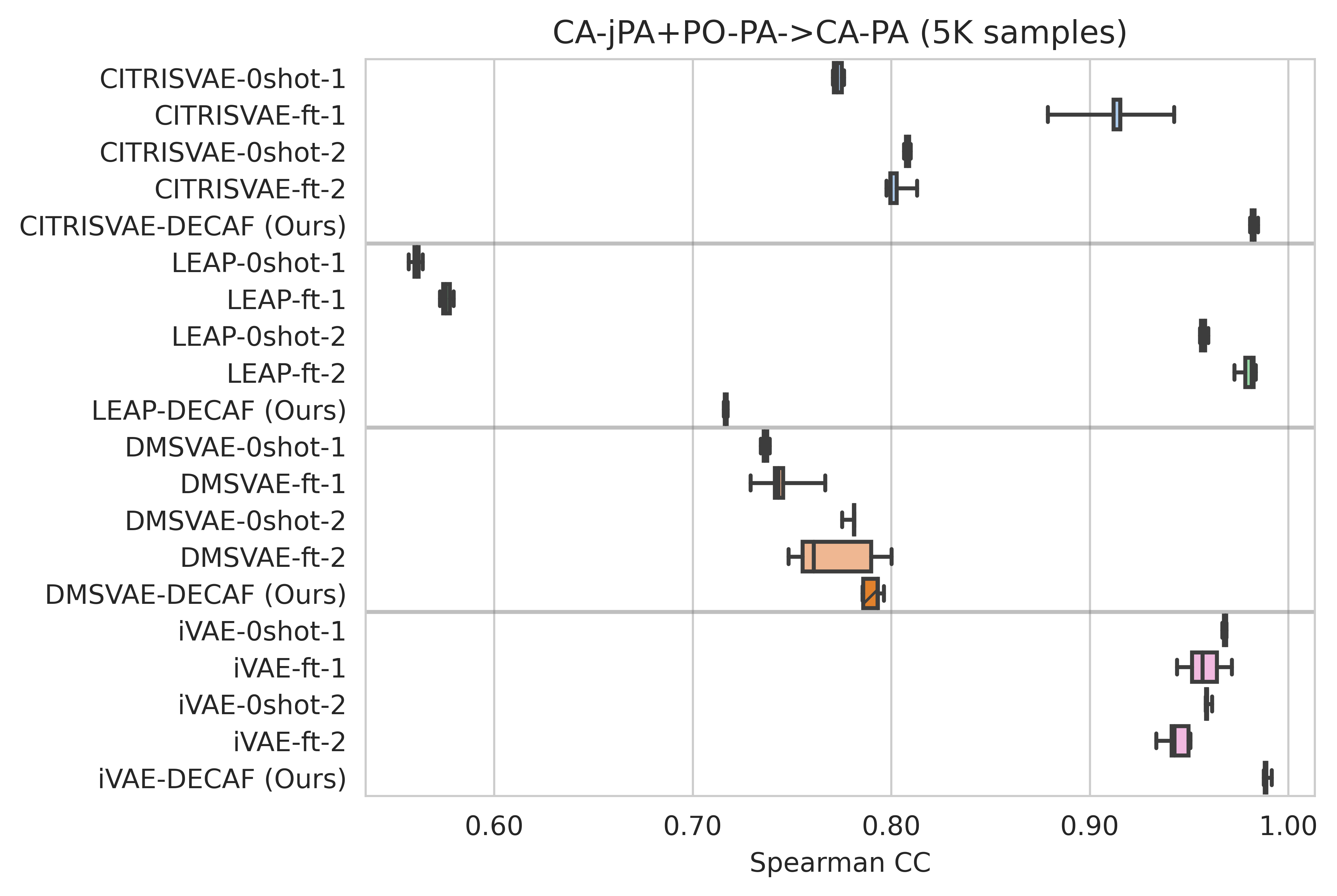}%
    \includegraphics[height=5cm]{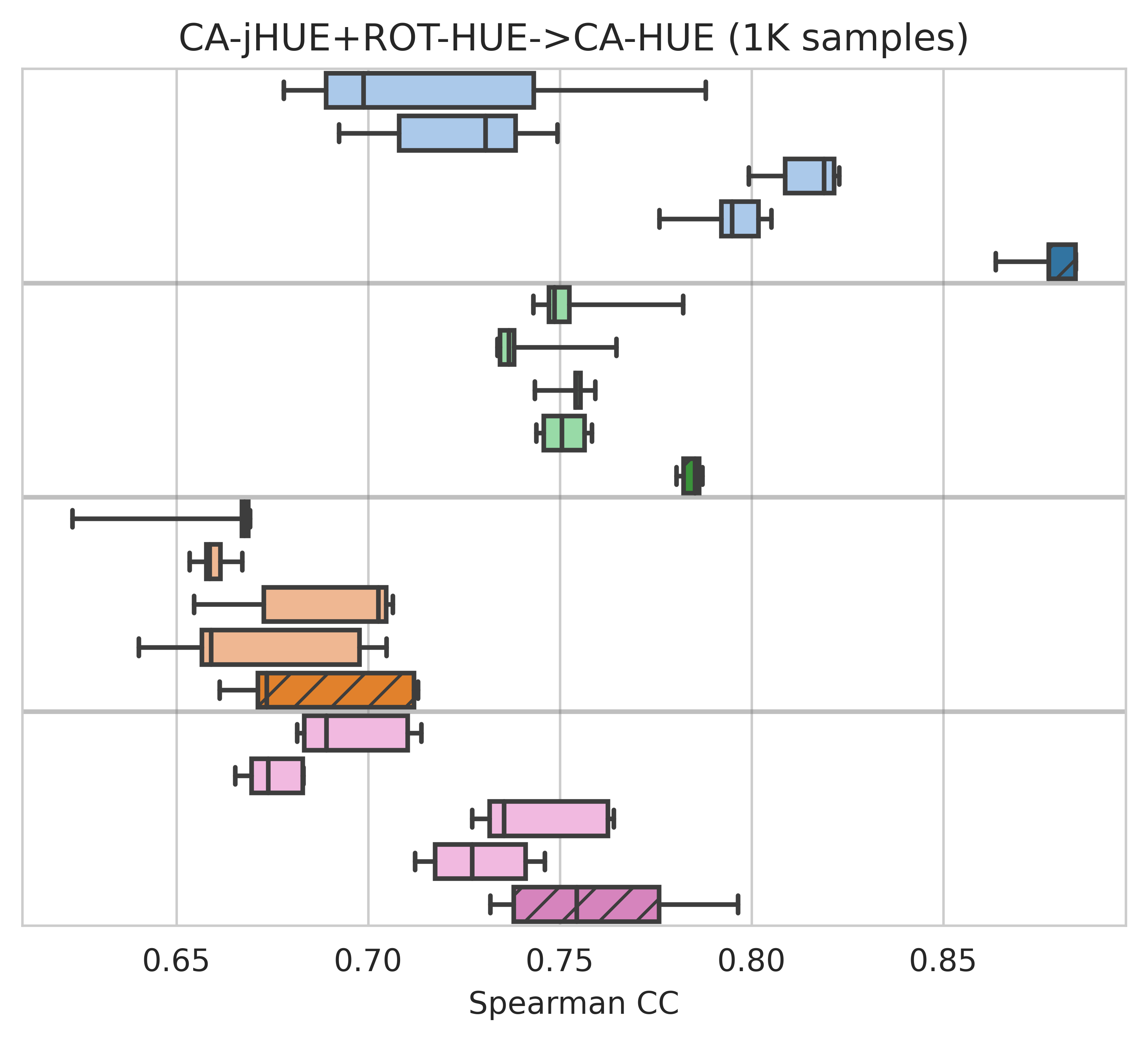}%
  }%
}
\setlength{\twosubht}{\ht\twosubbox}

\centering
\setlength{\twosubht}{\ht\twosubbox}

\subcaptionbox{InterventionalPong\label{fig:pong-comb-ca}}{%
  \includegraphics[height=\twosubht]{clear/results/results-pong-comb-ca.png}%
}
\subcaptionbox{Temporal Causal3DIdent\label{fig:c3d-comb-ca}}{%
  \includegraphics[height=\twosubht]{clear/results/results-c3d-comb-ca.png}%
}
\caption{Spearman CC (higher best, $\uparrow$) of inferred latents to the ground truth of all variables when composing representations. CRL approaches are color-coded, the proposed method has a darker color. \textbf{(a)} Composition of factors in InterventionalPong with sources \texttt{CA-jPA} and \texttt{PO-PA}. \textbf{(b)} Composition of factors in Causal3DIdent with sources \texttt{CA-jHUE} and \texttt{ROT-HUE}. }
\end{figure}

\noindent\textbf{InterventionalPong.}
We evaluate composing the causal representation in InterventionalPong where a new target environment models the ball position and the paddles accordingly to the factors of the first and second environment, respectively. We consider the first source environment S1 to model the ball-position with Cartesian coordinates but entangled paddles (CA-jPA) and the second environment S2 with polar ball position and independently intervened paddles (PO-PA). We aim to identify the causal factors in a target environement with Cartesian ball position and independent paddles, \texttt{CA}-\texttt{jPA}+\texttt{PO}-\texttt{PA}$\rightarrow$ \texttt{CA}-\texttt{PA}. Figure~\ref{fig:pong-comb-ca} shows the Spearman Combined Correlation when addressing the new composed environment. \OurApproach{} improves with respect to the baselines in all but LEAP causal representation approaches. \OurApproach{} highly depends on the quality of the source representations, as shown with the LEAP composition, where the source S1 CA-jPA fails to identify the position of the ball in its Cartesian coordinates. Nevertheless, \OurApproach{} correctly recognizes changing factors and building on the invariant factors from S2, improves over S1 transfers. However, the proposed approach falls behind adaptation from polar coordinates due to its good initial 0shot alignment. 

\smallskip
\noindent\textbf{Causal3DIdent.}
In the Temporal Causal3DIdent dataset, we consider two source environments: one with Cartesian position and jointly intervened hue (CA-jHUE), and another with a rotated coordinate system for position but independent interventions on hue (ROT-HUE), \texttt{CA}-\texttt{jHUE}+\texttt{ROT}-\texttt{HUE}$\rightarrow$ \texttt{CA}-\texttt{HUE}. 
The target environment composes the Cartesian position of the first source environment with the individual hue variables of the second environment, requiring the algorithms to identify which variables can be reused and combined from the sources.
As shown in Figure~\ref{fig:c3d-comb-ca}, \OurApproach{} finds the correct variables to compose and, especially for CITRISVAE, provides significant gains over the baselines, while only requiring 1k samples. Composing the representations with \OurApproach{} benefits the identification performance on both LEAP and iVAE approaches, while we observe that most of the variance overlaps for DMSVAE representations.

\smallskip
\noindent\textbf{Voronoi Benchmark.}
We assess the benefit of \OurApproach{} in the composition setting as we increase the number of target samples. The target combines the first three \texttt{REG} variables from S1 with the last two independently intervened factors \texttt{i} from S2, \texttt{REG-j}+\texttt{CH-i} $\rightarrow$ \texttt{REG-i}. As can be noted in Figure~\ref{fig:voronoi-comp-increasing-ca}, \OurApproach{} leverages the target samples only for the detection of changing causal factors and achieves close to perfect Spearman CC starting from 750 samples. The disentanglement is stable as we increase the number of samples, and it is competitive up to 10K samples. We observe that the proposed approach improves over all considered baselines. Notably, adaptation of representations and \OurApproach{} composition strategy outperform training from scratch on the target domain, showing the advantage of re-using previously learnt causal factors in the new environment.
\section{Related Work}
\label{app:extended-related-work}
\noindent\textbf{Disentanglement.} A lot of effort in representation learning has been devoted to obtaining a compact lower-dimensional representation of data as product of factors of variation \cite{bengio2013representation}. However, the general assumption of independent latents does not remove spurious solutions \cite{locatello2019challenging}. To overcome identifiability limitations, unsupervised disentanglement has been relaxed to employ some form of supervision \cite{locatello2020weakly, Locatello2020Disentangling}. Recent works from the nonlinear ICA~\cite{comon1994independent, hyvarinen1999nonlinear} have built on non-stationarity~\cite{hyvarinen2016unsupervised}, auxiliary information leading to conditionally independent latents~\cite{hyvarinen2019nonlinear}, or assumptions on the mixing function \cite{zhengidentifiability, gresele2021independent}.  As the independence assumption is often not met in real data it hinders the generalization capabilities of these methods \cite{dittadi2020transfer,trauble2021disentangled, dittadi2021generalization, roth2022disentanglement}. In contrast, this work allow for potential causal relationships among latents and investigates how to adapt the previously learnt representation to address a new unseen target environment.

\smallskip
\noindent\textbf{Causal Representation Learning (CRL).}
Recent work in CRL \cite{lippe2022citris, lachapelle2022disentanglement, yao2021learning, lippe2022icitris, lippe2023biscuit, yao2022temporally} identify causal variables and relations in a temporal sequence setting where a system may be affected by interventions, i.e., we have access to consequent observations and performed actions relating them. In this setting, \citet{lippe2022citris} consider multidimensional causal factors and leverages known intervention targets to disentangle them. In contrast, DMSVAE~\cite{lachapelle2022disentanglement} builds on recent results on nonlinear ICA to show how sparsity in the transition function and intervention targets constrain the problem to be identifiable. Building on non-stationarity and independence of exogenous noises, LEAP~\cite{yao2021learning} identifies causal factors thanks to an observable auxiliary variable that modulates the noise distribution in different regimes.
\citet{brehmer2022weakly} consider a counterfactual learning scenario instead. Following previous work in multi-view nonlinear ICA~\cite{locatello2020weakly, von2021self}, they cast the problem as a weakly supervised generative process where we observe samples before and after atomic and perfect interventions.
As opposed to the CRL literature that does not consider available pre-trained representations to identify the causal factors in a new domain, this work focuses on a multi-environment setting where we can transfer from source representations and leverage them to address a target environment where few samples are present.

\smallskip
\noindent\textbf{Modularity.}  
Modular causally-inspired representations have been explored in the literature. \citet{parascandolo2018learning} investigate how training with a winner-takes-all scheme guides the specialization of causal mechanisms: independent modules compete on observed samples and only those maximizing an heuristic activation function get updated. By specializing on their input, a modular representation emerges and reverses the effect of the unknown generating mechanisms. Similarly, an explicitly modular architecture composed of multiple almost-independent subsystems model a dynamic setting in \cite{goyal2019recurrent}. The subsystems compete based on the strength of their activations on the observed input, and most firing ones update their internal state.
\citet{besserve2018counterfactuals} exploit unit-level counterfactual statements to seek for modular structure in generative models and define disentanglement based on available transformations of data.
No guarantees are provided on the recovered modules \eg a module can model multiple causal variables at the same time. In contrast, this work does not seek for a modular representation of the data: we build on the modular nature of causal representations where causal variables are identified up to an explicit identification class to investigate how to detect changed causal factors and how to alter the representation to address a new related environment.

\section{Conclusions}
We introduce \OurApproach{}, a framework that is a first step towards adapting and composing causal representations. Our approach detects changing causal variables in a new environment and provides a method to adapt them with a limited amount of target samples. Experimental results on three datasets show the benefit of re-using and composing learnt causal representations when applied to different causal representation approaches.
\OurApproach{} constructs accurate target representations. We envision a setup where a bank of re-usable factors are available. Future work involves leveraging the available causal factors to aid learning of the dynamics in the new domain, identifying changed causal factors and relaxing the assumption on the observation of intervention targets.

\acks{We would like to thank Fan Feng, Gianluca Scarpellini and Andrea Maracani for the useful discussions. We gratefully acknowledge the HPC infrastructure and the Support Team at Fondazione Istituto Italiano di Tecnologia.}

\bibliography{main}

\newpage
\appendix
\section{Implementation details and hyperparameters}
\label{app:experimental-details}
In this section we provide further details on the implementation. 

\label{app:implementation-details}
\subsection{Source models}
\paragraph{VAE architecture} In order to have the different CRL approaches achieving their highest performance on source environments, we tested different variants of the same architecture. A convolutional encoder outputs the mean and standard deviation parameters of independent Gaussians. After sampling, the embeddings are decoded for reconstruction. For computational reasons, the specific architecture depends on the dataset. In Voronoi Benchmark and InterventionalPong the encoder is a 5-layer CNN + 2-layer MLP with a hidden dimension of 32. The decoder uses a symmetric architecture to the encoder (2-layer MLP and 5-layer deconv). In Temporal Causal3DIdent we followed the architecture in~\cite{lippe2022citris} and employed a 10-layer CNN  and a 10-layer Resnet~\cite{he2016deep} decoder with a hidden dimension of 64.

On Voronoi Benchmark and Interventional Pong datasets we found an autoregressive flow prior \cite{rezende2015variational, kingma2016improved} to be beneficial on CITRISVAE and DMSVAE, following the architecture in ~\cite{lippe2022citris}. The Gaussian samples from the encoder are fed to a 4-layer normalizing flow including  Activation Normalization~\cite{kingma2018glow}, Invertible $1 \times 1$ convolutions~\cite{kingma2018glow} and autoregressive affine coupling layers.

\paragraph{Transition prior} The transition prior accepts as input the current time step and some auxiliary information to predict the next time step. In CITRISVAE the transition prior is a 2-layer MLP fed with $z^t$ and $I^{t+1}$ as input to predict $z^{t+1}$. Other baselines employ a 3-layer MLP. Following~\cite{lippe2022citris}, we adapted the iVAE prior to accept as input the concatenation of the current time step $z^t$ and the intervention target $I^{t+1}$. Similarly, both LEAP and DMSVAE priors accept as input a masked version of the concatenation $[z^t, I^{t+1}]$  where the mask is learned during training. Due to the density of the temporal graph of both Voronoi Benchmark and InternventionalPong, we found that restricting DMSVAE to learn the action mask only proved beneficial for the approach. 

All the source models are trained with a batch size of 512 samples using AdamW~\cite{loshchilov2017decoupled} optimization with a learning rate of 1e-3 and Cosine Warmup scheduler. We used the Swish~\cite{hendrycks2016gaussian, ramachandran2017searching} non-linearity. We regularized the source models to avoid overfitting on the source data by controlling for the source training epochs and adding a $L^2$-norm loss on the representation with hyperparameter $\beta_{\text{reg}}$. We summarized the used hyperparameters in Tab.~\ref{tab:app-hyperparams-source}.

\begin{table}[t!]
    \centering
    \caption{Summary of the hyperparameters for all source models trained on the Voronoi benchmark, InterventionalPong and Temporal Causal3DIdent dataset,}
    \label{tab:app-hyperparams-source}
    \small
    \begin{tabular}{lcccc}
        \toprule
        \multicolumn{5}{c}{\cellcolor{gray!25}{\textbf{Voronoi benchmark/InterventionalPong}}}\\
        \text{Hyperparameter} & \text{CITRISVAE} & \text{LEAP} & \text{DMVAE} & \text{iVAE} \\
        \midrule
        Learning rate & \multicolumn{4}{c}{---- 1e-3 ----}\\
        Learning rate warmup & \multicolumn{4}{c}{---- Cosine Warmup (100 steps) ----}\\
        Optimizer & \multicolumn{4}{c}{---- AdamW \cite{loshchilov2017decoupled} ----}\\
        Batch size & \multicolumn{4}{c}{---- 512 ----}\\
        Number of epochs & 75(V)/125(P) & 100(V)/200(P) & 75(V)/175(P) & 100(V)/ 200(P)\\
        KLD Factor ($\beta$) & \multicolumn{3}{c}{---- 1.0 ----} & 0.5\\
        Num latents & \multicolumn{4}{c}{---- 16 ----}\\
        Model variant & VAE+NF & VAE & VAE+NF & VAE \\
        Encoder & \multicolumn{4}{c}{---- 5 layer CNN + 2 linear layers ----}\\
        Prior layers & 2 & 3 & 3 & 3 \\
        Decoder & \multicolumn{4}{c}{---- 5 layer (deconv-)CNN + 2 linear layers ----}\\
        Hidden dimensionality & \multicolumn{4}{c}{---- 32 ----}\\
        Activation function & \multicolumn{4}{c}{---- Swish \cite{ramachandran2017searching} ----}\\ 
        Target classifier weight & 2 & \multicolumn{3}{c}{---- n.a. ----} \\
        Sparsity regularizer  & n.a & \multicolumn{2}{c}{---- 0.01 ----} & n.a. \\
        Discriminator weight  & n.a & 0.05 & \multicolumn{2}{c}{---- n.a. ----} \\
        \bottomrule
        & \\
        \multicolumn{5}{c}{\cellcolor{gray!25}{\textbf{Temporal Causal3DIdent dataset}}}\\
        \text{Hyperparameter} & \text{CITRISVAE} & \text{LEAP} & \text{DMVAE} & \text{iVAE} \\
        \midrule
        Learning rate & \multicolumn{4}{c}{---- 1e-3 ----}\\
        Learning rate warmup & \multicolumn{4}{c}{---- Cosine Warmup (100 steps) ----}\\
        Optimizer & \multicolumn{4}{c}{---- AdamW \cite{ramachandran2017searching} ----}\\
        Batch size & \multicolumn{4}{c}{---- 512 ----}\\
        Number of epochs & \multicolumn{4}{c}{---- 600 ----} \\
        KLD Factor ($\beta$) & \multicolumn{4}{c}{---- 1 ----}\\
        Num latents & \multicolumn{4}{c}{---- 32 ----}\\
        Model variant & VAE+NF & VAE & VAE+NF & VAE \\
        Encoder & \multicolumn{4}{c}{---- 10-layer CNN ----}\\
        Prior layers & 2 & 3 & 3 & 3 \\
        Decoder & \multicolumn{4}{c}{---- 10-layer ResNet ----} \\
        Hidden dimensionality & \multicolumn{4}{c}{---- 64 ----}\\
        Activation function & \multicolumn{4}{c}{---- Swish \cite{ramachandran2017searching} ----}\\ 
        Target classifier weight & 2 & \multicolumn{3}{c}{---- n.a. ----} \\
        Sparsity regularizer  & n.a & \multicolumn{2}{c}{---- 0.01 ----} & n.a. \\
        Discriminator weight  & n.a & 0.1 & \multicolumn{2}{c}{---- n.a. ----} \\
        \bottomrule
    \end{tabular}
\end{table}

\begin{table}[t!]
    \centering
    \caption{Summary of the hyperparameters used for addressing the target environment.}
    \label{tab:app-hyperparams-adapt}
    
    \small
    \begin{tabular}{lcccc}
    \toprule
        \multicolumn{4}{c}{\textbf{\cellcolor{gray!25}{\OurApproach{} Adaptation}}}\\
        \text{Hyperparameter} & \text{Voronoi Benchmark} & \text{InterventionalPong} & \text{Temporal Causal3DIdent}  \\
        \midrule
        Learning rate & \multicolumn{3}{c}{---- 1e-2 ----}\\
        Learning rate warmup & \multicolumn{3}{c}{---- Cosine Warmup (100 steps) ----}\\
        Optimizer & \multicolumn{3}{c}{---- AdamW \cite{loshchilov2017decoupled} ----}\\
        Batch size & \multicolumn{3}{c}{---- 1024 ----}\\
        Number of epochs & \multicolumn{3}{c}{---- 5000 ----}\\
        KLD Factor ($\beta$) & \multicolumn{3}{c}{---- 1 ----}\\
        Hidden dimensionality & \multicolumn{3}{c}{---- 64 ----}\\
        Activation function & \multicolumn{3}{c}{---- Swish \cite{ramachandran2017searching} ----}\\ 
        Target classifier weight & \multicolumn{3}{c}{---- 2 ----} \\
        Num flows & 2 & \multicolumn{2}{c}{---- 4 ----} \\
        At Least one ($\beta_{\text{ALO}}$) & 4 & \multicolumn{2}{c}{---- 2 ----} \\
        $L^2$-Norm regularizer ($\beta_{\text{reg}}$) & 4 & \multicolumn{2}{c}{---- 2 ----} \\
        Changed module threshold ($\tau$) & 0.15 & 0.2 & 0.1 \\
        \bottomrule
        & \\
        \multicolumn{4}{c}{\cellcolor{gray!25}{\textbf{Fine-tuning}}}\\
        \text{Hyperparameter} & \text{Voronoi Benchmark} & \text{InterventionalPong} & \text{Temporal Causal3dIdent}  \\
        \midrule
        Learning rate & \multicolumn{3}{c}{---- 1e-3 ----}\\
        Learning rate warmup & \multicolumn{3}{c}{---- Cosine Warmup (100 steps) ----}\\
        Optimizer & \multicolumn{3}{c}{---- AdamW \cite{loshchilov2017decoupled} ----}\\
        Batch size & \multicolumn{3}{c}{---- 512 ----}\\
        Number of epochs & \multicolumn{3}{c}{---- 2500 ----}\\
        \bottomrule
        & \\
        \multicolumn{4}{c}{\cellcolor{gray!25}{\textbf{\OurApproach{} Composition}}}\\
        \text{Hyperparameter} & \text{Voronoi Benchmark} & \text{InterventionalPong} & \text{Temporal Causal3DIdent}  \\
        \midrule
        Learning rate & \multicolumn{3}{c}{---- 1e-3 ----}\\
        Learning rate warmup & \multicolumn{3}{c}{---- Cosine Warmup (100 steps) ----}\\
        Optimizer & \multicolumn{3}{c}{---- AdamW \cite{loshchilov2017decoupled} ----}\\
        Batch size & \multicolumn{3}{c}{---- 512 ----}\\
        Number of epochs & \multicolumn{3}{c}{---- 10 ----}\\
        Changed module threshold ($\tau$) & 0.15 & 0.2 & 0.1 \\
        \bottomrule
    \end{tabular}
    
\end{table}

\subsection{Adaptation and Composition}
\paragraph{Fine-tuning.} The fine-tuning approach resumes the training of the model with the same causal representation strategy of the source model, \eg a model pre-trained with the CITRISVAE strategy adapts to the new environment using the same CITRISVAE algorithm. Fine-tuning adapts the model with 2500 epochs and a batch size of 512 using AdamW optimizer with a learning rate of 1e-3 and Cosine Warmup scheduler. 

\paragraph{Adaptation.} We implemented the adaptation approach using an autoregressive normalizing flow \cite{rezende2015variational} following~\cite{lippe2022citris}. The flow is based on the MADE \cite{germain2015made} architecture with 16 neurons per latent variable. The flow includes Activation Normalization and $1 \times 1$ invertible convolutions. The depth of the flow depends on the dataset. As a flow prior, we employed a 2-layer autoregressive network that follows the same MADE architecture as the normalizing flow. For each latent variable, the flow outputs the parameters of a Gaussian distribution. \OurApproach{} adapts the model in 5000 epochs with a batch size of 1024 samples. We optimize using AdamW with a learning rate of 1e-2 and weight decay 5e-3. We applied the same Cosine Warmup scheduler as in the fine-tuning strategy.

\paragraph{Composition.} 
\OurApproach{} stitches together the causal factor representations of modules that are detected to be invariant with respect to the target environement. Since the latent to factors assignment allows for a variable number of latents per factor, we cannot guarantee that the resulting representation matches the dimensionality of the pretrained autoencoder. To this end, we learn a projection function $\rho$ projecting the representation to the same dimensionality as the source embedding. Thus, we freeze the representation model and learn it on the source data  via reconstruction. In practice we parameterize $\rho$ with a 2-layer feedforward network having 128 hidden dimensionality and Swish non-linearity. The projection function is trained with AdamW, a learning rate of 1e-3 and batch size 512.

We report the hyperparameters used for adaptation in Tab.~\ref{tab:app-hyperparams-adapt}.

\begin{figure}[h!]
\centering
  \includegraphics[width=\linewidth]{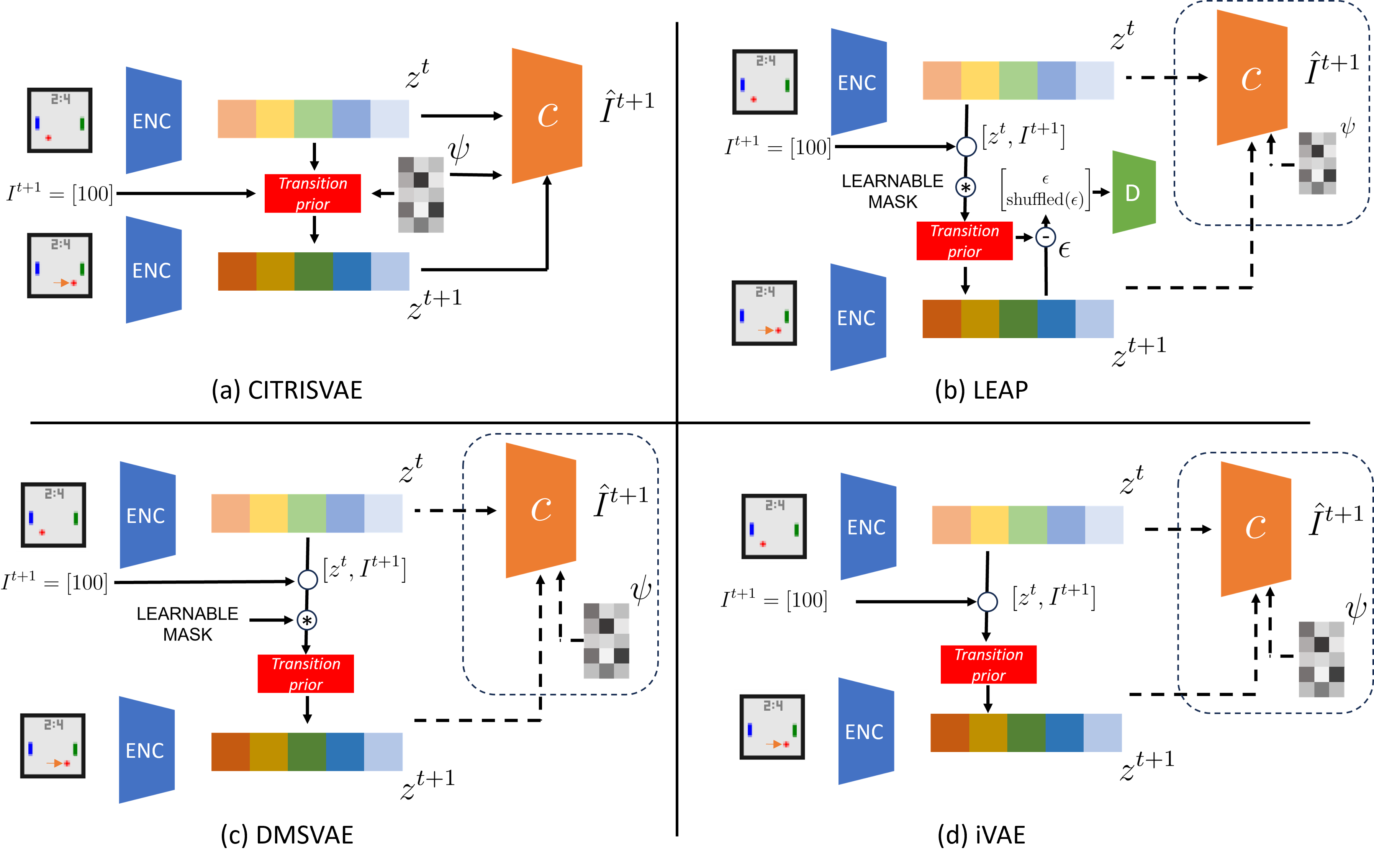}
  \label{fig:crl-approaches}
\caption{Visualization of the adaptation of the different CRL appoaches for the TRIS setting and learning of the target classifier on a pre-trained representation. (a) CITRISVAE~\cite{lippe2022citris} is used as is and makes available the classifier and the assignment $\psi$ for later re-use. (b) LEAP~\cite{yao2021learning}: since intervention targets are a source of non-stationarity, the previous time step and the intervention target are concatenated and masked to condition the LEAP transition prior. (c) DMSVAE~\cite{lachapelle2022disentanglement} conditions the transition prior on the concatenation of previous time step and intervention targets, masked according to the learnt graph. (d) iVAE~\cite{khemakhem2020variational} conditions the prior on the concatenation of previous time step and intervention targets. For LEAP, DMSVAE and iVAE  we learn a target classifier and the assignment $\psi$ on top of the frozen representation.}
\end{figure}
e\section{Full Results}
\label{app:full-results}
Tables~\ref{tab:app-voronoi-re-reg2ch},~\ref{tab:app-pong-re-ca2po}, and~\ref{tab:app-c3d-re-ca2rot} report the complete results on the adaptation setting using the correlation diagonal (diag) and off-diagonal (off-diag). Similarly, Tables~\ref{tab:app-voronoi-co-reg},~\ref{tab:app-pong-co-ca},~\ref{tab:app-c3d-co-ca} report the correlation metrics for the considered composition settings.

\clearpage
\subsection{Other composition settings}
\begin{figure}
\centering
\includegraphics[width=0.8\textwidth]{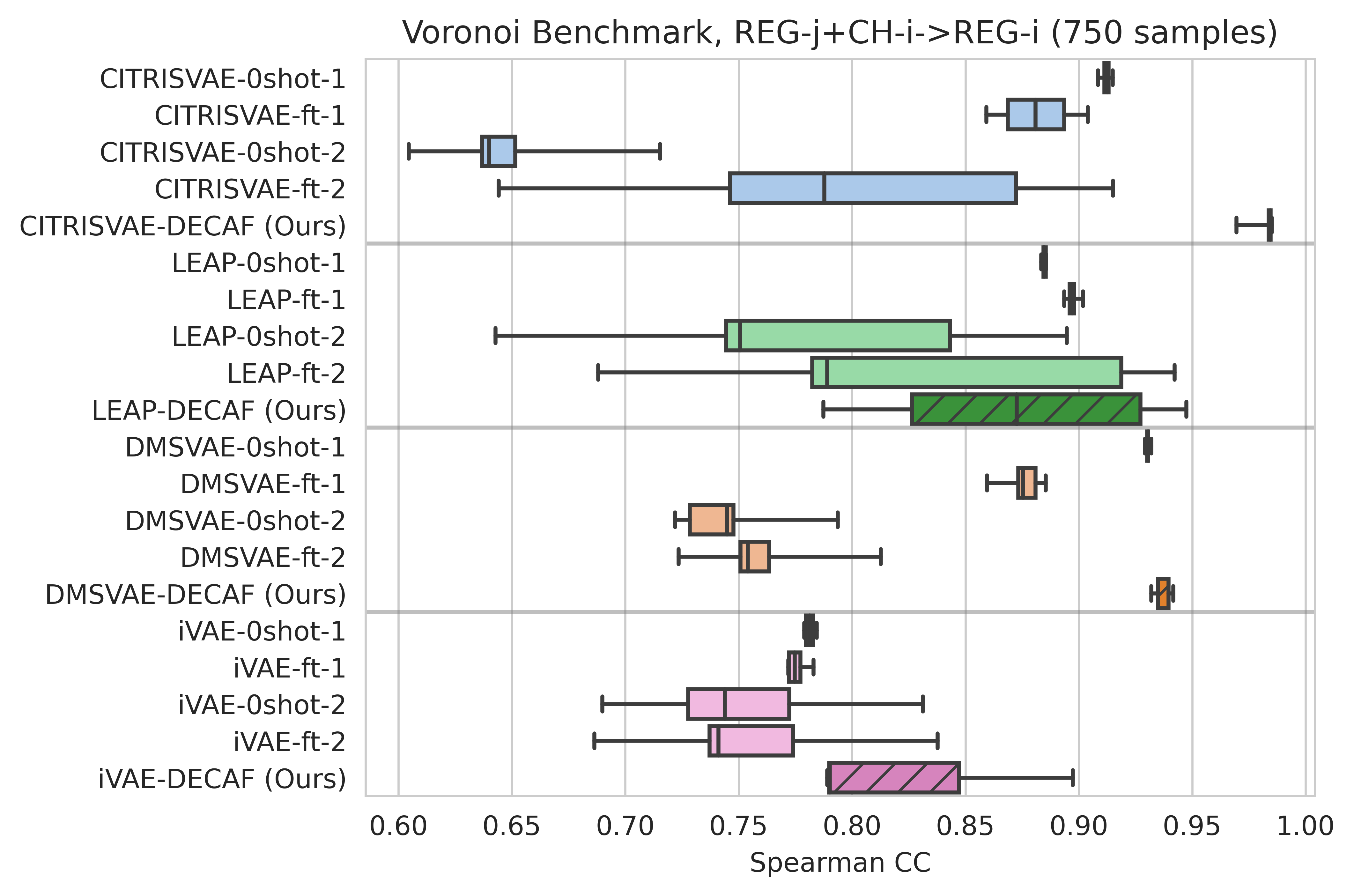}
\caption{Spearman CC ($\uparrow$) of inferred latents to the ground truth of all variables when composing representations in Voronoi Benchmark with sources \texttt{REG-j} and \texttt{CH-i}}
\label{fig:app-results-voronoi-composition}
\end{figure}

In Figure~\ref{fig:app-results-voronoi-composition} we report results on the composition setting in Voronoi Benchmark.

\subsection{Increasing number of target samples}
\begin{figure}
\centering
\includegraphics[width=0.3\textwidth]{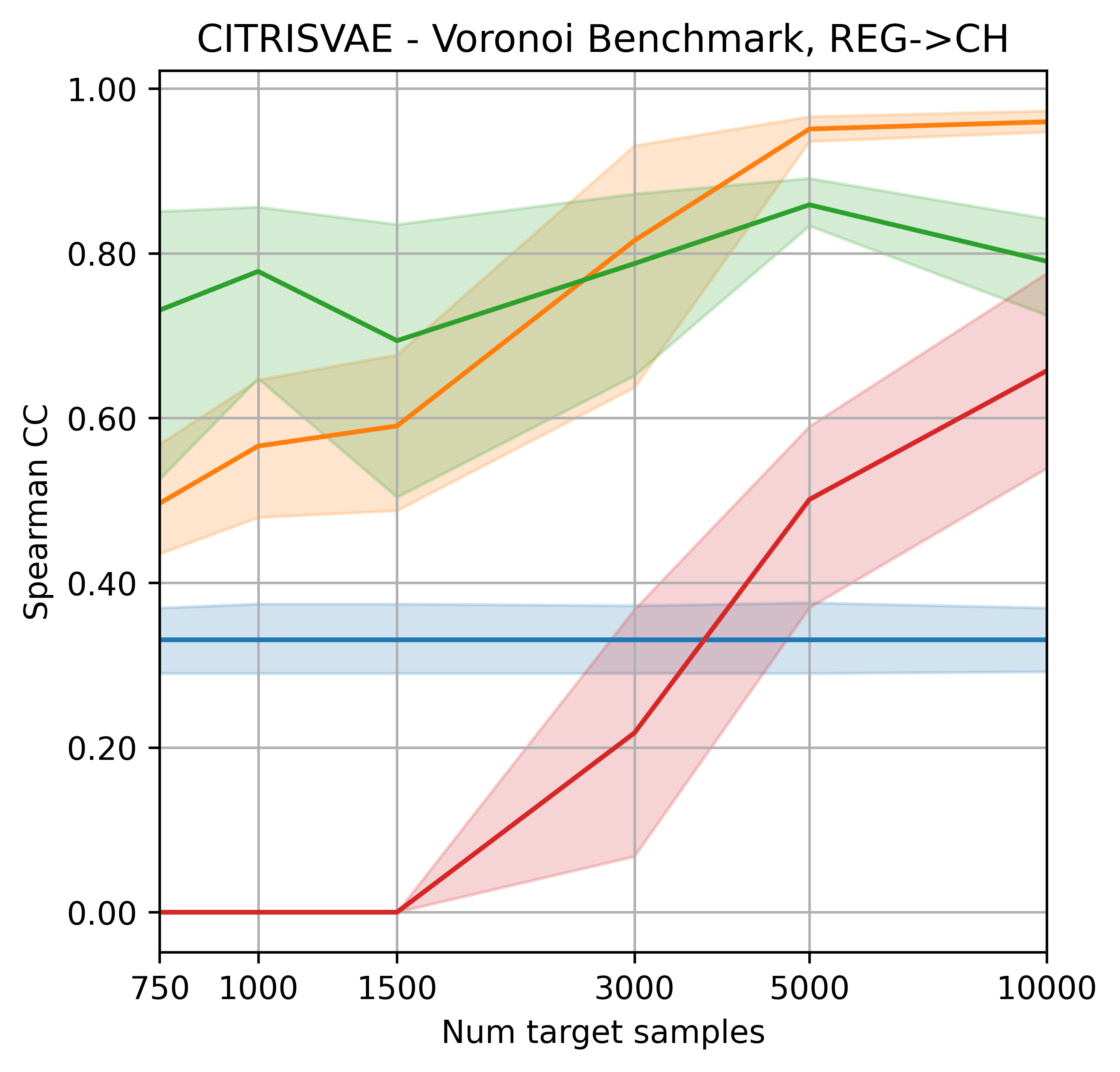}
\includegraphics[width=0.3\textwidth]{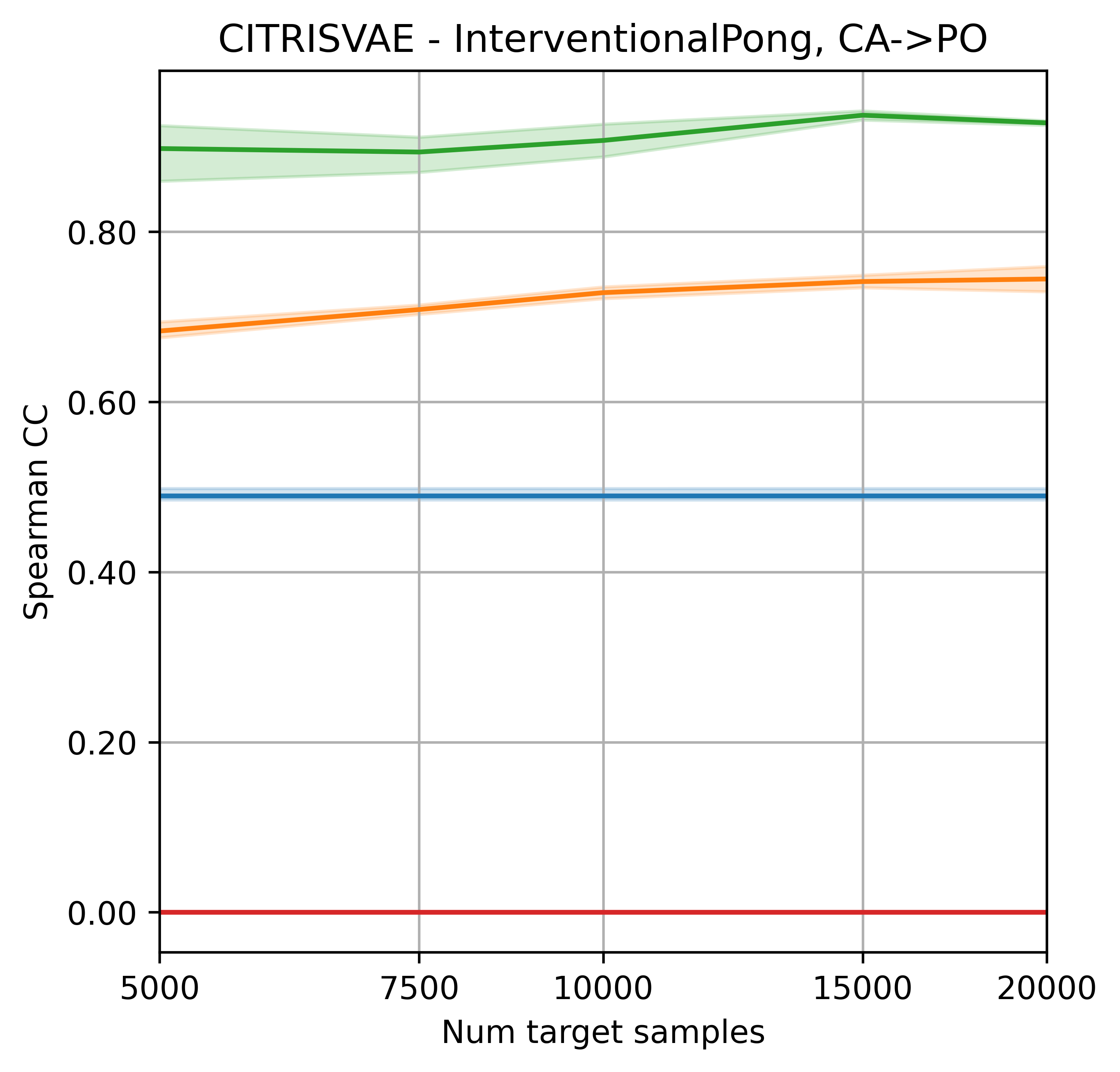}
\includegraphics[width=0.3\textwidth]{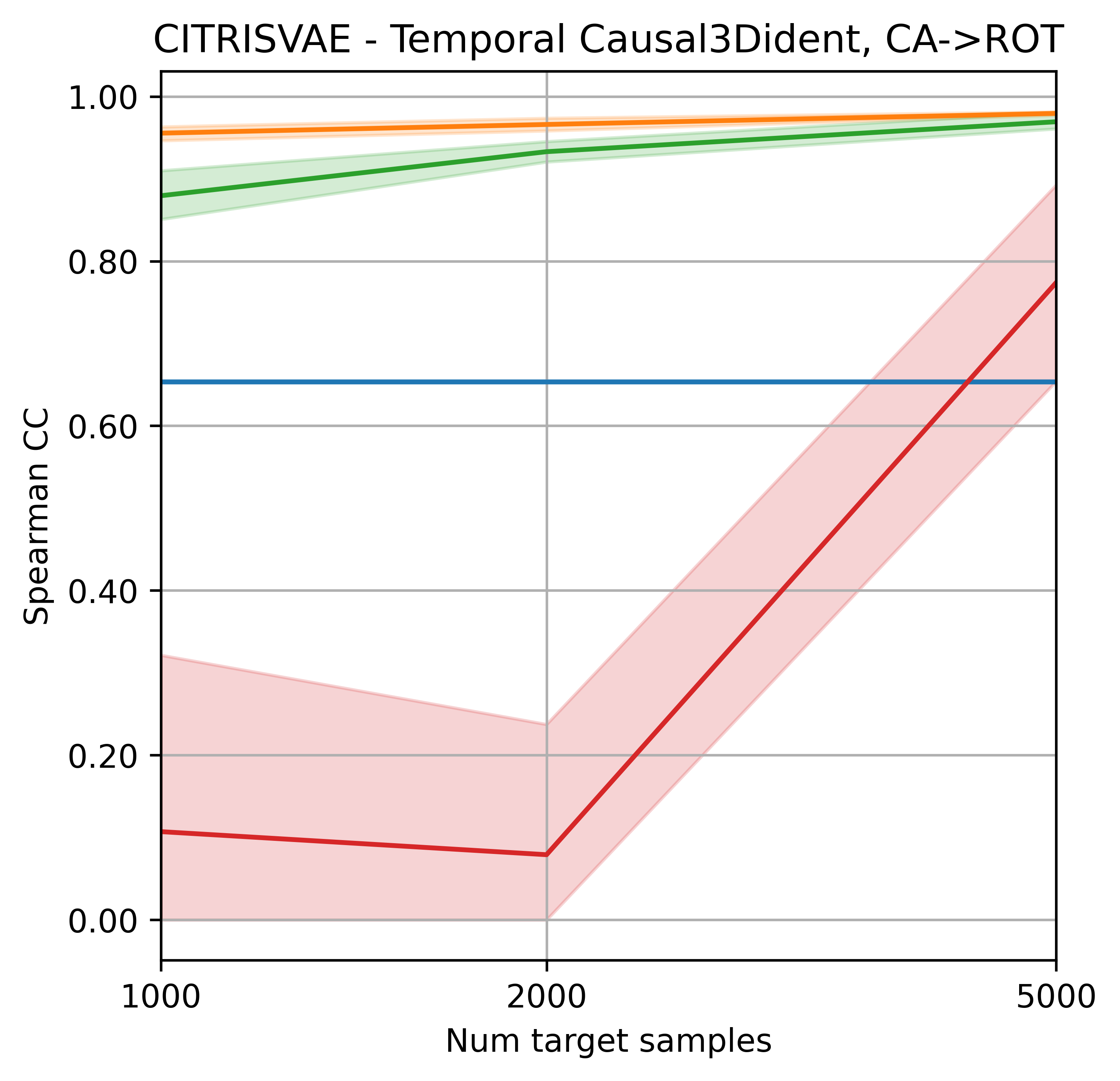}

\smallskip
\includegraphics[width=0.3\textwidth]{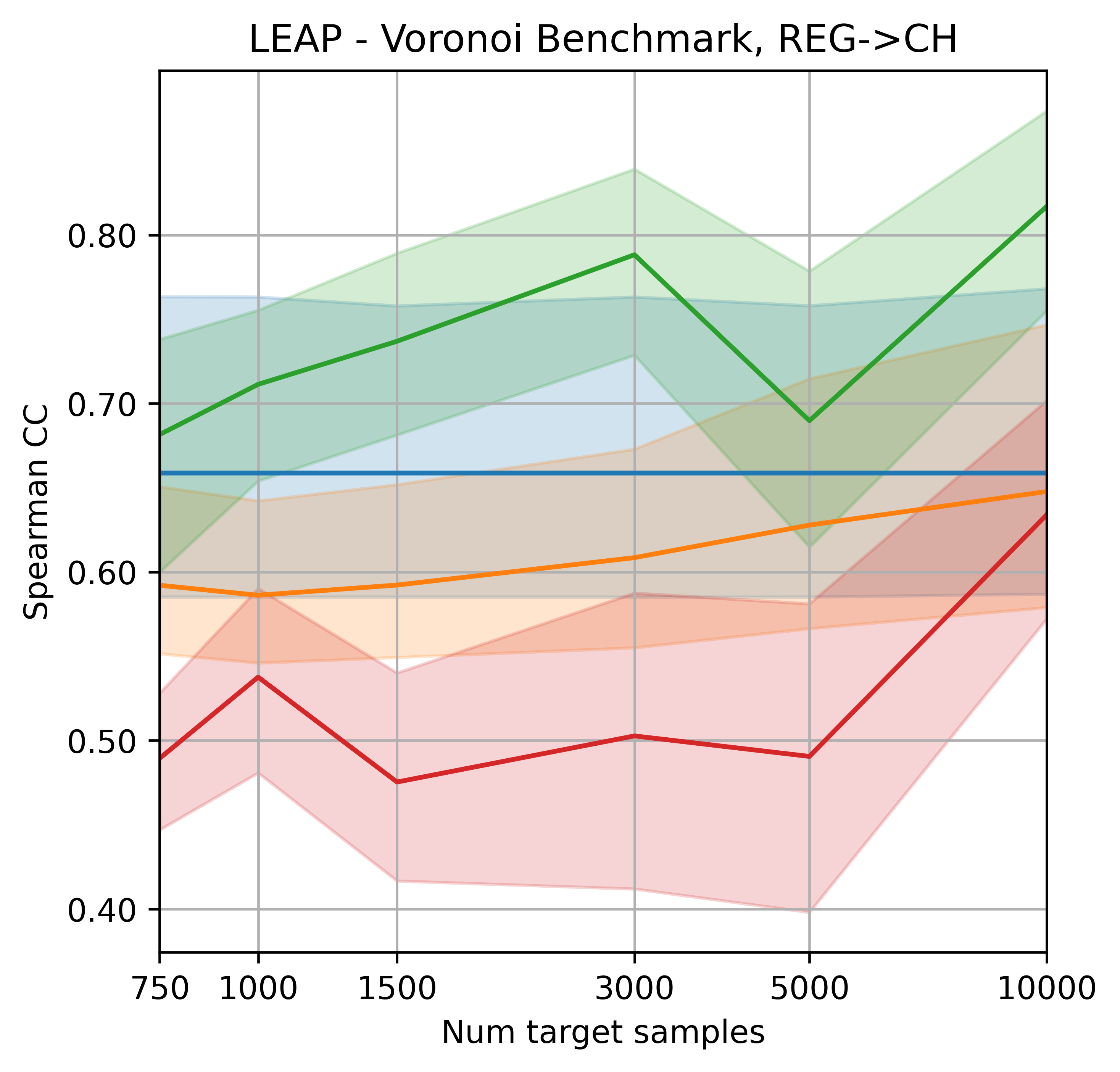}
\includegraphics[width=0.3\textwidth]{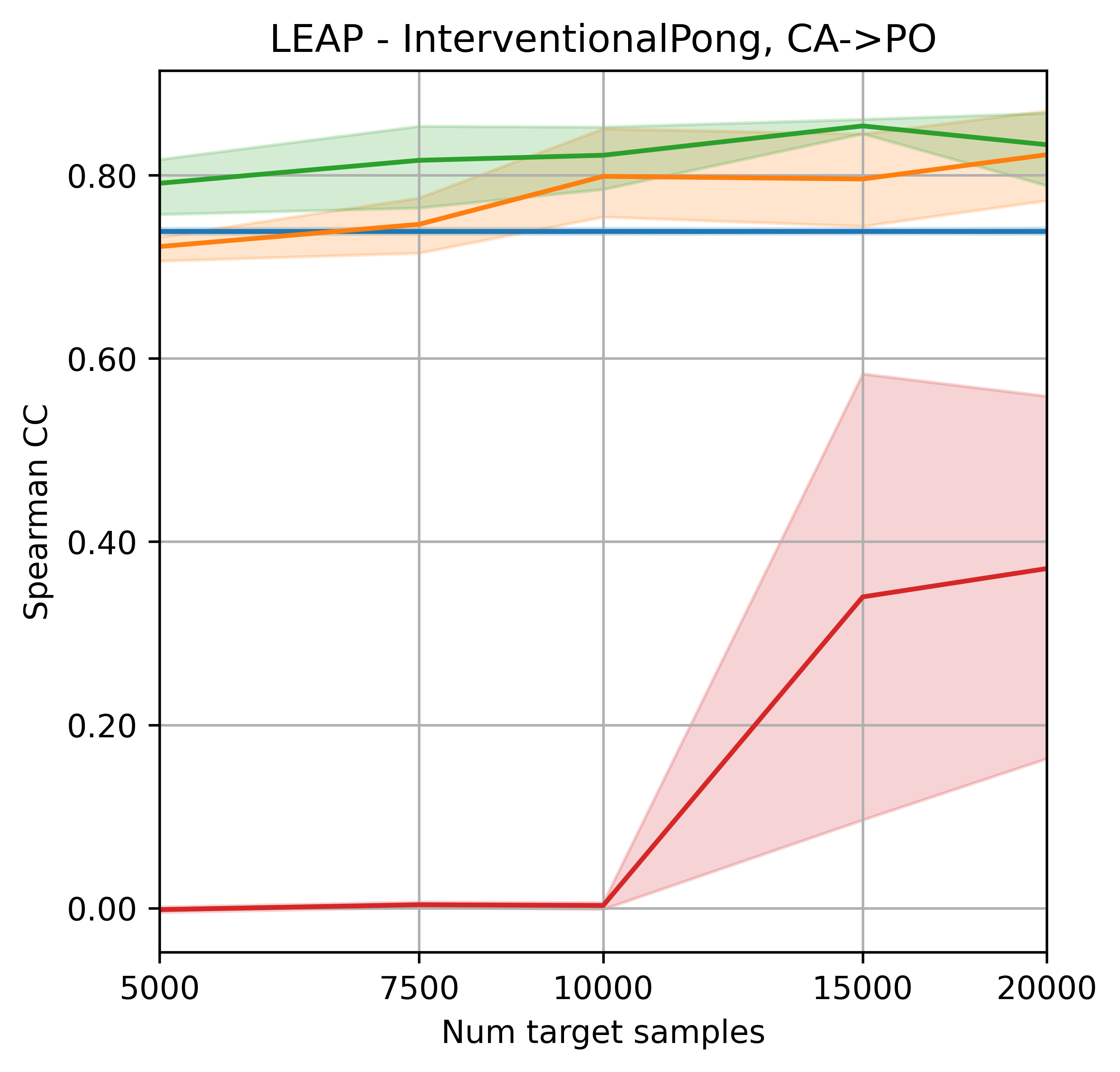}
\includegraphics[width=0.3\textwidth]{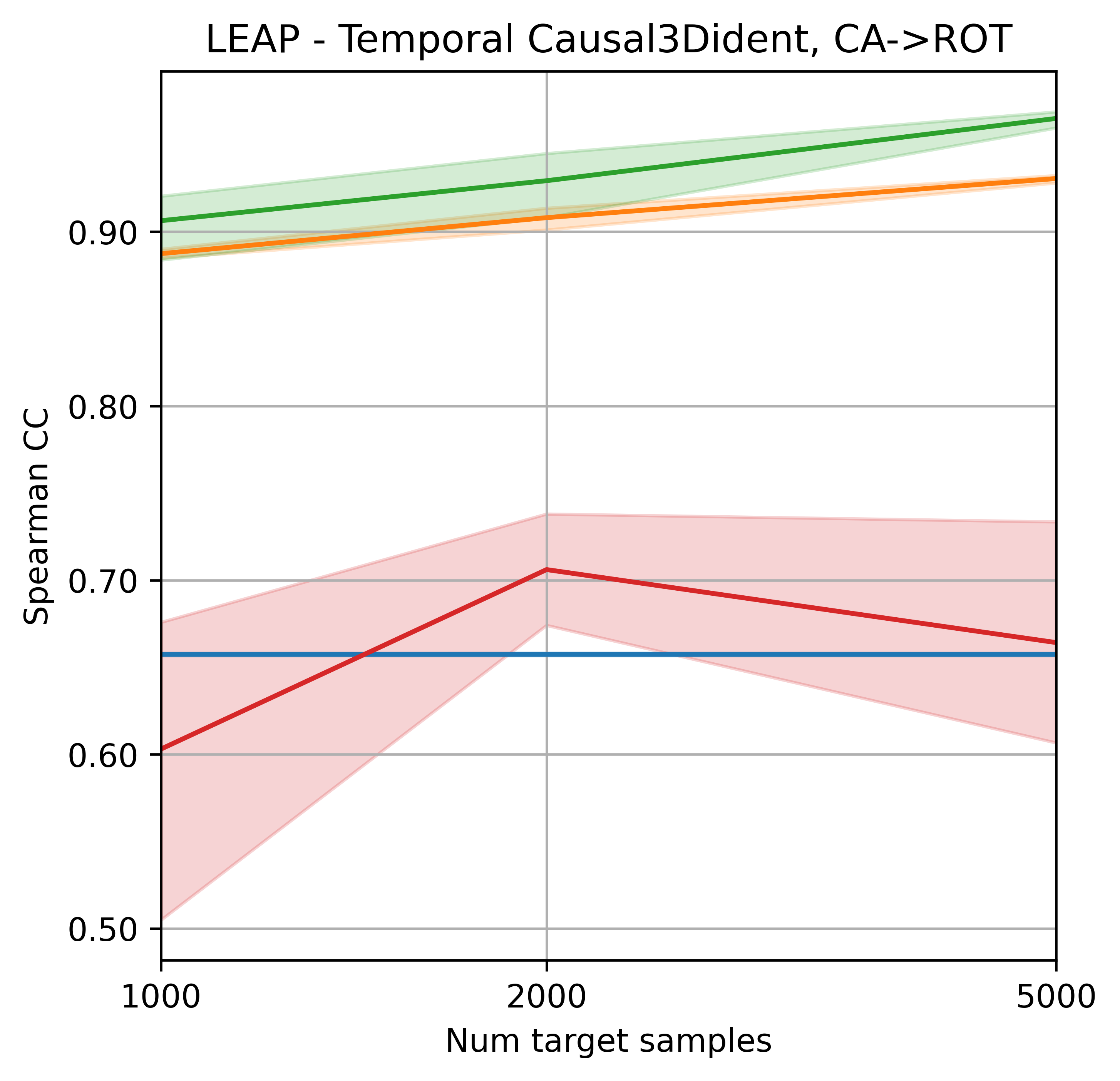}

\smallskip
\includegraphics[width=0.3\textwidth]{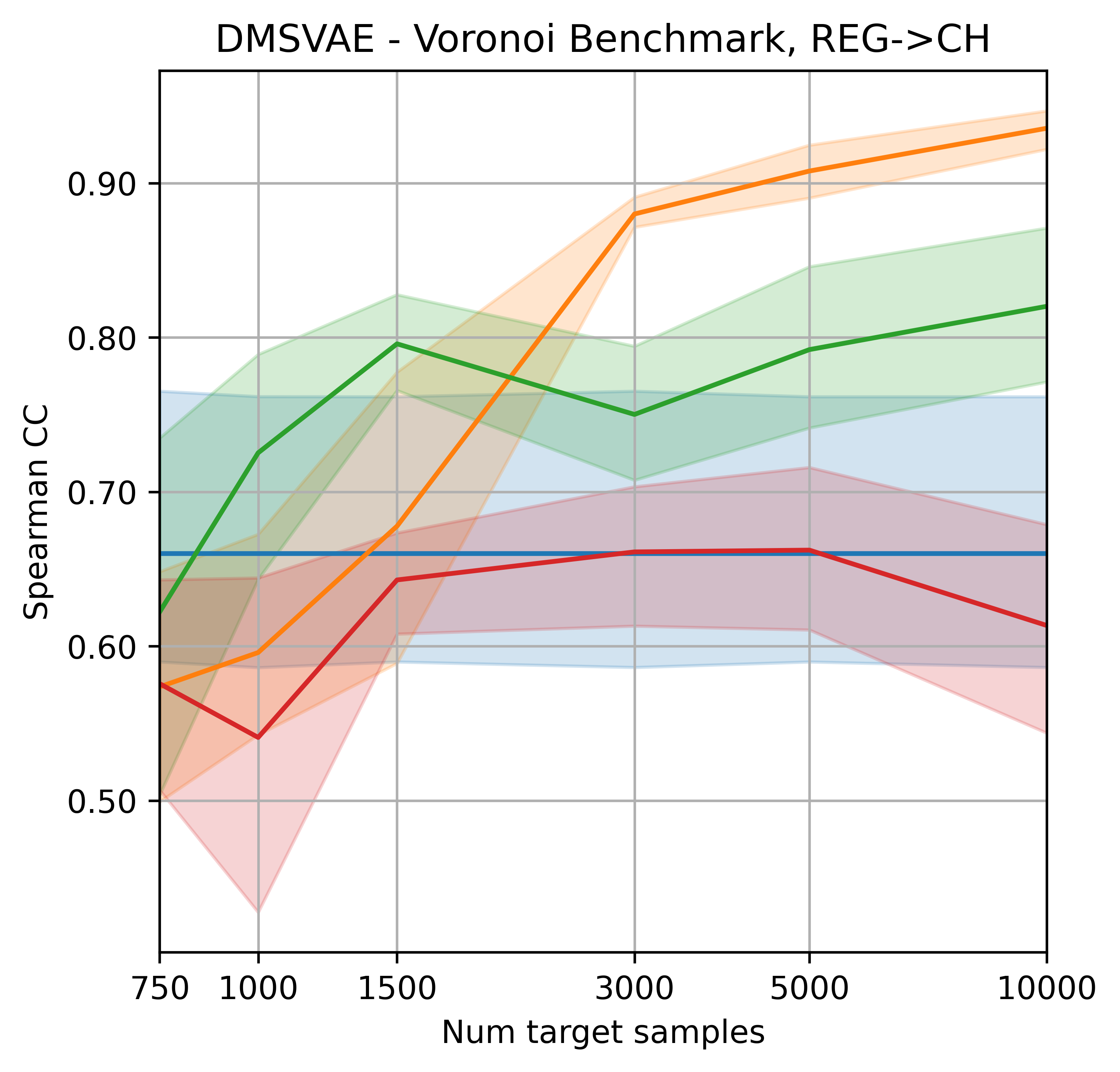}
\includegraphics[width=0.3\textwidth]{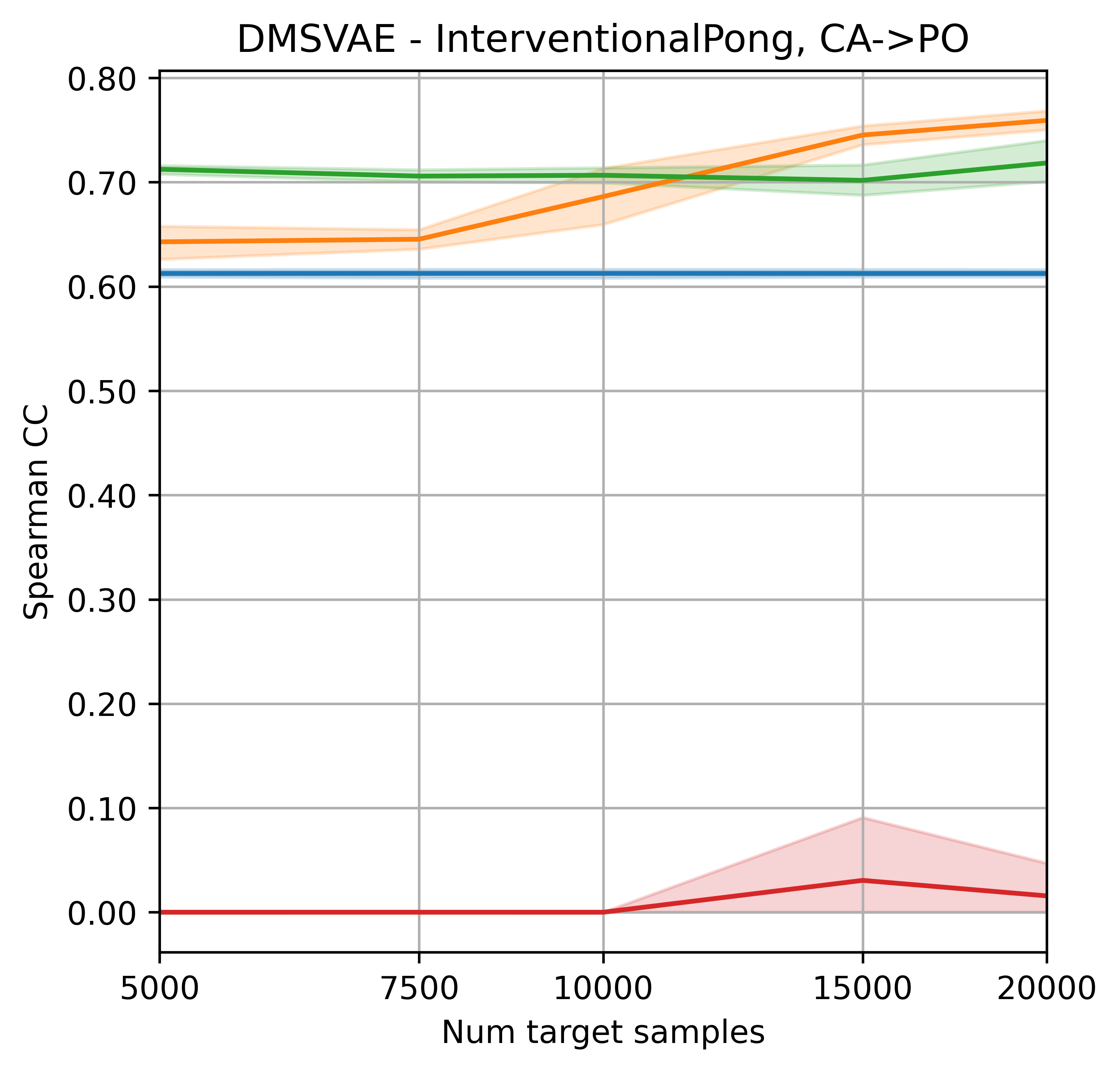}
\includegraphics[width=0.3\textwidth]{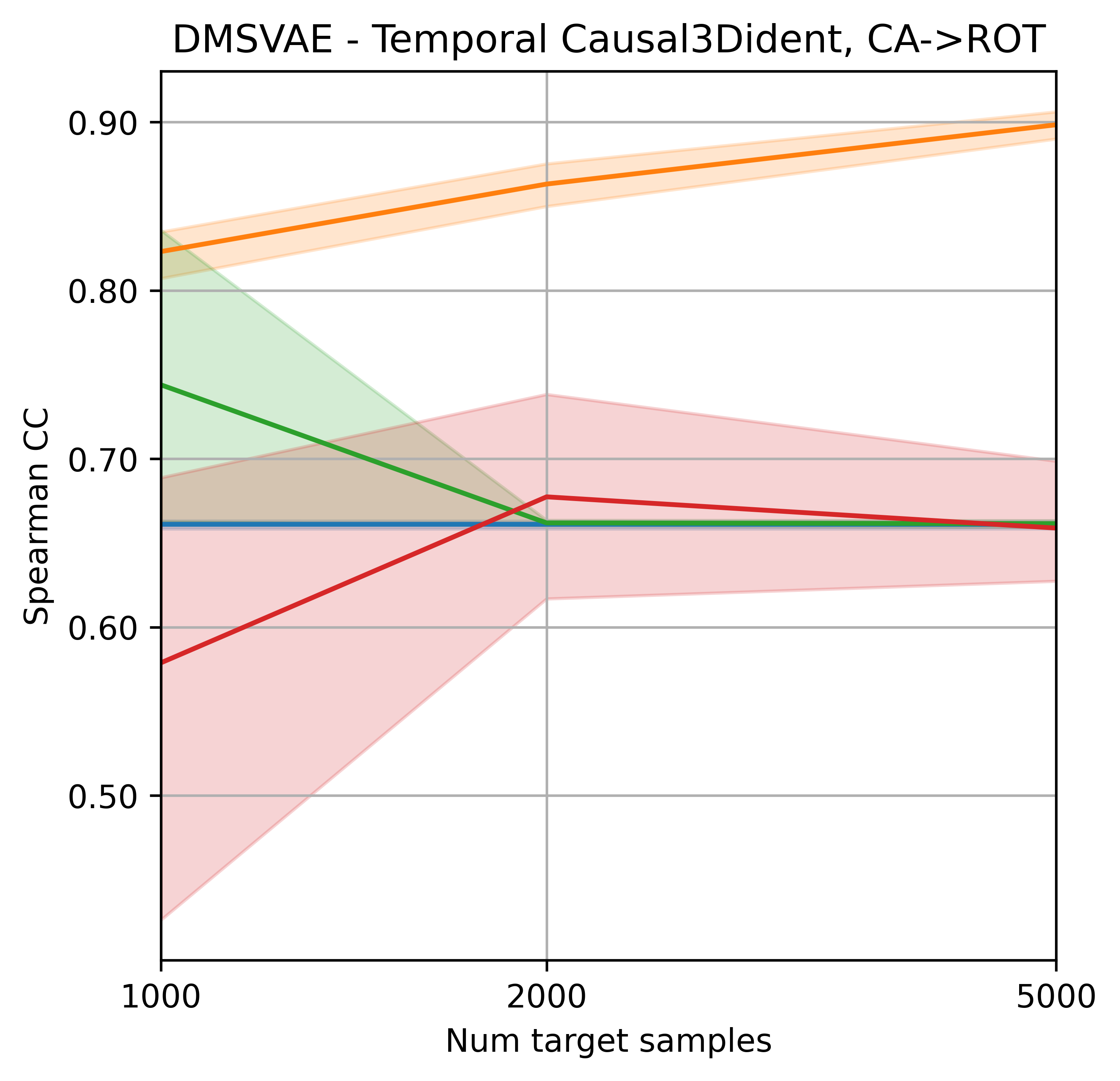}

\smallskip
\includegraphics[width=0.3\textwidth]{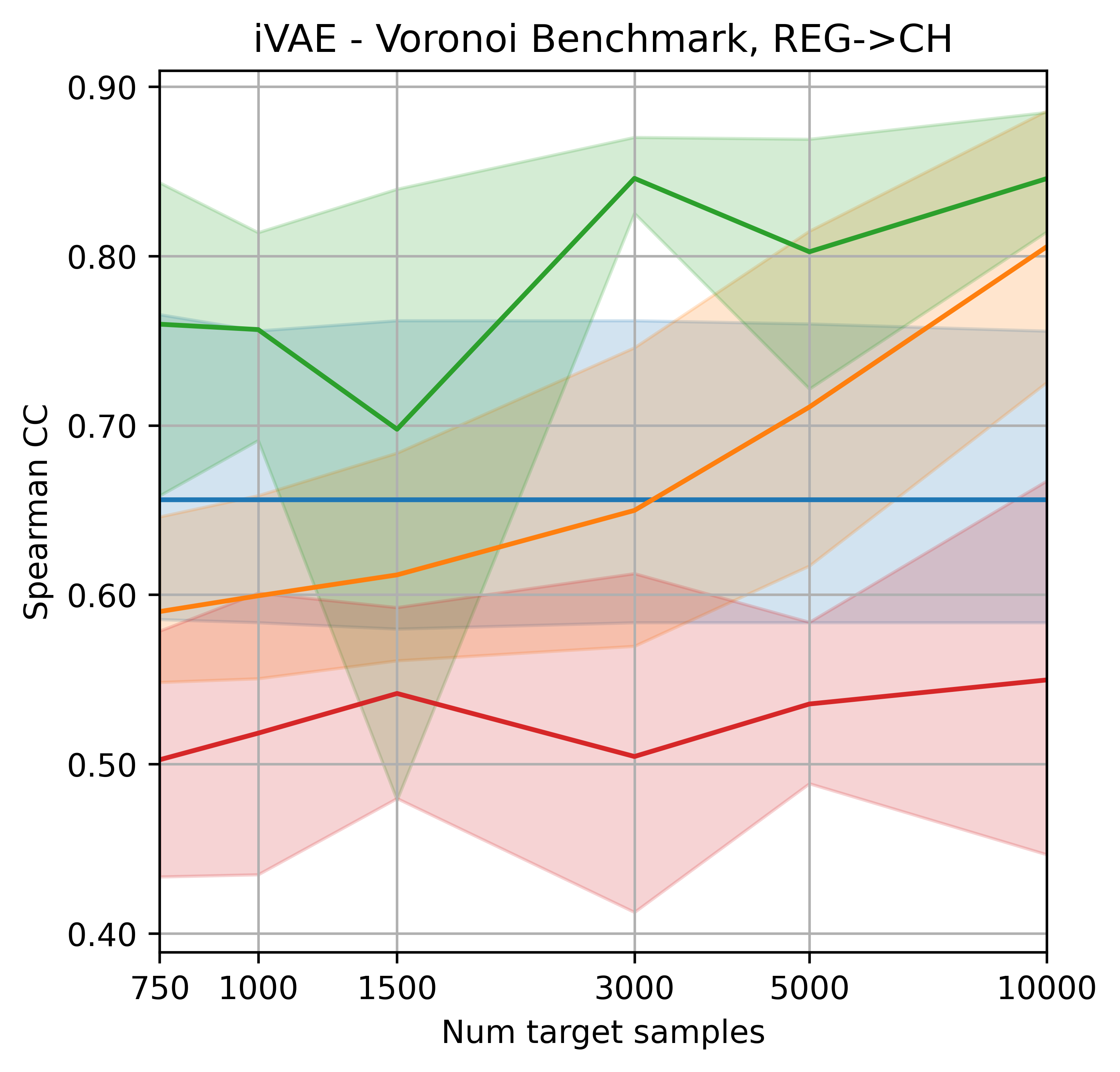}
\includegraphics[width=0.3\textwidth]{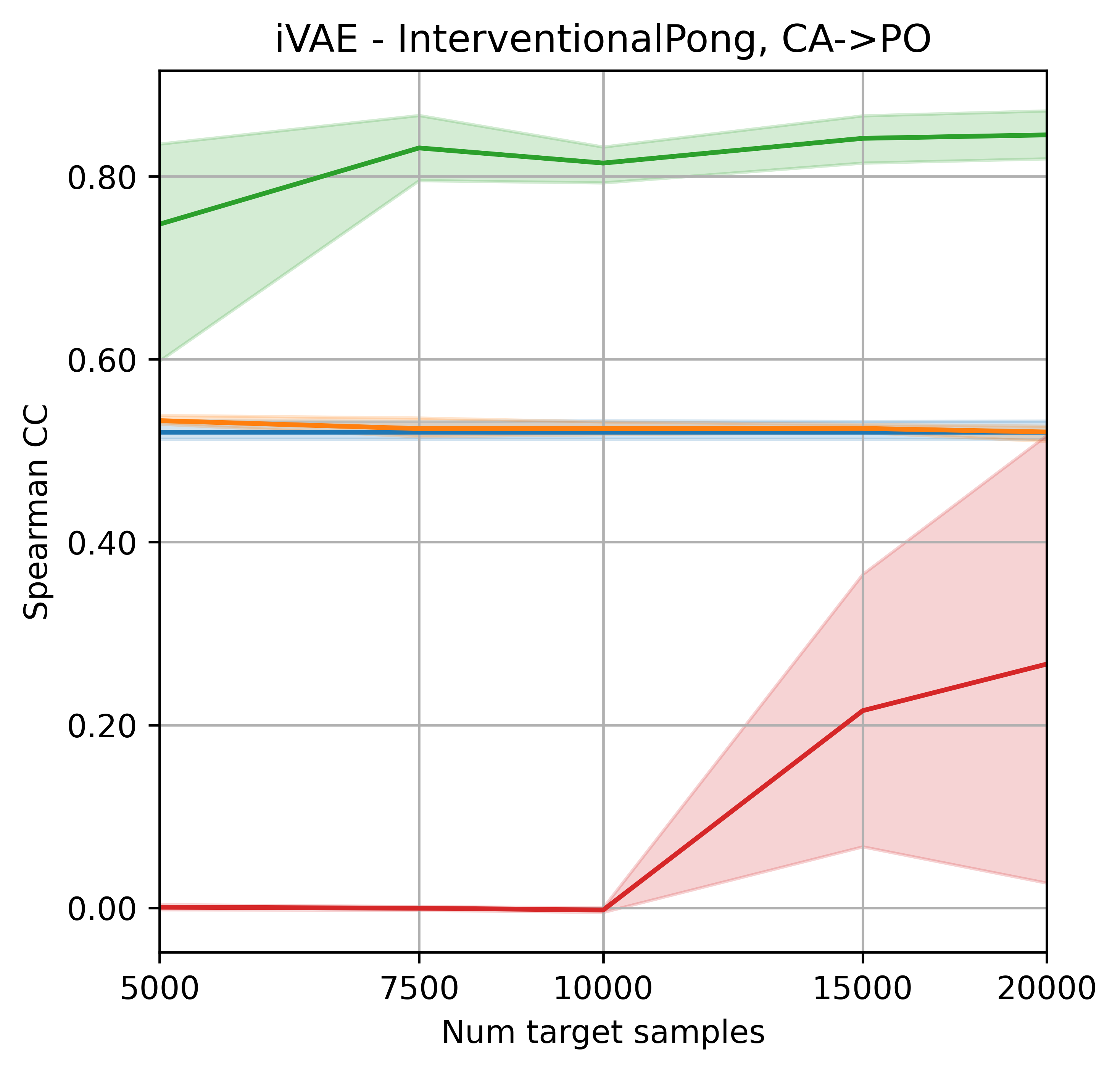}
\includegraphics[width=0.3\textwidth]{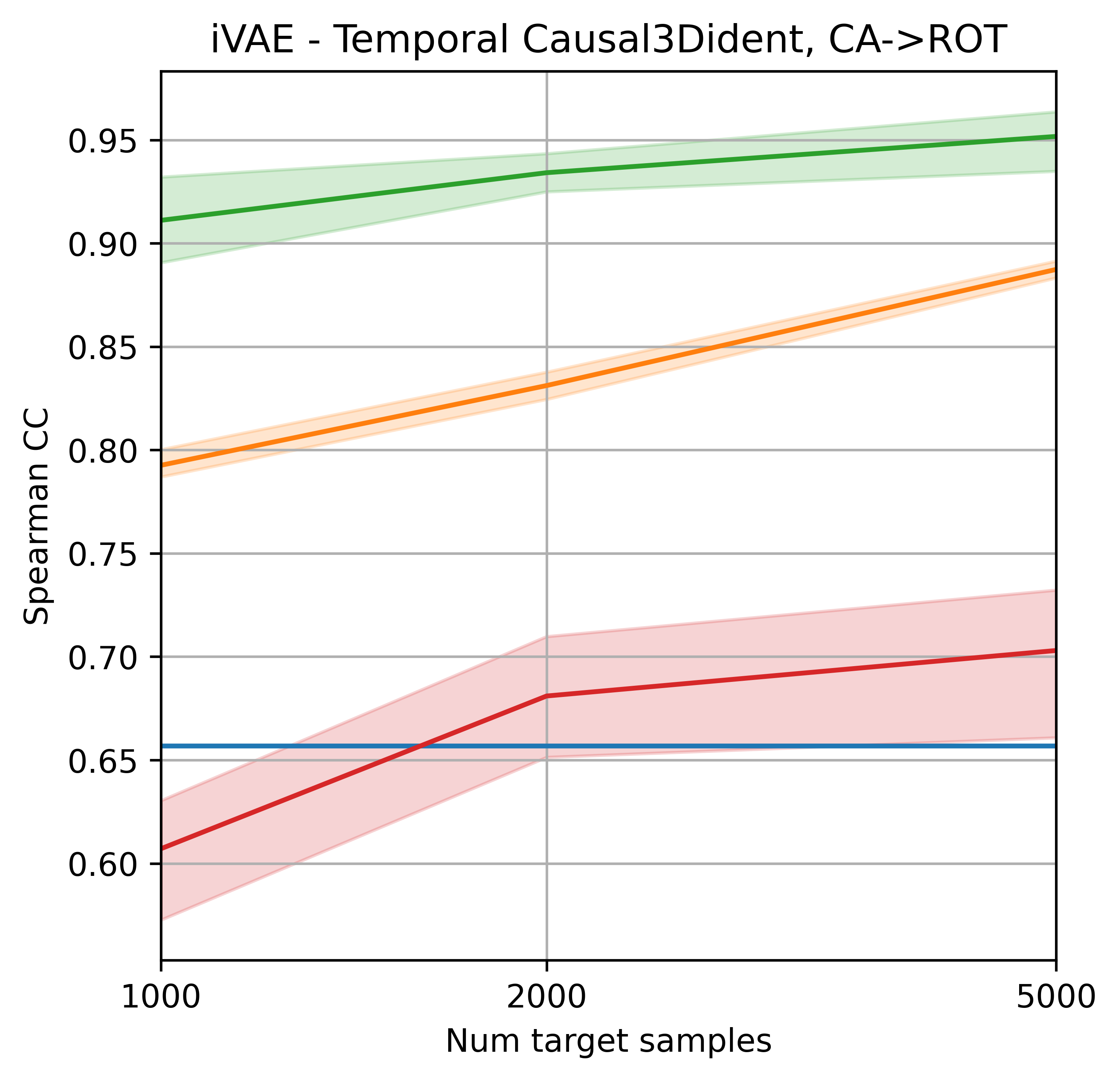}
\centering

\smallskip
\includegraphics[width=0.5\textwidth]{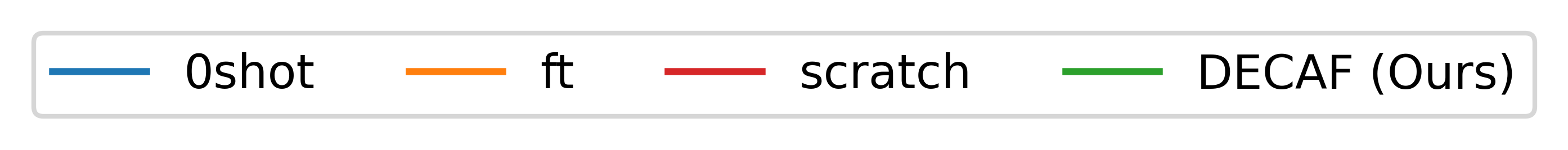}
\caption{Adaptation with increasing number of target samples. \textbf{Rows}: CITRISVAE, LEAP, DMSVAE and iVAE. \textbf{Columns:} Voronoi Benchmark, InterventionalPong, Temporal Causal3dIdent.}
\label{fig:app-results-increasing}
\end{figure}

\begin{figure}
\centering
\includegraphics[width=0.3\textwidth]{clear/results/results-voronoi-comp-CITRISVAE-increasing.png}
\includegraphics[width=0.3\textwidth]{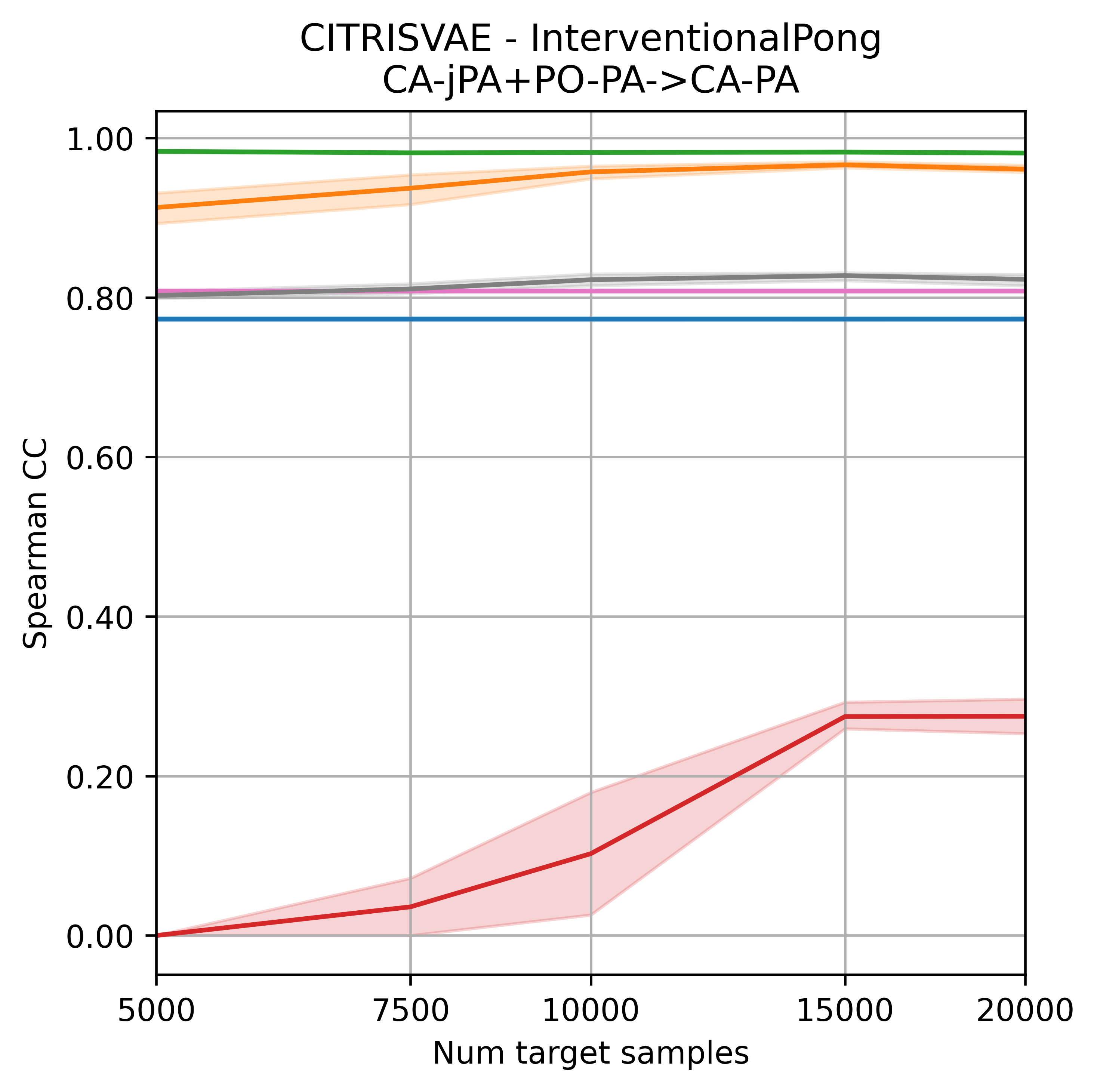}
\includegraphics[width=0.3\textwidth]{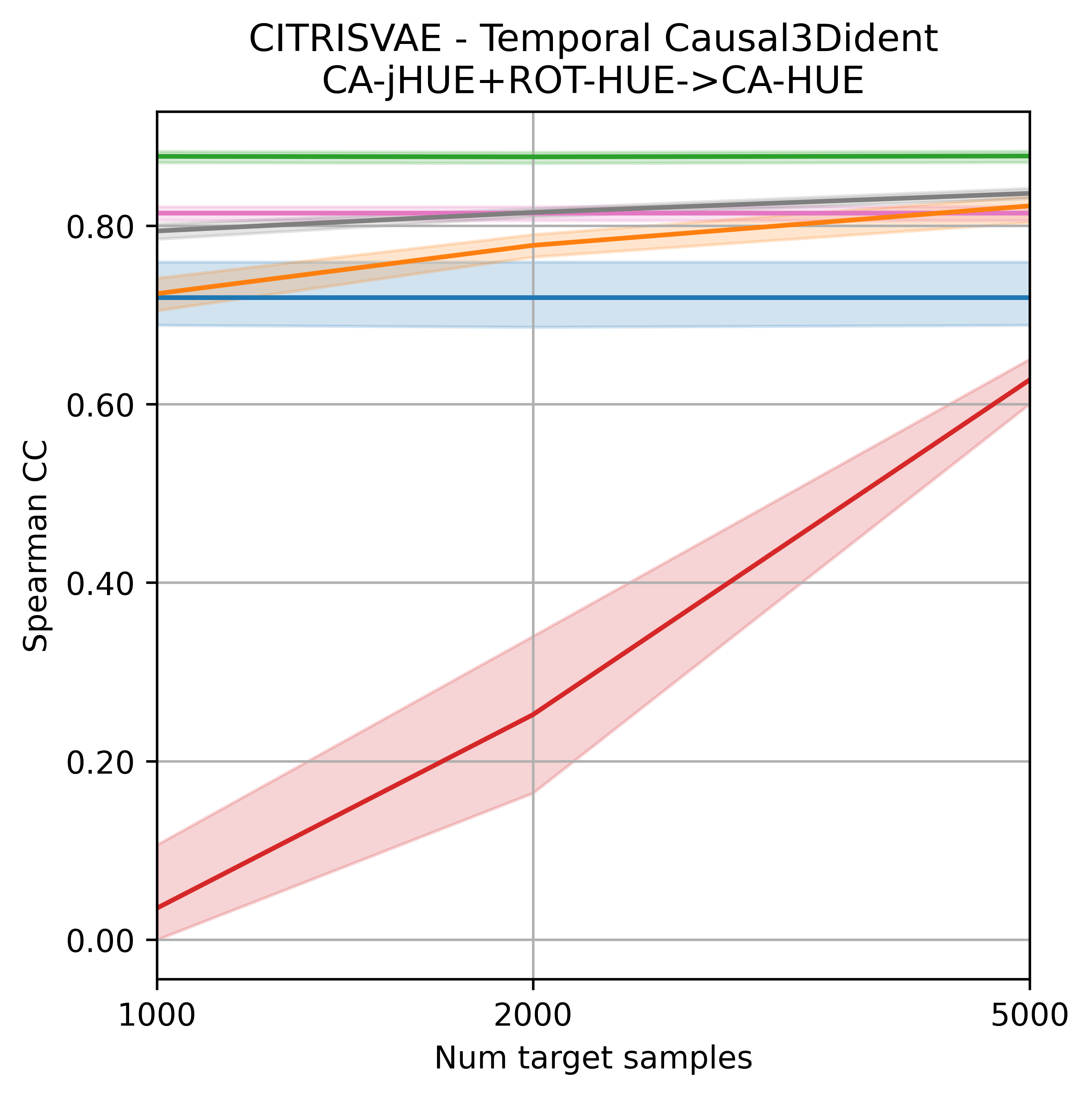}

\smallskip
\includegraphics[width=0.3\textwidth]{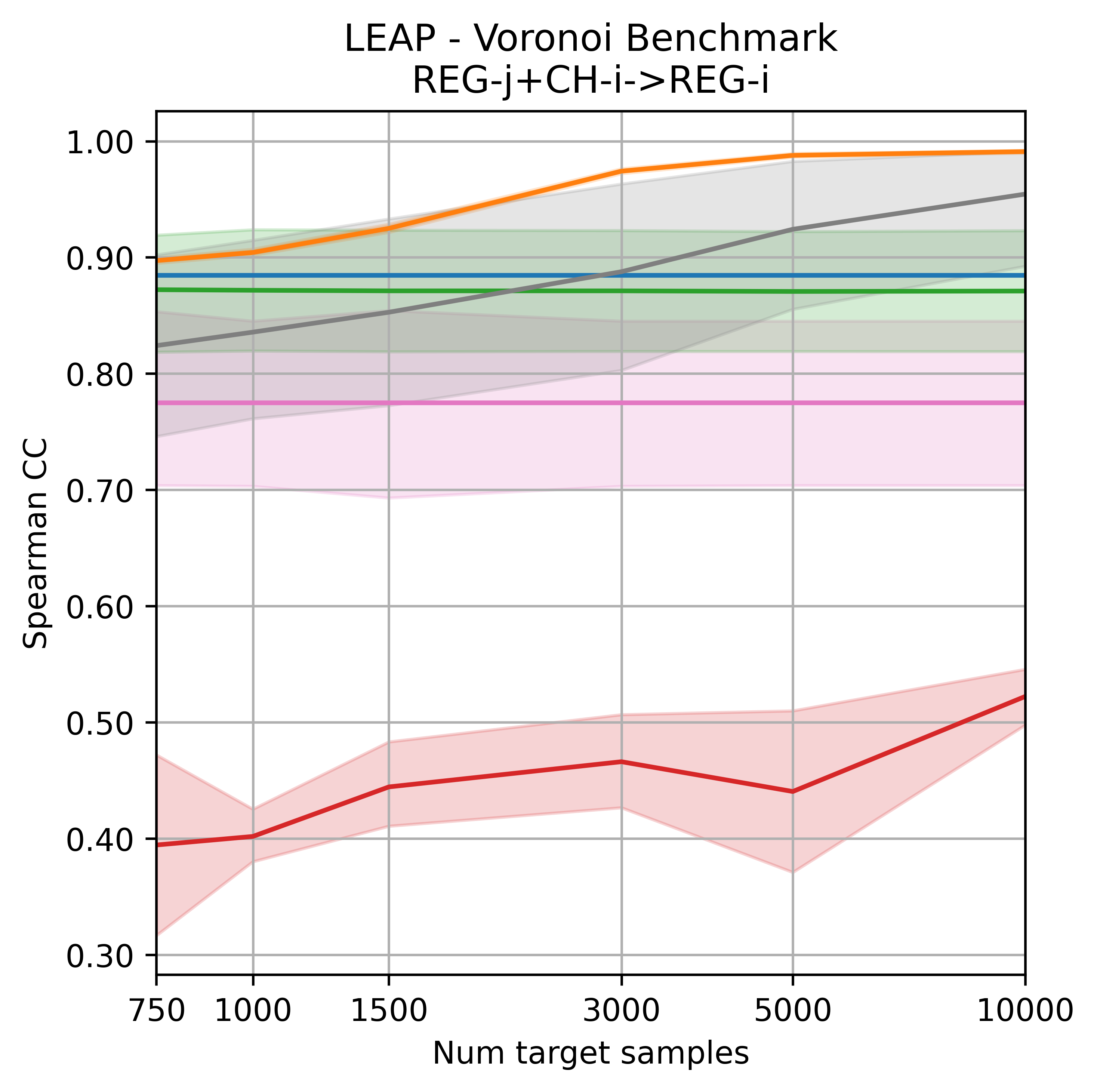}
\includegraphics[width=0.3\textwidth]{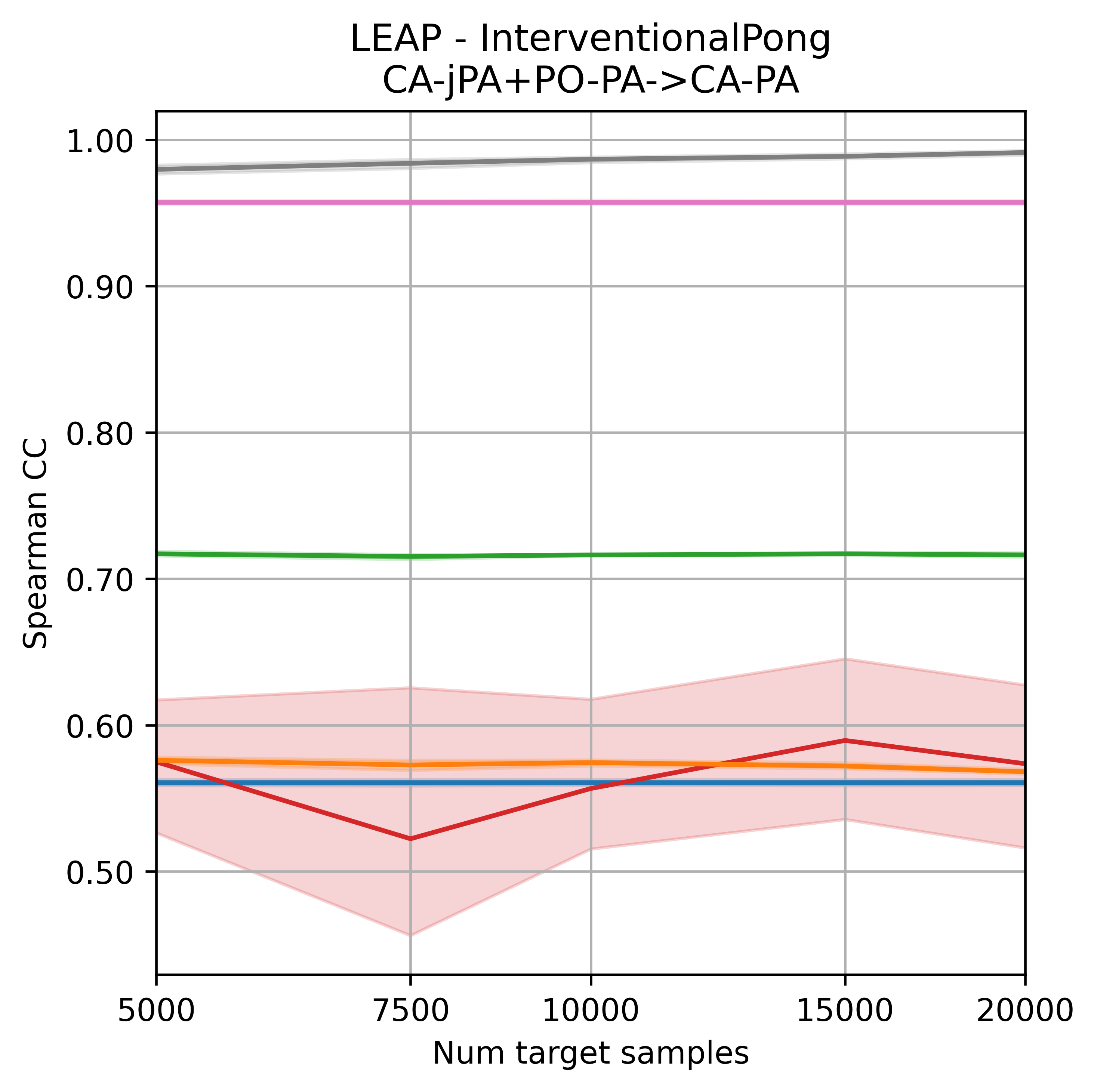}
\includegraphics[width=0.3\textwidth]{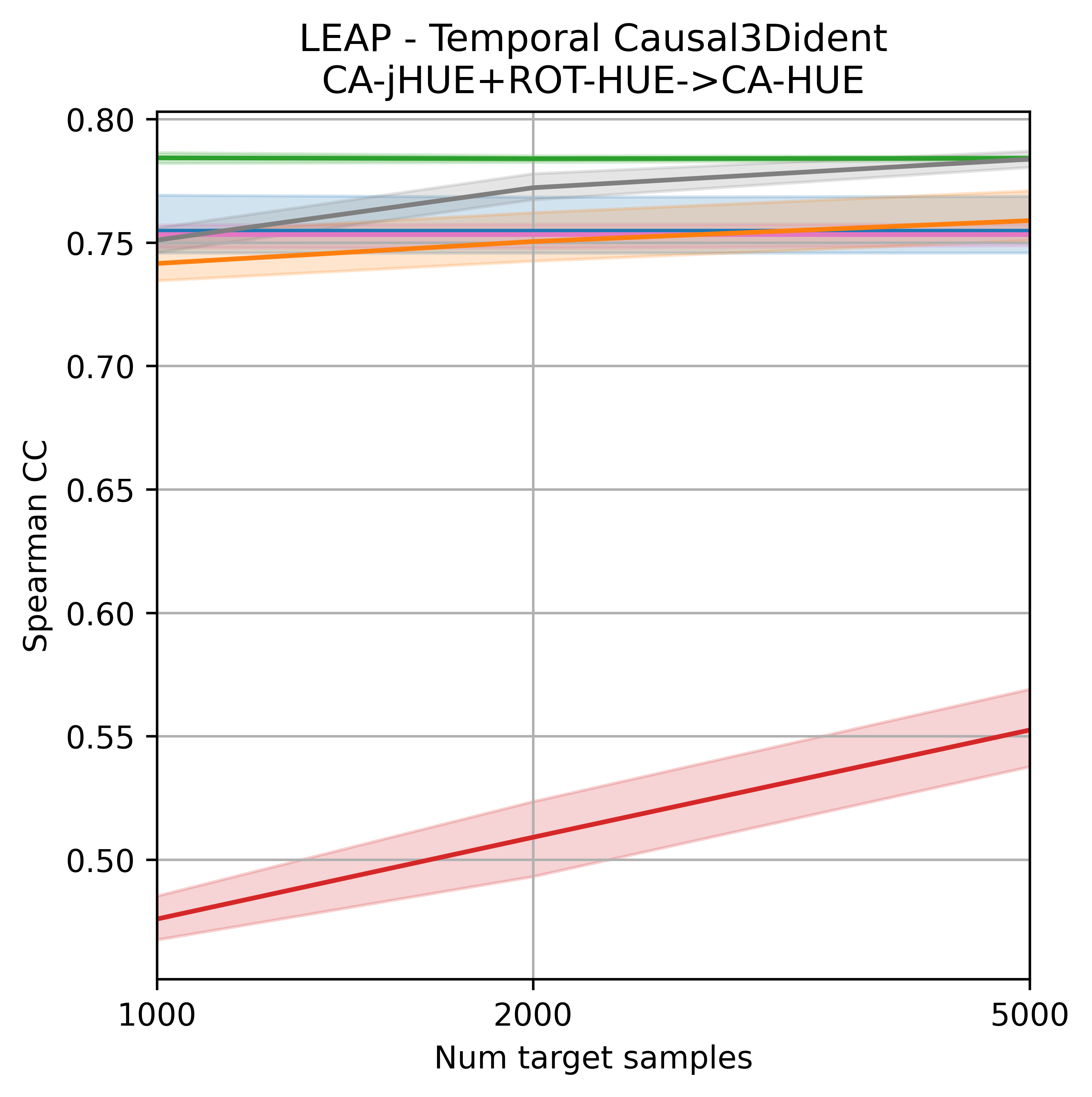}

\smallskip
\includegraphics[width=0.3\textwidth]{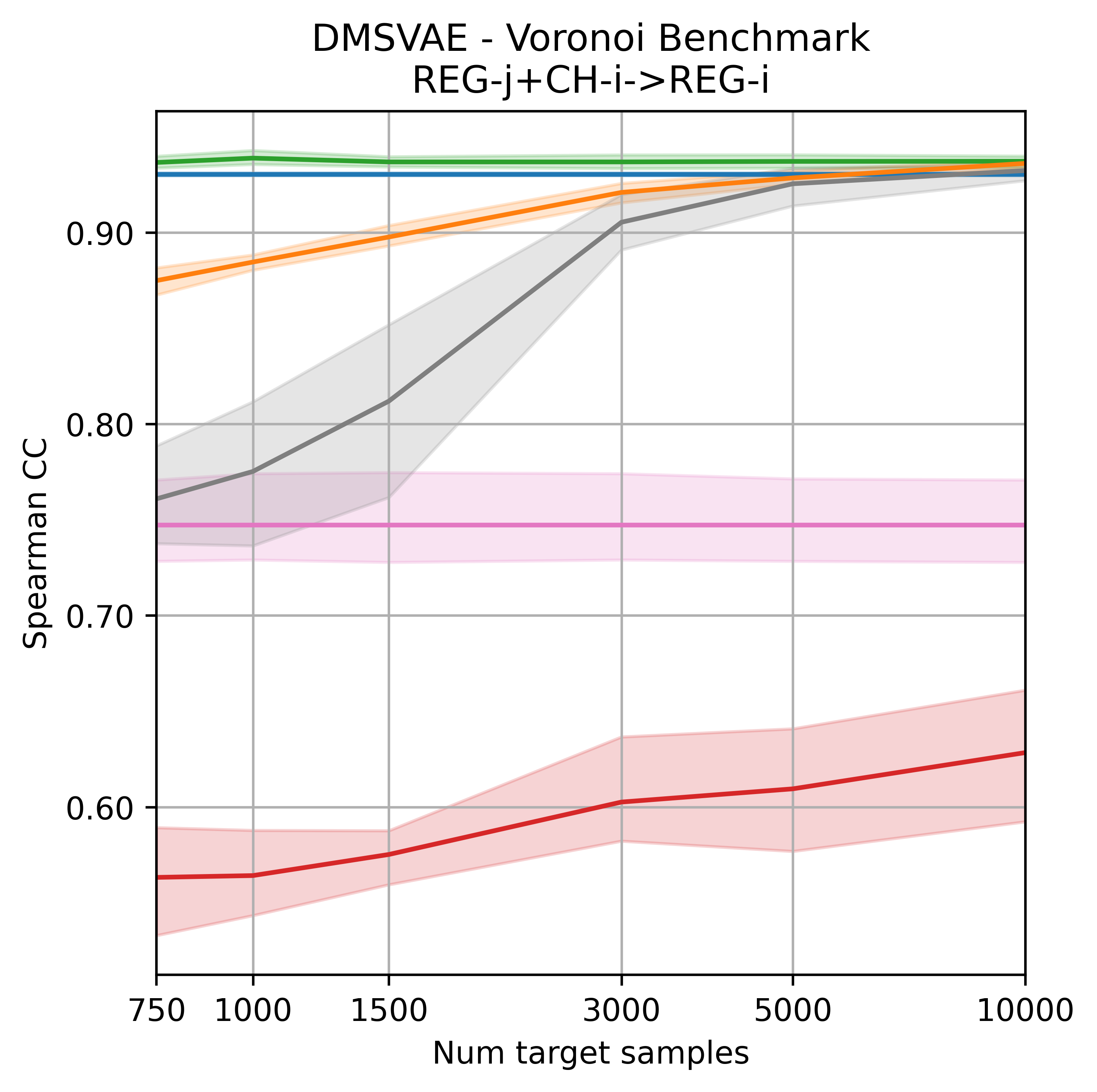}
\includegraphics[width=0.3\textwidth]{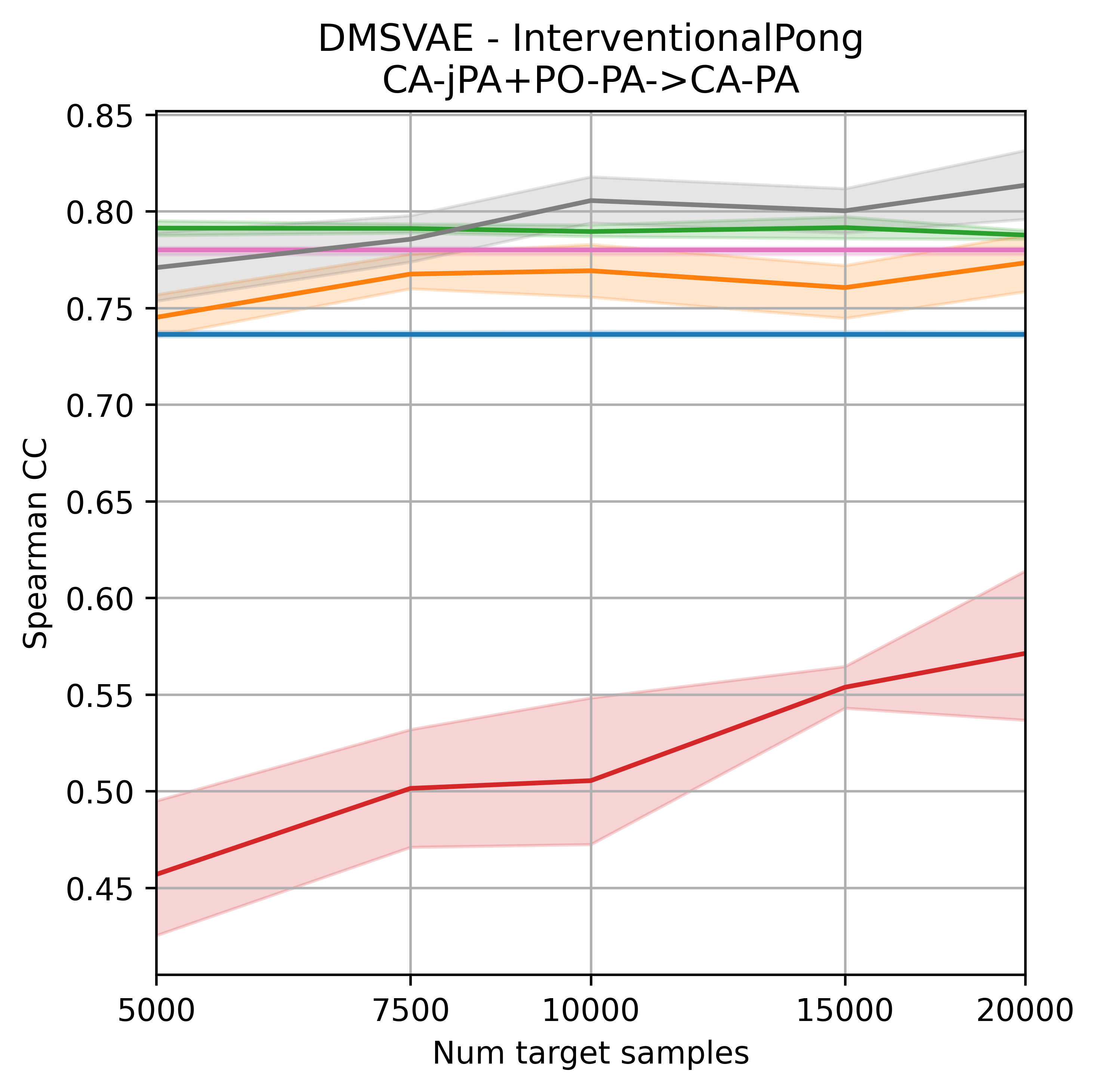}
\includegraphics[width=0.3\textwidth]{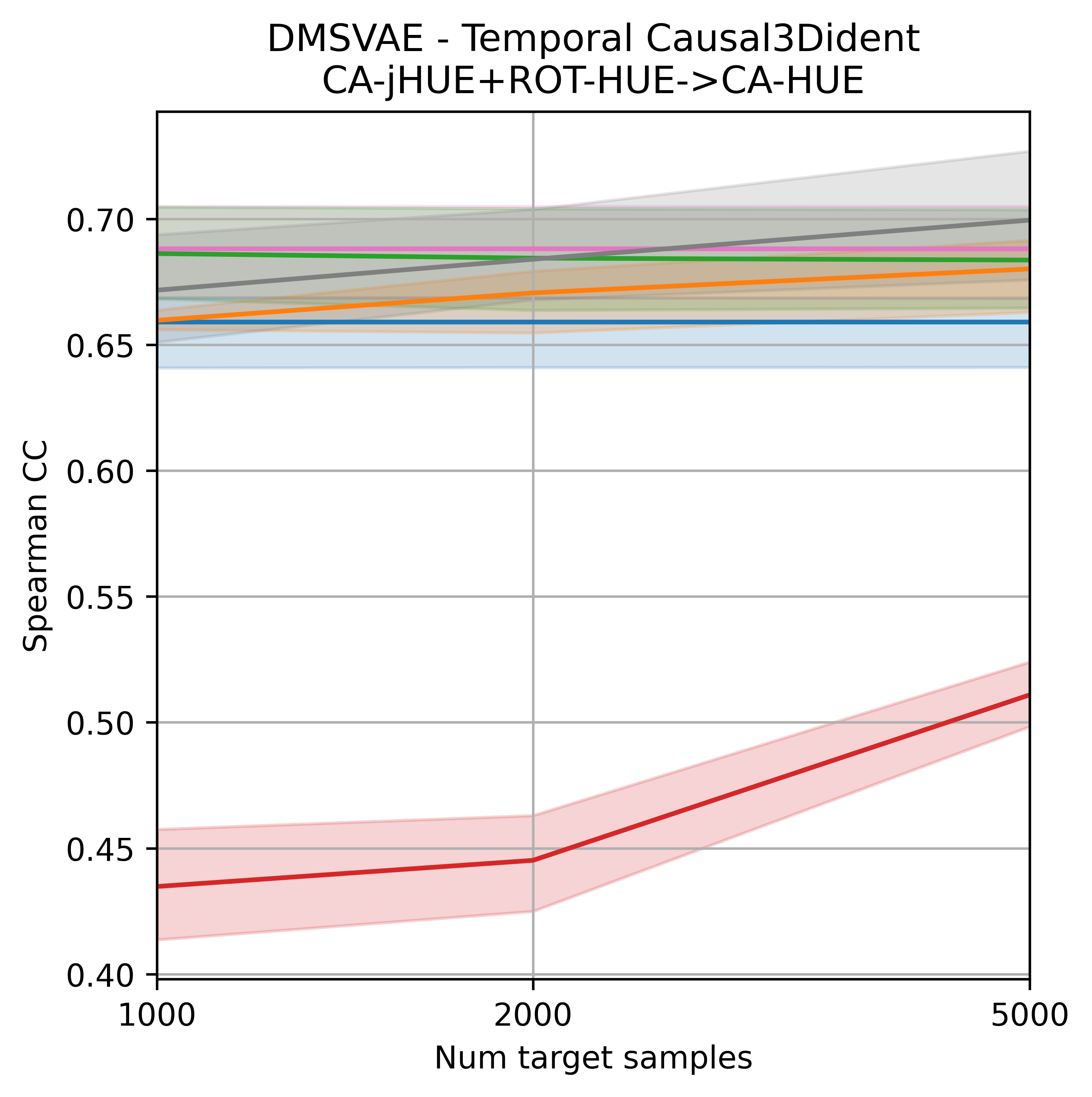}

\smallskip
\includegraphics[width=0.3\textwidth]{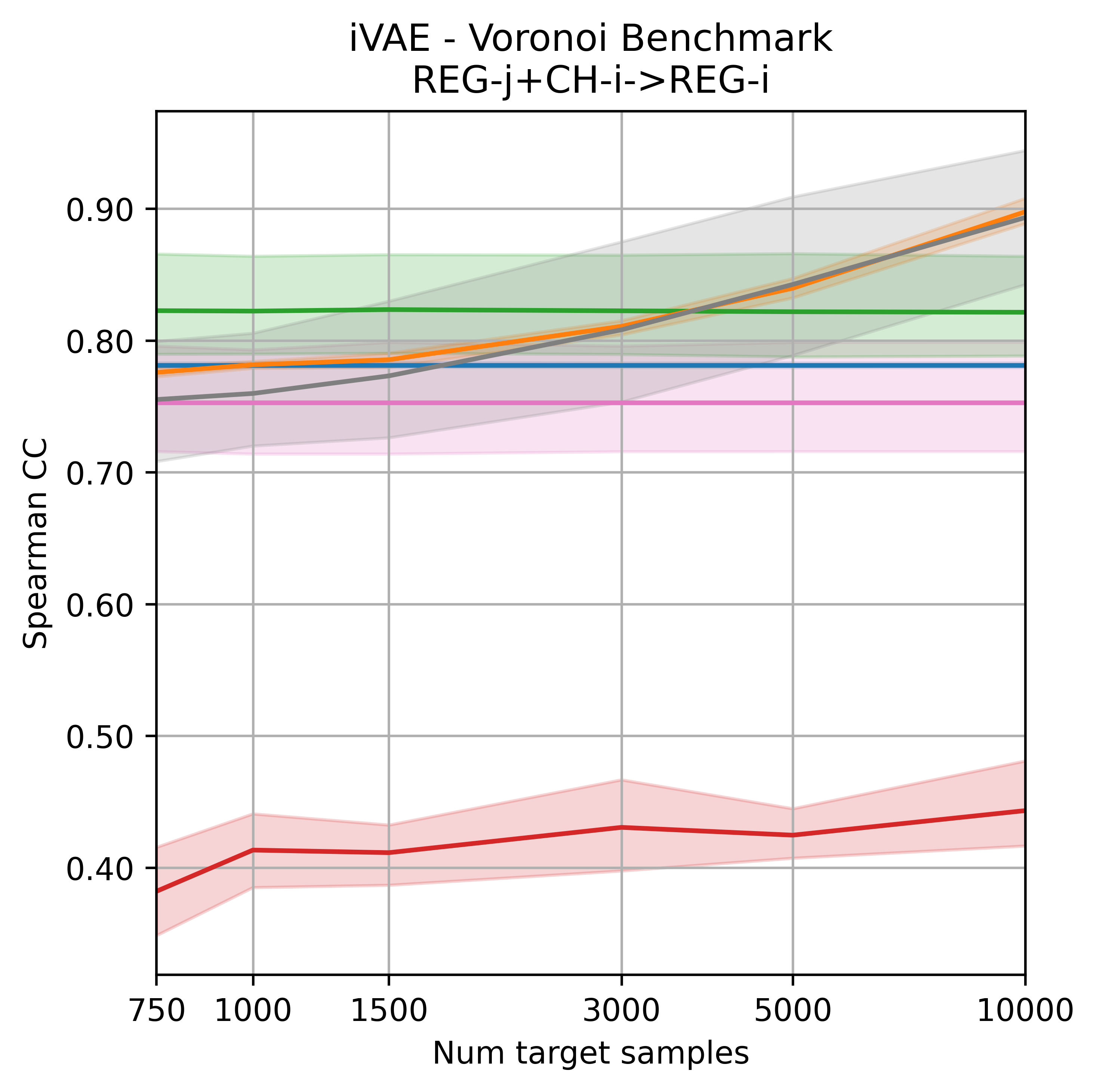}
\includegraphics[width=0.3\textwidth]{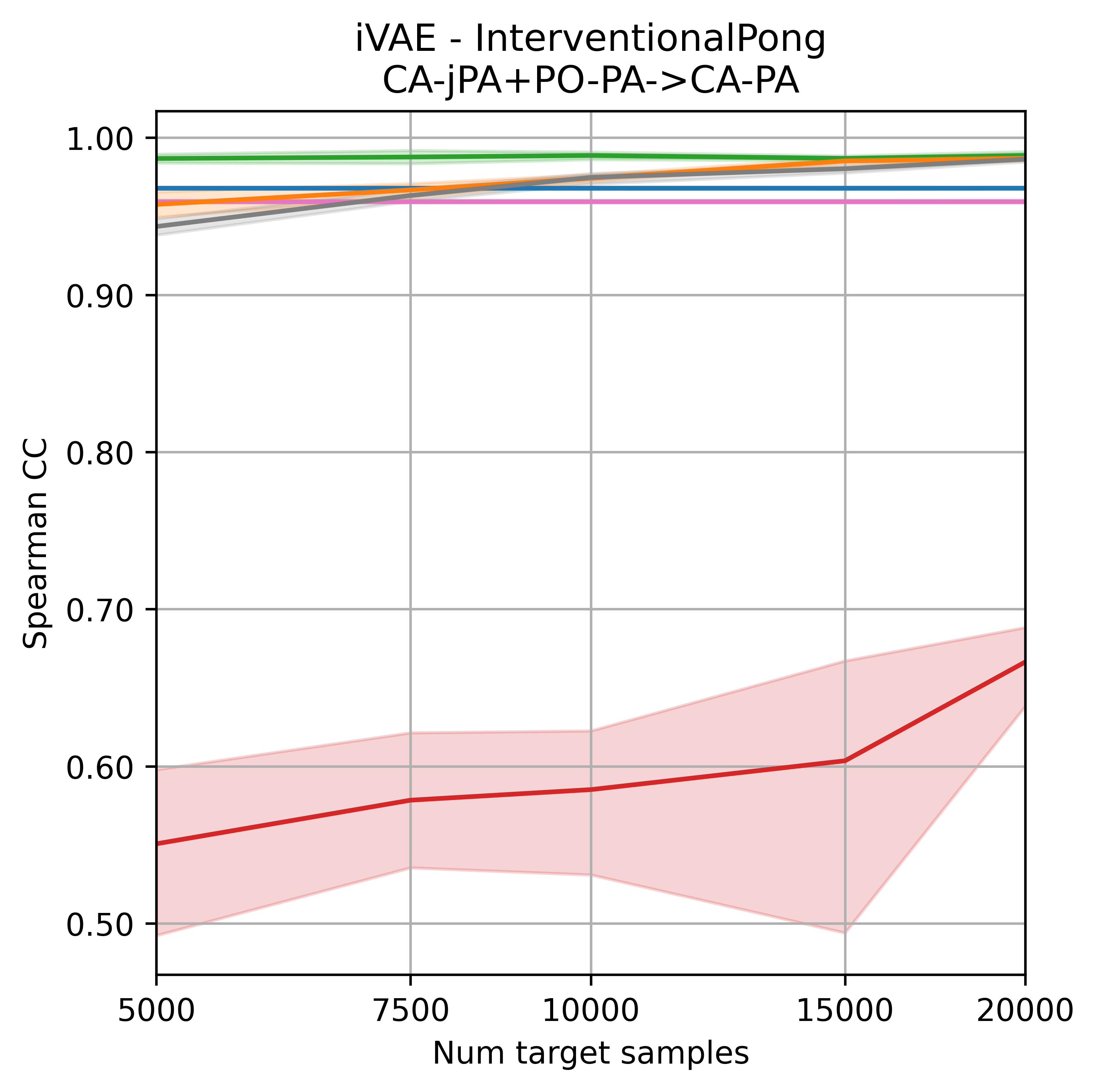}
\includegraphics[width=0.3\textwidth]{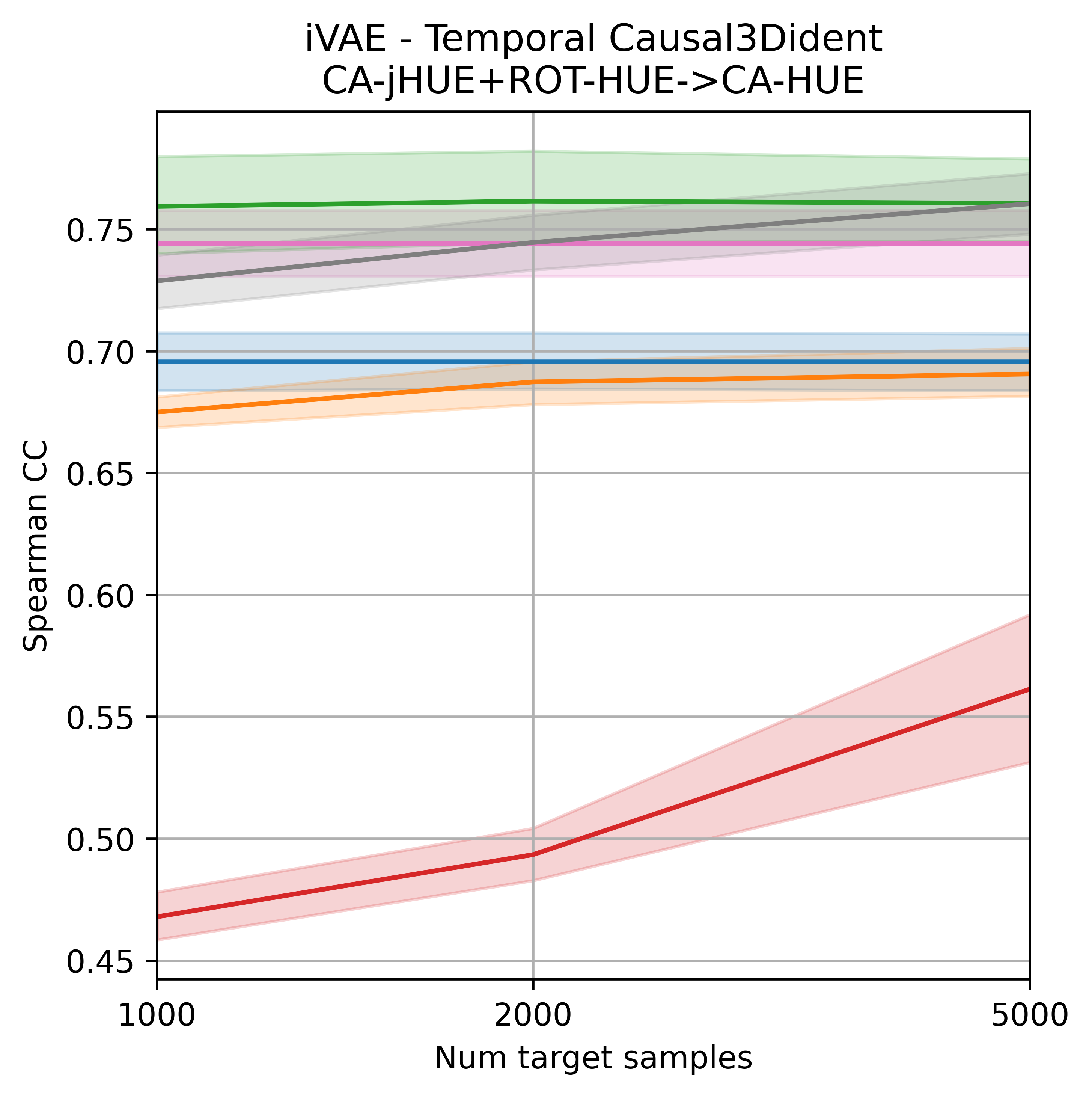}

\centering
\smallskip
\includegraphics[width=0.8\textwidth]{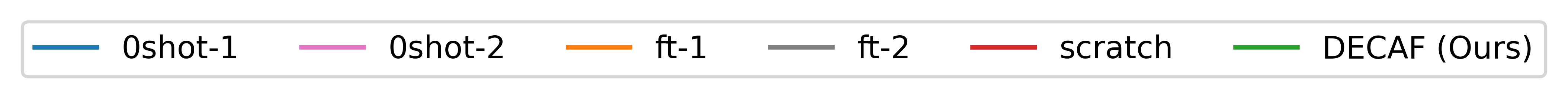}

\caption{Composition with increasing number of target samples. \textbf{Rows}: CITRISVAE, LEAP, DMSVAE and iVAE. \textbf{Columns:} Voronoi Benchmark, InterventionalPong, Temporal Causal3dIdent.}
\label{fig:app-results-comp-increasing}
\end{figure}

In Figure~\ref{fig:app-results-increasing} we report results on the adaptation of causal representations when increasing the number of target samples. Similarly, Figure~\ref{fig:app-results-comp-increasing} reports the composition results when increasing the number of target samples.

\subsection{Detection of changed factors}
\begin{figure}
\centering
\includegraphics[width=0.3\textwidth]{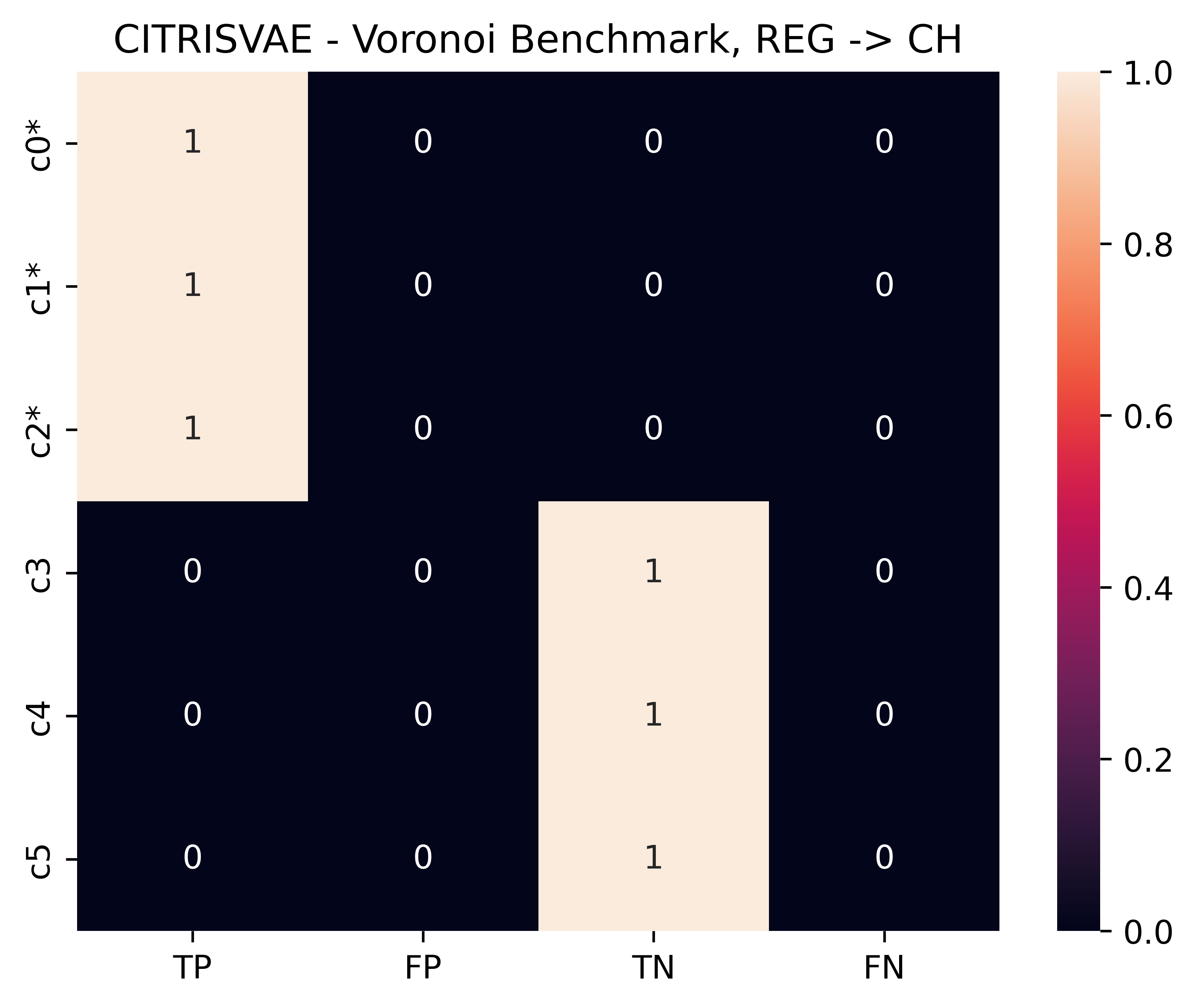}
\includegraphics[width=0.3\textwidth]{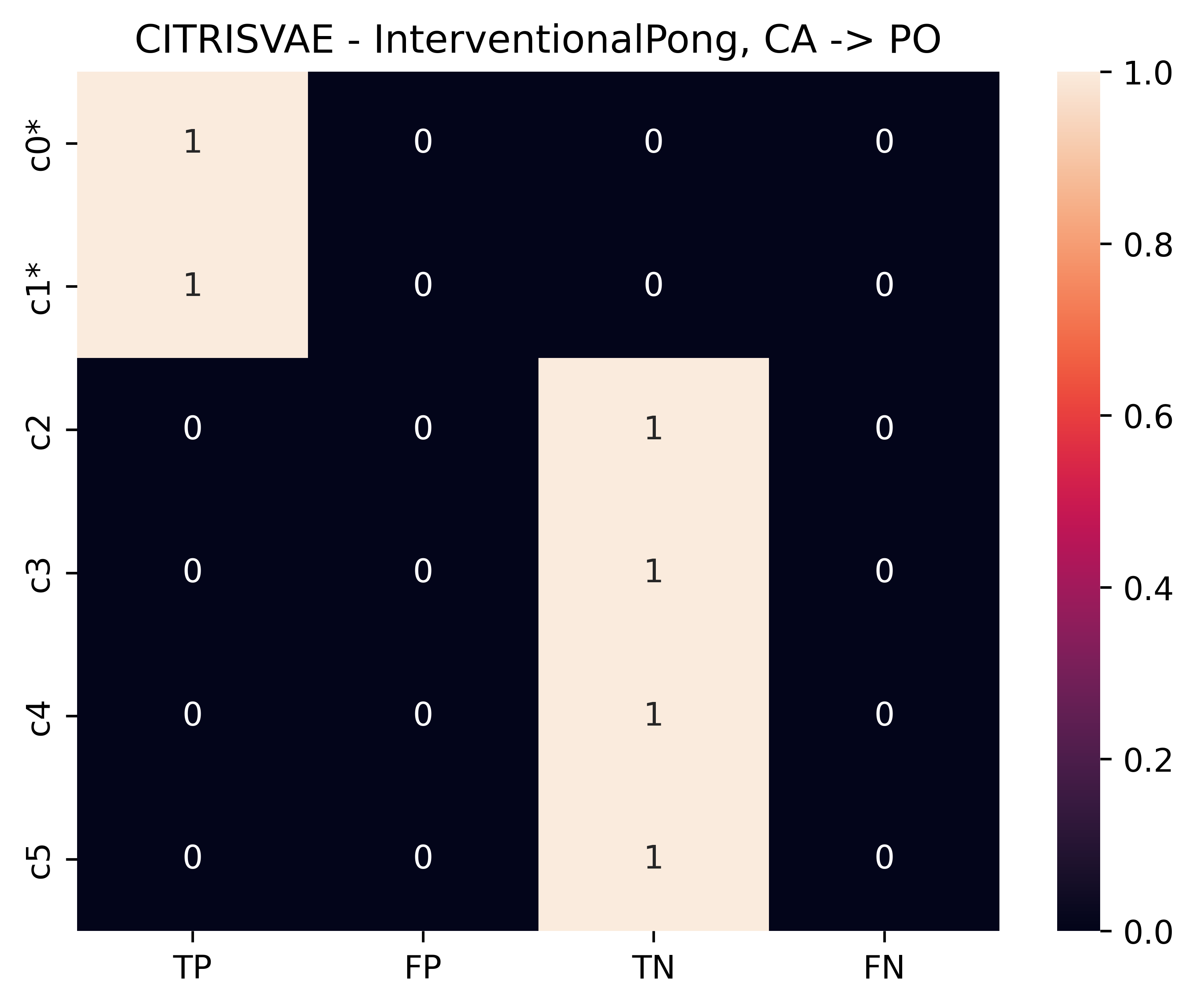}
\includegraphics[width=0.3\textwidth]{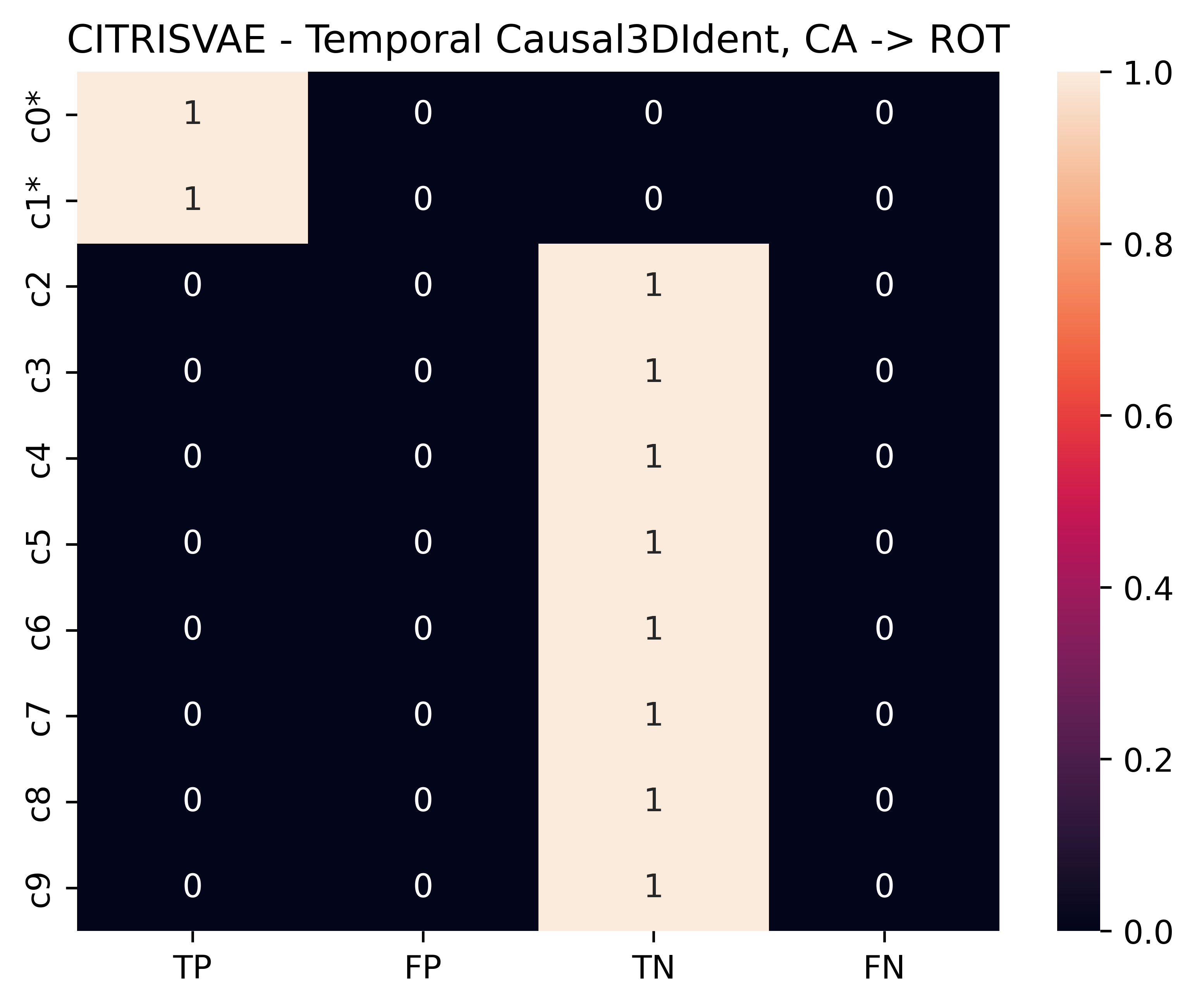}

\smallskip
\includegraphics[width=0.3\textwidth]{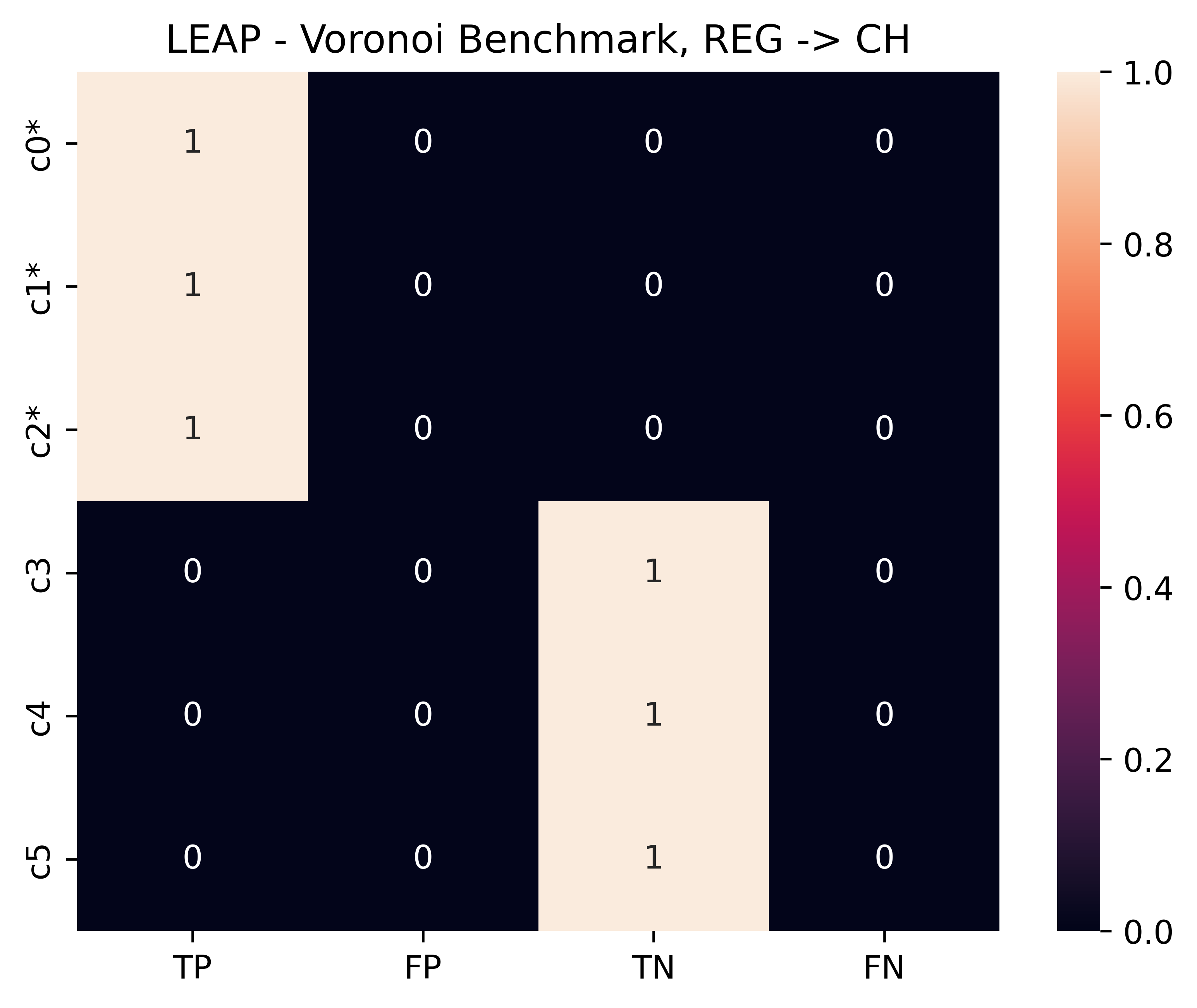}
\includegraphics[width=0.3\textwidth]{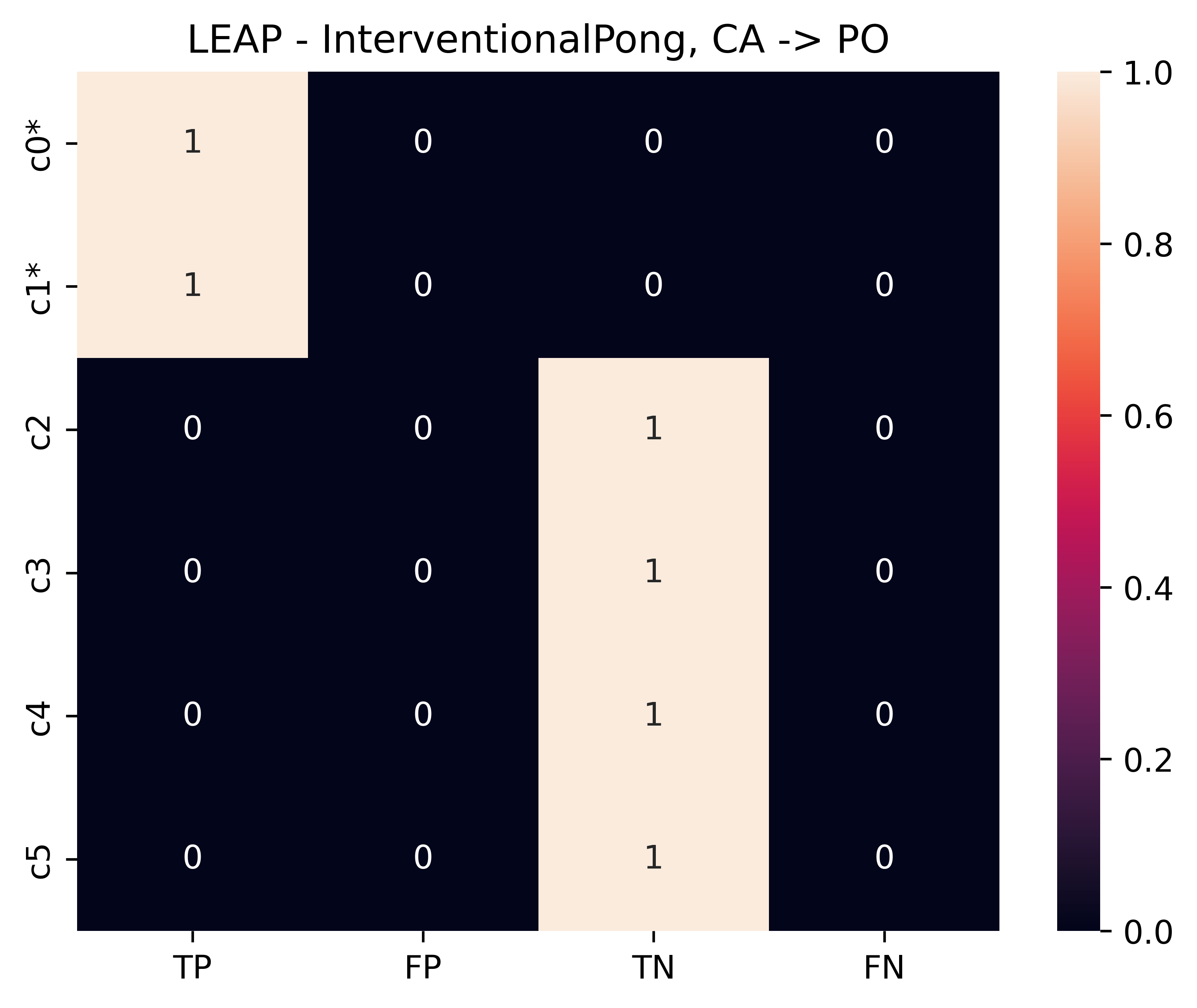}
\includegraphics[width=0.3\textwidth]{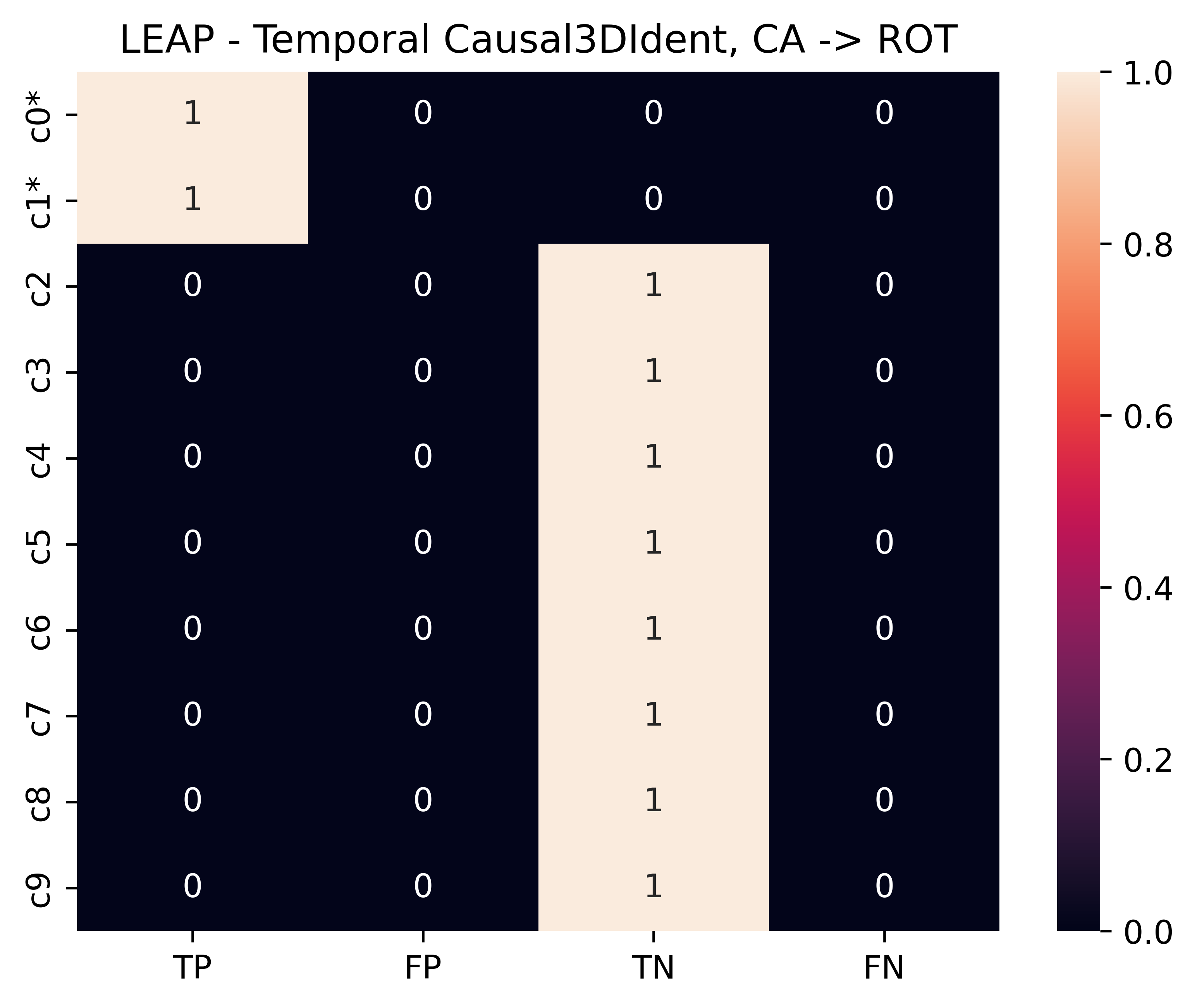}

\smallskip
\includegraphics[width=0.3\textwidth]{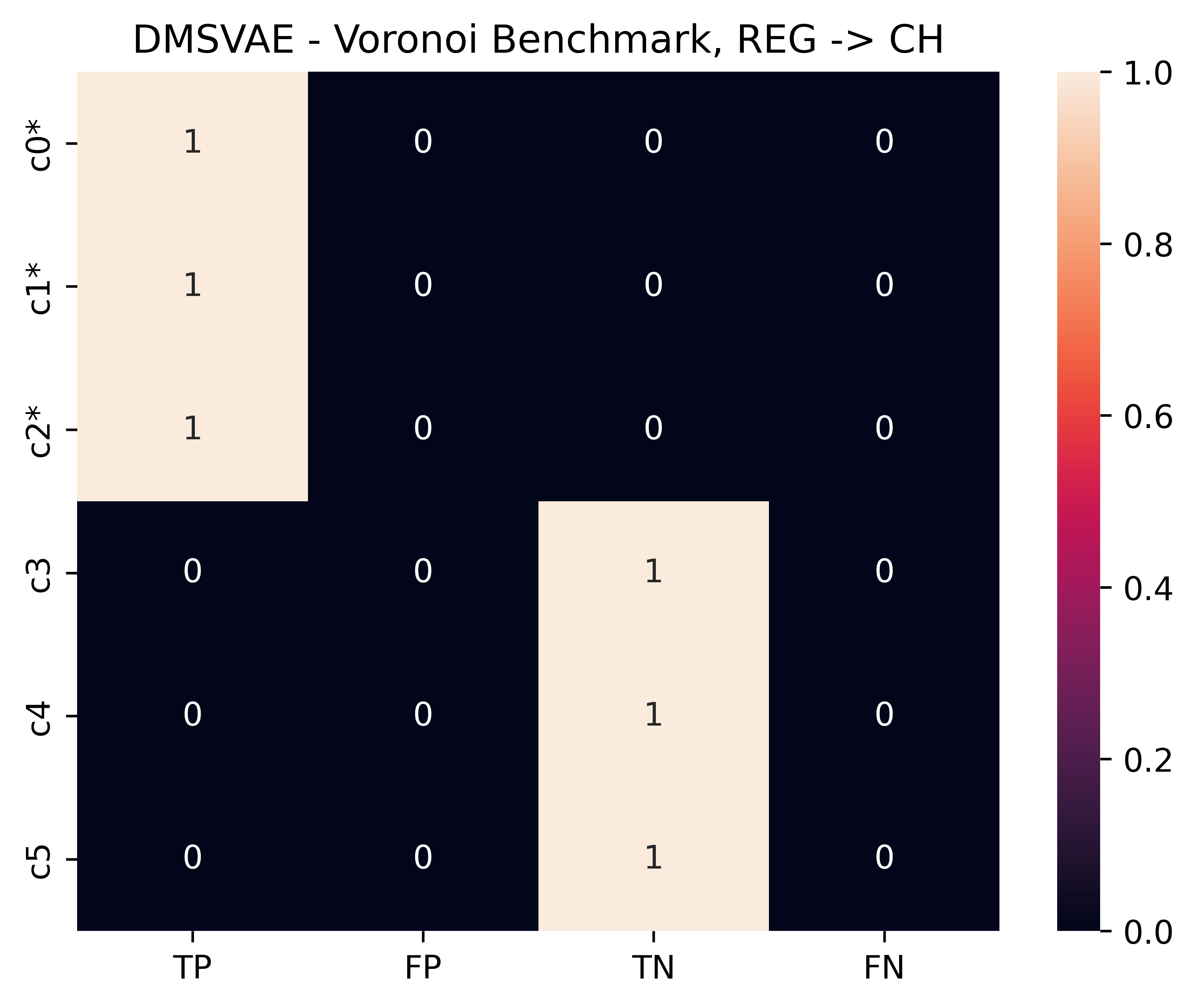}
\includegraphics[width=0.3\textwidth]{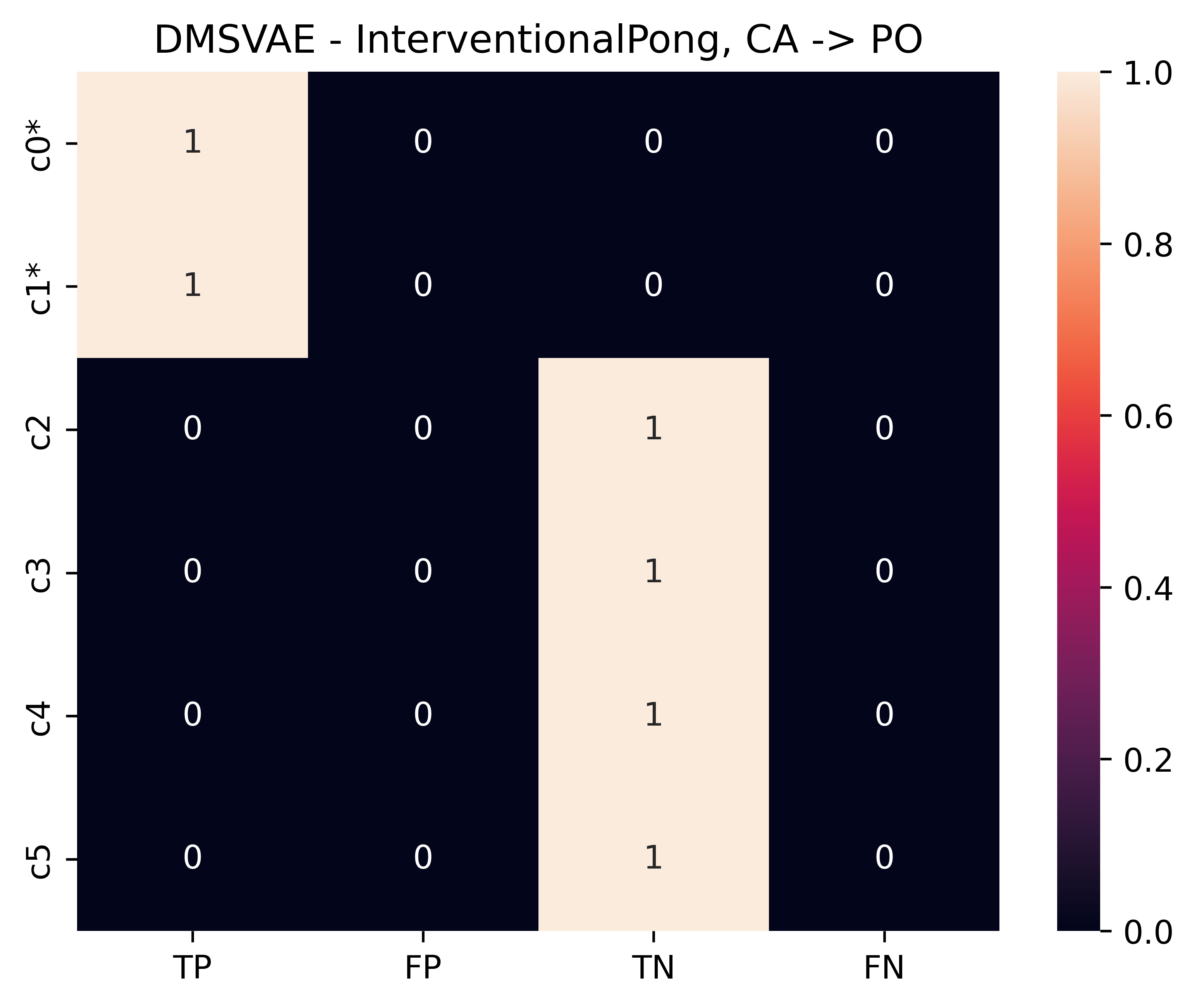}
\includegraphics[width=0.3\textwidth]{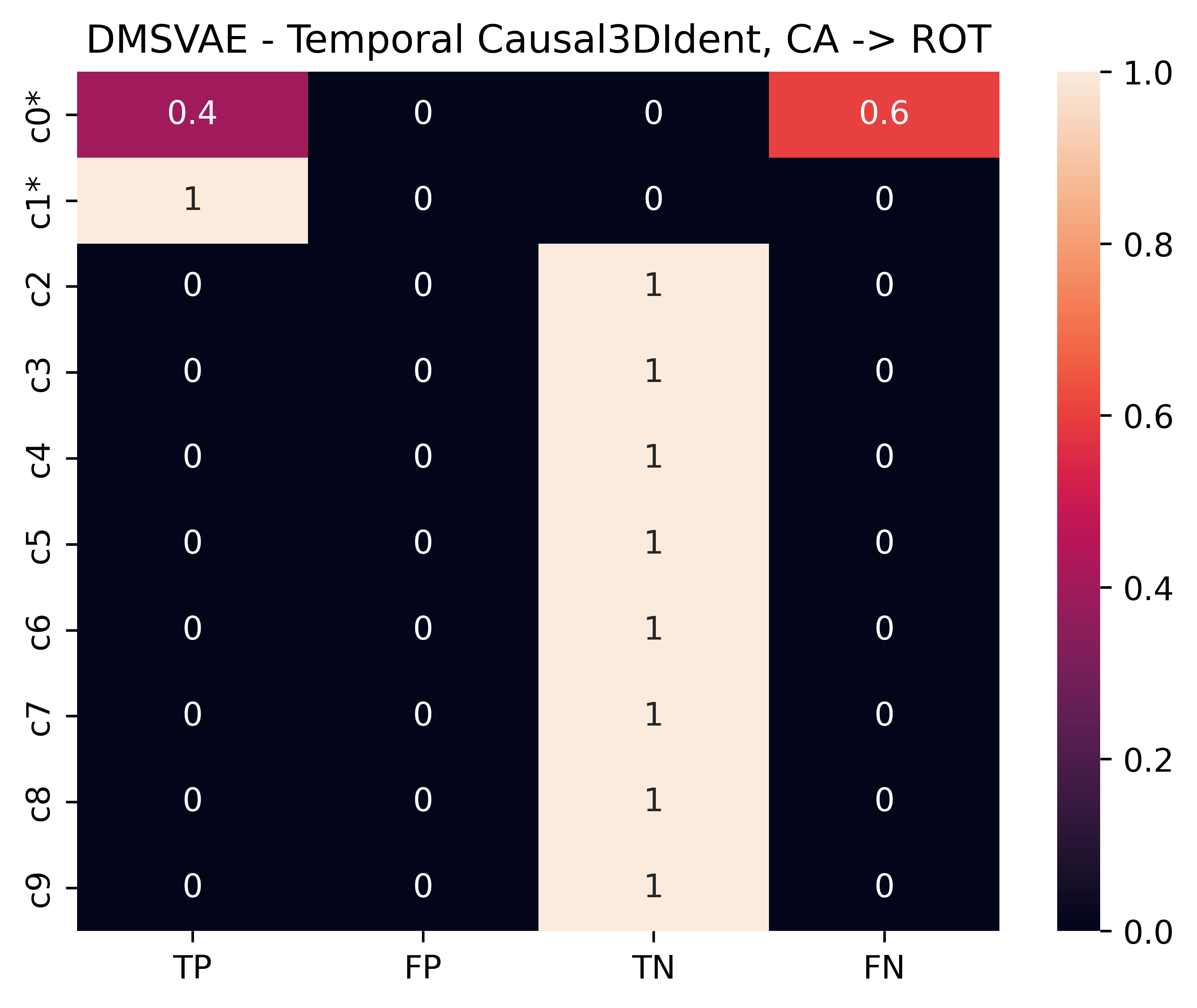}

\smallskip
\includegraphics[width=0.3\textwidth]{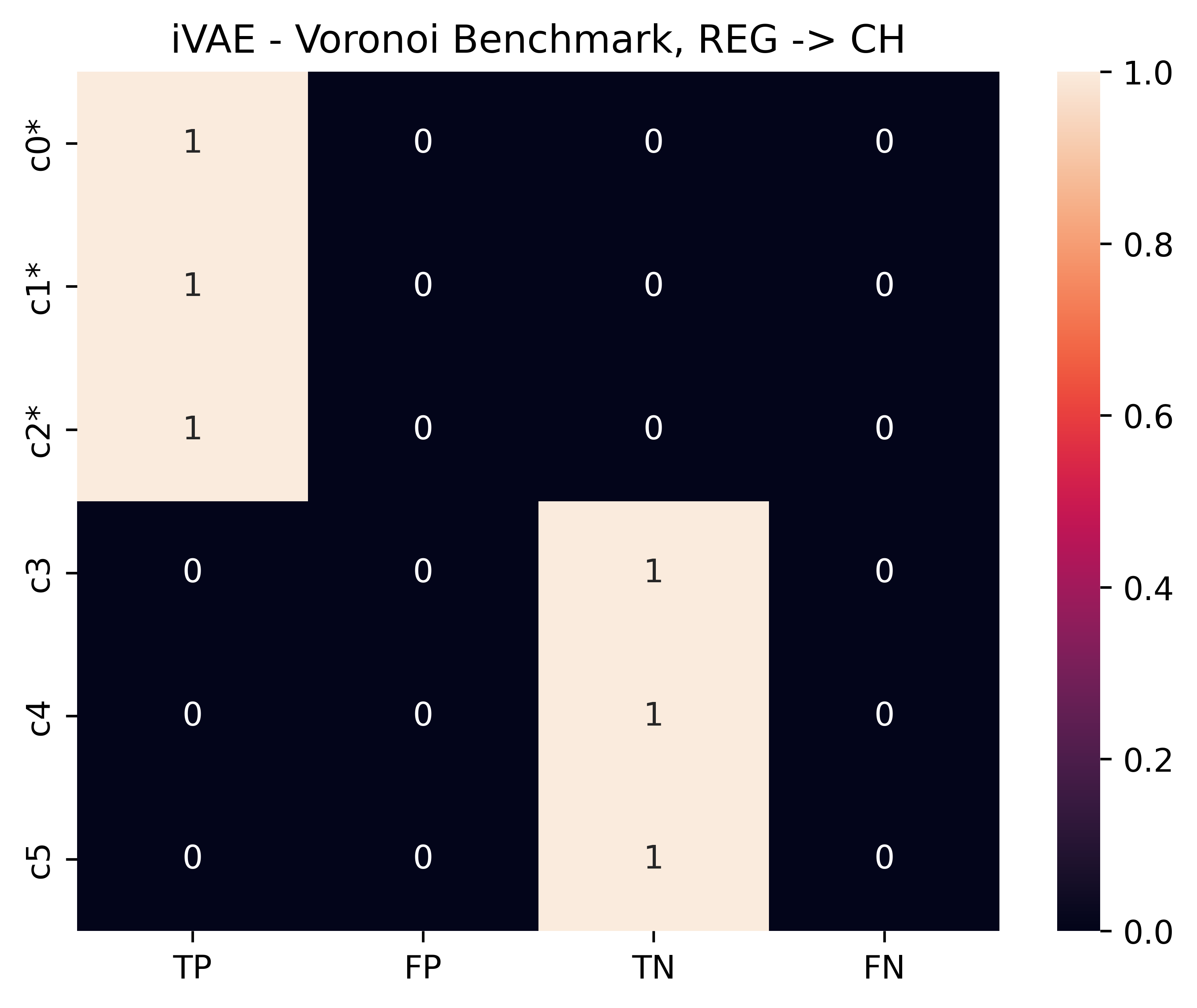}
\includegraphics[width=0.3\textwidth]{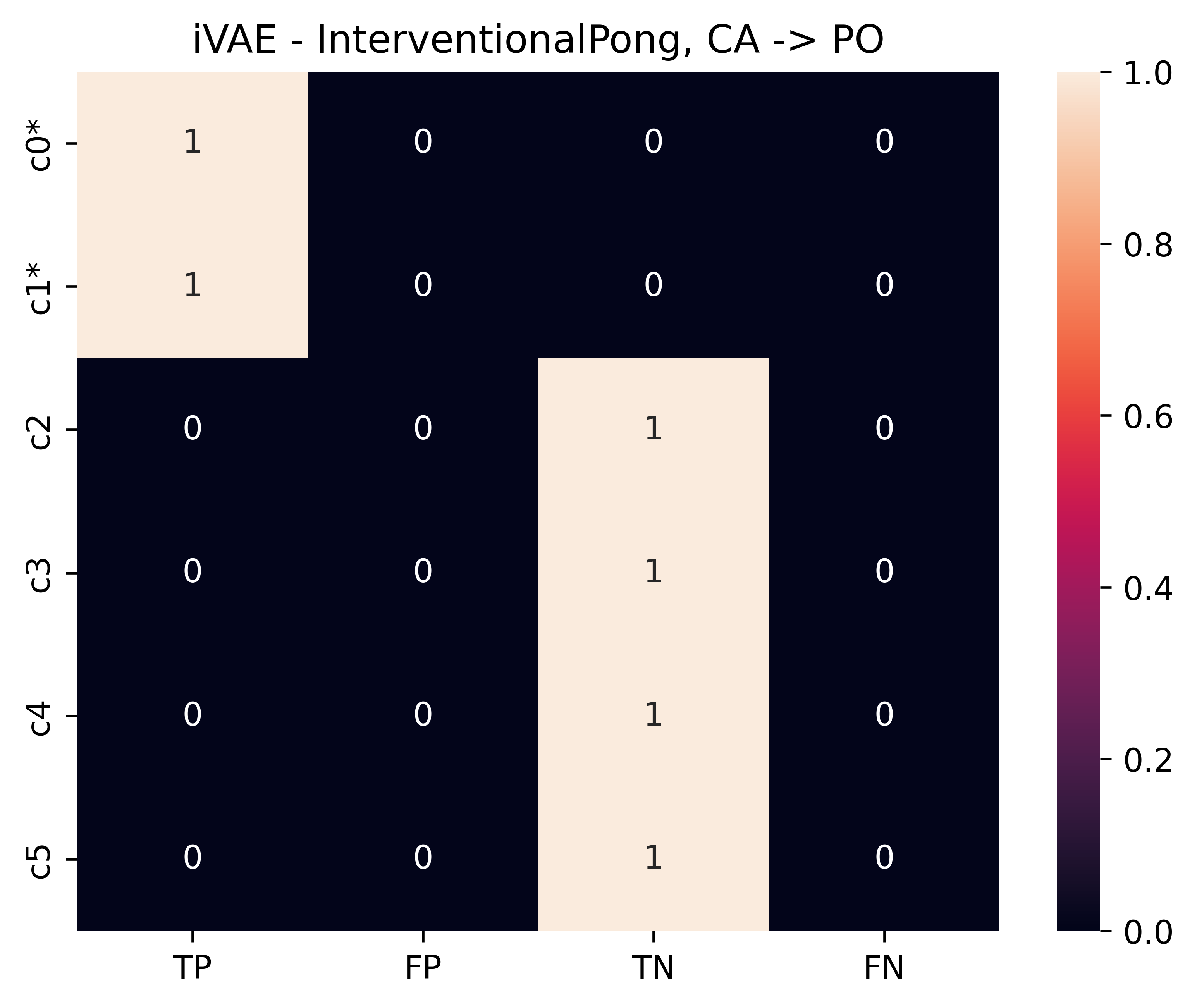}
\includegraphics[width=0.3\textwidth]{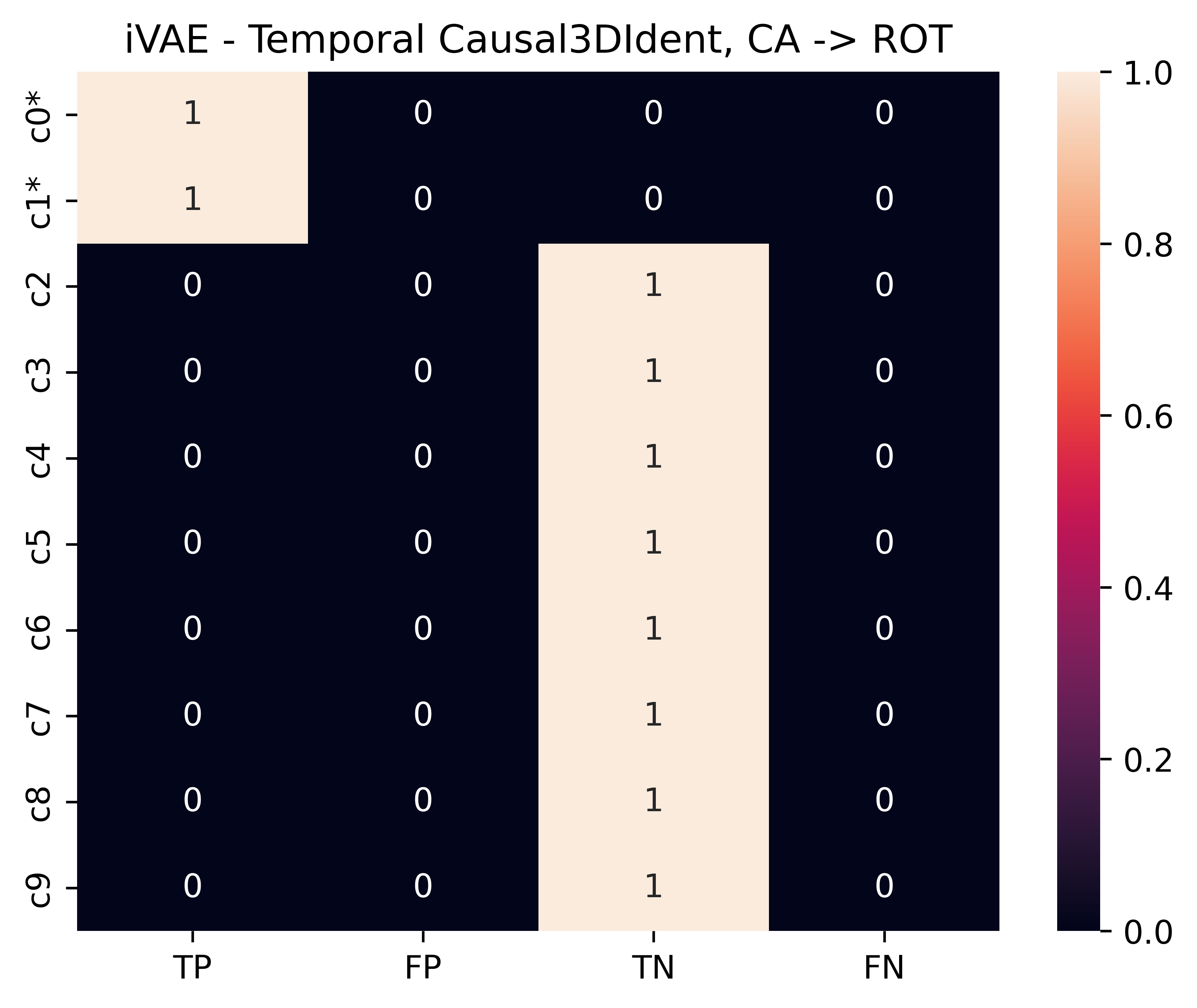}

\caption{Confusion matrix of the chagend factors detection. \textbf{Rows}: CITRISVAE, LEAP, DMSVAE and iVAE. \textbf{Columns:} Voronoi Benchmark, InterventionalPong, Temporal Causal3dIdent.}
\label{fig:app-results-detection-method}
\end{figure}

In Figure~\ref{fig:app-results-detection-method} we report the confusion matrix of the changed factor detection grouping by dataset and method.

\begin{table*}[t!]
    \centering
    \caption{Voronoi Benchmark, REG $\rightarrow$ CH (750 samples)}
    \label{tab:app-voronoi-re-reg2ch}
    \resizebox{0.8\textwidth}{!}{
    \begin{tabular}{llcccc}
        \toprule
        Approach & Adaptation & $R^2$ diag $\uparrow$ & $R^2$ off-diag $\downarrow$ & Spearman diag $\uparrow$ & Spearman off-diag $\downarrow$\\ 
        \midrule
CITRISVAE & 0shot & 0.24 \scriptsize{$\pm$ 0.09} & 0.56 \scriptsize{$\pm$ 0.05} & 0.40 \scriptsize{$\pm$ 0.10} & 0.71 \scriptsize{$\pm$ 0.03}\\ 
& ft & 0.37 \scriptsize{$\pm$ 0.14} & 0.35 \scriptsize{$\pm$ 0.07} & 0.57 \scriptsize{$\pm$ 0.13} & 0.56 \scriptsize{$\pm$ 0.06}\\ 
& DECAF (Ours) & \textbf{0.72} \scriptsize{$\pm$ 0.24} & \textbf{0.18} \scriptsize{$\pm$ 0.22} & \textbf{0.83} \scriptsize{$\pm$ 0.18} & \textbf{0.34} \scriptsize{$\pm$ 0.23}\\ 
\midrule 
LEAP & 0shot & 0.67 \scriptsize{$\pm$ 0.15} & \textbf{0.23} \scriptsize{$\pm$ 0.10} & 0.78 \scriptsize{$\pm$ 0.13} &\textbf{ 0.43} \scriptsize{$\pm$ 0.12}\\ 
& ft & 0.60 \scriptsize{$\pm$ 0.08} & 0.29 \scriptsize{$\pm$ 0.07} & 0.74 \scriptsize{$\pm$ 0.09} & 0.50 \scriptsize{$\pm$ 0.07}\\ 
& DECAF (Ours) & \textbf{0.79} \scriptsize{$\pm$ 0.09} & 0.25 \scriptsize{$\pm$ 0.12} & \textbf{0.88} \scriptsize{$\pm$ 0.06} & 0.44 \scriptsize{$\pm$ 0.09}\\ 
\midrule 
DMSVAE & 0shot & \textbf{0.67} \scriptsize{$\pm$ 0.16} & \textbf{0.23} \scriptsize{$\pm$ 0.10} & \textbf{0.78} \scriptsize{$\pm$ 0.13} & \textbf{0.42} \scriptsize{$\pm$ 0.12}\\ 
& ft & 0.63 \scriptsize{$\pm$ 0.10} & 0.33 \scriptsize{$\pm$ 0.10} & \textbf{0.78} \scriptsize{$\pm$ 0.06} & 0.55 \scriptsize{$\pm$ 0.09}\\ 
& DECAF (Ours) & 0.54 \scriptsize{$\pm$ 0.18} & 0.26 \scriptsize{$\pm$ 0.14} & 0.70 \scriptsize{$\pm$ 0.14} & 0.44 \scriptsize{$\pm$ 0.16}\\ 
\midrule 
iVAE & 0shot & 0.67 \scriptsize{$\pm$ 0.16} & 0.23 \scriptsize{$\pm$ 0.10} & 0.78 \scriptsize{$\pm$ 0.14} & 0.43 \scriptsize{$\pm$ 0.12}\\ 
& ft & 0.59 \scriptsize{$\pm$ 0.08} & 0.29 \scriptsize{$\pm$ 0.06} & 0.73 \scriptsize{$\pm$ 0.09} & 0.50 \scriptsize{$\pm$ 0.07}\\ 
& DECAF (Ours) & \textbf{0.70} \scriptsize{$\pm$ 0.16} & \textbf{0.13} \scriptsize{$\pm$ 0.11} & \textbf{0.82} \scriptsize{$\pm$ 0.11} & \textbf{0.29} \scriptsize{$\pm$ 0.13}\\ 
        \bottomrule
    \end{tabular}
    }
\end{table*}
\begin{table*}[t!]
    \centering
    \caption{InterventionalPong, CA$\rightarrow$PO (5K samples)}
    \label{tab:app-pong-re-ca2po}
    \resizebox{0.8\textwidth}{!}{
    \begin{tabular}{llcccc}
        \toprule
        Approach & Adaptation & $R^2$ diag $\uparrow$ & $R^2$ off-diag $\downarrow$ & Spearman diag $\uparrow$ & Spearman off-diag $\downarrow$\\ 
        \midrule

CITRISVAE & 0shot & 0.60 \scriptsize{$\pm$ 0.01} & 0.60 \scriptsize{$\pm$ 0.01} & 0.53 \scriptsize{$\pm$ 0.01} & 0.55 \scriptsize{$\pm$ 0.01}\\ 
& ft & 0.77 \scriptsize{$\pm$ 0.01} & 0.34 \scriptsize{$\pm$ 0.02} & 0.69 \scriptsize{$\pm$ 0.01} & 0.33 \scriptsize{$\pm$ 0.02}\\ 
& DECAF (Ours) & \textbf{0.93} \scriptsize{$\pm$ 0.03} & \textbf{0.09} \scriptsize{$\pm$ 0.04} & \textbf{0.94} \scriptsize{$\pm$ 0.03} & \textbf{0.14} \scriptsize{$\pm$ 0.06}\\ 
\midrule 
LEAP & 0shot & \textbf{0.85} \scriptsize{$\pm$ 0.01} & 0.24 \scriptsize{$\pm$ 0.01} & \textbf{0.87} \scriptsize{$\pm$ 0.01} & 0.36 \scriptsize{$\pm$ 0.01}\\ 
& ft & 0.64 \scriptsize{$\pm$ 0.02} & \textbf{0.16} \scriptsize{$\pm$ 0.02} & 0.72 \scriptsize{$\pm$ 0.02} & 0.28 \scriptsize{$\pm$ 0.02}\\ 
& DECAF (Ours) & 0.84 \scriptsize{$\pm$ 0.04} & 0.18 \scriptsize{$\pm$ 0.07} & 0.86 \scriptsize{$\pm$ 0.06} & \textbf{0.26} \scriptsize{$\pm$ 0.03}\\ 
\midrule 
DMSVAE & 0shot & 0.50 \scriptsize{$\pm$ 0.01} & 0.25 \scriptsize{$\pm$ 0.01} & 0.57 \scriptsize{$\pm$ 0.00} & 0.33 \scriptsize{$\pm$ 0.01}\\ 
& ft & 0.53 \scriptsize{$\pm$ 0.04} & 0.18 \scriptsize{$\pm$ 0.03} & 0.59 \scriptsize{$\pm$ 0.04} & 0.30 \scriptsize{$\pm$ 0.01}\\ 
& DECAF (Ours) & \textbf{0.61} \scriptsize{$\pm$ 0.01} & \textbf{0.14} \scriptsize{$\pm$ 0.01} & \textbf{0.65} \scriptsize{$\pm$ 0.01} & \textbf{0.21} \scriptsize{$\pm$ 0.01}\\ 
\midrule 
iVAE & 0shot & 0.59 \scriptsize{$\pm$ 0.04} & 0.53 \scriptsize{$\pm$ 0.01} & 0.53 \scriptsize{$\pm$ 0.03} & 0.49 \scriptsize{$\pm$ 0.02}\\ 
& ft & 0.58 \scriptsize{$\pm$ 0.03} & 0.51 \scriptsize{$\pm$ 0.01} & 0.55 \scriptsize{$\pm$ 0.02} & 0.48 \scriptsize{$\pm$ 0.03}\\ 
& DECAF (Ours) & \textbf{0.71} \scriptsize{$\pm$ 0.17} & \textbf{0.20} \scriptsize{$\pm$ 0.19} & \textbf{0.77} \scriptsize{$\pm$ 0.19} & \textbf{0.27} \scriptsize{$\pm$ 0.15}\\
        \bottomrule
    \end{tabular}
    }
\end{table*}

\begin{table*}[t!]
    \centering
    \caption{Temporal Causal3dIdent, CA$\rightarrow$ROT (1K samples)}
    \label{tab:app-c3d-re-ca2rot}
    \resizebox{0.8\textwidth}{!}{
    \begin{tabular}{llcccc}
        \toprule
        Approach & Adaptation & $R^2$ diag $\uparrow$ & $R^2$ off-diag $\downarrow$ & Spearman diag $\uparrow$ & Spearman off-diag $\downarrow$\\ 
        \midrule
CITRISVAE & 0shot & 0.76 \scriptsize{$\pm$ 0.00} & 0.28 \scriptsize{$\pm$ 0.00} & 0.87 \scriptsize{$\pm$ 0.00} & 0.48 \scriptsize{$\pm$ 0.00}\\ 
& ft & \textbf{0.95} \scriptsize{$\pm$ 0.01} & \textbf{0.01} \scriptsize{$\pm$ 0.01} & \textbf{0.98} \scriptsize{$\pm$ 0.00} & \textbf{0.06} \scriptsize{$\pm$ 0.02}\\ 
& DECAF (Ours) & 0.92 \scriptsize{$\pm$ 0.04} & 0.05 \scriptsize{$\pm$ 0.03} & 0.96 \scriptsize{$\pm$ 0.02} & 0.19 \scriptsize{$\pm$ 0.06}\\ 
\midrule 
LEAP & 0shot & 0.75 \scriptsize{$\pm$ 0.00} & 0.28 \scriptsize{$\pm$ 0.00} & 0.87 \scriptsize{$\pm$ 0.00} & 0.47 \scriptsize{$\pm$ 0.00}\\ 
& ft & 0.93 \scriptsize{$\pm$ 0.00} & 0.07 \scriptsize{$\pm$ 0.00} & 0.96 \scriptsize{$\pm$ 0.00} & 0.18 \scriptsize{$\pm$ 0.00}\\ 
& DECAF (Ours) & \textbf{0.95} \scriptsize{$\pm$ 0.01} & \textbf{0.03} \scriptsize{$\pm$ 0.02} & \textbf{0.97} \scriptsize{$\pm$ 0.01} & \textbf{0.15} \scriptsize{$\pm$ 0.04}\\ 
\midrule 
DMSVAE & 0shot & 0.66 \scriptsize{$\pm$ 0.03} & 0.23 \scriptsize{$\pm$ 0.01} & 0.81 \scriptsize{$\pm$ 0.02} & 0.44 \scriptsize{$\pm$ 0.01}\\ 
& ft & \textbf{0.81} \scriptsize{$\pm$ 0.03} & \textbf{0.09} \scriptsize{$\pm$ 0.00} & \textbf{0.90} \scriptsize{$\pm$ 0.02} & \textbf{0.24} \scriptsize{$\pm$ 0.02}\\ 
& DECAF (Ours) & 0.73 \scriptsize{$\pm$ 0.08} & 0.16 \scriptsize{$\pm$ 0.10} & 0.85 \scriptsize{$\pm$ 0.05} & 0.33 \scriptsize{$\pm$ 0.15}\\ 
\midrule 
iVAE & 0shot & 0.75 \scriptsize{$\pm$ 0.00} & 0.28 \scriptsize{$\pm$ 0.00} & 0.87 \scriptsize{$\pm$ 0.00} & 0.47 \scriptsize{$\pm$ 0.00}\\ 
& ft & 0.87 \scriptsize{$\pm$ 0.00} & 0.15 \scriptsize{$\pm$ 0.01} & 0.93 \scriptsize{$\pm$ 0.00} & 0.31 \scriptsize{$\pm$ 0.01}\\ 
& DECAF (Ours) & \textbf{0.95} \scriptsize{$\pm$ 0.02} & \textbf{0.03} \scriptsize{$\pm$ 0.01} & \textbf{0.97} \scriptsize{$\pm$ 0.01} & \textbf{0.14} \scriptsize{$\pm$ 0.04}\\

        \bottomrule
    \end{tabular}
    }
\end{table*}

\begin{table*}[t!]
    \centering
    \caption{Voronoi Benchmark, REG-j+CH-i$\rightarrow$REG-i (750 samples) with sources \texttt{REG-j} and \texttt{CH-i}.}
    \label{tab:app-voronoi-co-reg}
    \resizebox{0.8\textwidth}{!}{
    \begin{tabular}{llcccc}
        \toprule
        Approach & Adaptation & $R^2$ diag $\uparrow$ & $R^2$ off-diag $\downarrow$ & Spearman diag $\uparrow$ & Spearman off-diag $\downarrow$\\ 
        \midrule
CITRISVAE & 0shot-1 & \textbf{0.99} \scriptsize{$\pm$ 0.00} & 0.08 \scriptsize{$\pm$ 0.00} & \textbf{1.00} \scriptsize{$\pm$ 0.00} & 0.16 \scriptsize{$\pm$ 0.00}\\ 
& ft-1 & 0.95 \scriptsize{$\pm$ 0.01} & 0.06 \scriptsize{$\pm$ 0.01} & 0.98 \scriptsize{$\pm$ 0.00} & 0.20 \scriptsize{$\pm$ 0.03}\\ 
& 0shot-2 & 0.60 \scriptsize{$\pm$ 0.04} & 0.29 \scriptsize{$\pm$ 0.05} & 0.69 \scriptsize{$\pm$ 0.05} & 0.39 \scriptsize{$\pm$ 0.04}\\ 
& ft-2 & 0.79 \scriptsize{$\pm$ 0.13} & 0.14 \scriptsize{$\pm$ 0.10} & 0.87 \scriptsize{$\pm$ 0.11} & 0.27 \scriptsize{$\pm$ 0.11}\\ 
& DECAF (Ours) & \textbf{0.99} \scriptsize{$\pm$ 0.00} & \textbf{0.00} \scriptsize{$\pm$ 0.01} & \textbf{1.00} \scriptsize{$\pm$ 0.00} & \textbf{0.03} \scriptsize{$\pm$ 0.01}\\ 
\midrule 
LEAP & 0shot-1 & 0.91 \scriptsize{$\pm$ 0.00} & \textbf{0.08} \scriptsize{$\pm$ 0.00} & 0.95 \scriptsize{$\pm$ 0.00} & 0.17 \scriptsize{$\pm$ 0.00}\\ 
& ft-1 & \textbf{0.93} \scriptsize{$\pm$ 0.00} & 0.06 \scriptsize{$\pm$ 0.00} & \textbf{0.96} \scriptsize{$\pm$ 0.00} & \textbf{0.16} \scriptsize{$\pm$ 0.00}\\ 
& 0shot-2 & 0.80 \scriptsize{$\pm$ 0.13} & 0.15 \scriptsize{$\pm$ 0.08} & 0.88 \scriptsize{$\pm$ 0.08} & 0.30 \scriptsize{$\pm$ 0.10}\\ 
& ft-2 & 0.84 \scriptsize{$\pm$ 0.13} & 0.11 \scriptsize{$\pm$ 0.09} & 0.88 \scriptsize{$\pm$ 0.11} & 0.22 \scriptsize{$\pm$ 0.11}\\ 
& DECAF (Ours) & 0.90 \scriptsize{$\pm$ 0.08} & \textbf{0.08} \scriptsize{$\pm$ 0.10} & 0.92 \scriptsize{$\pm$ 0.09} & \textbf{0.16} \scriptsize{$\pm$ 0.11}\\ 
\midrule 
DMSVAE & 0shot-1 & 0.97 \scriptsize{$\pm$ 0.00} & \textbf{0.02} \scriptsize{$\pm$ 0.00} & \textbf{0.99} \scriptsize{$\pm$ 0.00} & 0.12 \scriptsize{$\pm$ 0.00}\\ 
& ft-1 & 0.92 \scriptsize{$\pm$ 0.01} & 0.08 \scriptsize{$\pm$ 0.01} & 0.96 \scriptsize{$\pm$ 0.00} & 0.20 \scriptsize{$\pm$ 0.02}\\ 
& 0shot-2 & 0.77 \scriptsize{$\pm$ 0.05} & 0.17 \scriptsize{$\pm$ 0.03} & 0.85 \scriptsize{$\pm$ 0.05} & 0.33 \scriptsize{$\pm$ 0.03}\\ 
& ft-2 & 0.78 \scriptsize{$\pm$ 0.05} & 0.15 \scriptsize{$\pm$ 0.03} & 0.87 \scriptsize{$\pm$ 0.03} & 0.32 \scriptsize{$\pm$ 0.03}\\ 
& DECAF (Ours) & \textbf{0.98} \scriptsize{$\pm$ 0.00} & \textbf{0.02} \scriptsize{$\pm$ 0.00} & \textbf{0.99} \scriptsize{$\pm$ 0.00} & \textbf{0.11} \scriptsize{$\pm$ 0.01}\\ 
\midrule 
iVAE & 0shot-1 & \textbf{0.79} \scriptsize{$\pm$ 0.00} & 0.17 \scriptsize{$\pm$ 0.00} & \textbf{0.87} \scriptsize{$\pm$ 0.00} & 0.29 \scriptsize{$\pm$ 0.00}\\ 
& ft-1 & \textbf{0.79} \scriptsize{$\pm$ 0.00} & 0.15 \scriptsize{$\pm$ 0.00} & \textbf{0.87} \scriptsize{$\pm$ 0.00} & 0.30 \scriptsize{$\pm$ 0.01}\\ 
& 0shot-2 & 0.74 \scriptsize{$\pm$ 0.09} & 0.17 \scriptsize{$\pm$ 0.04} & 0.84 \scriptsize{$\pm$ 0.06} & 0.32 \scriptsize{$\pm$ 0.05}\\ 
& ft-2 & 0.75 \scriptsize{$\pm$ 0.09} & 0.16 \scriptsize{$\pm$ 0.05} & 0.83 \scriptsize{$\pm$ 0.07} & 0.30 \scriptsize{$\pm$ 0.05}\\ 
& DECAF (Ours) & 0.73 \scriptsize{$\pm$ 0.08} & \textbf{0.04} \scriptsize{$\pm$ 0.05} & 0.76 \scriptsize{$\pm$ 0.10} & \textbf{0.08} \scriptsize{$\pm$ 0.08}\\ 
\midrule

        \bottomrule
    \end{tabular}
    }
\end{table*}

\begin{table*}[t!]
    \centering
    \caption{InterventionalPong, CA-jPA+PO-PA$\rightarrow$CA-PA (5K samples) with sources \texttt{CA-jPA} and \texttt{PO-PA}.}
    \label{tab:app-pong-co-ca}
    \resizebox{0.8\textwidth}{!}{
    \begin{tabular}{llcccc}
        \toprule
        Approach & Adaptation & $R^2$ diag $\uparrow$ & $R^2$ off-diag $\downarrow$ & Spearman diag $\uparrow$ & Spearman off-diag $\downarrow$\\ 
        \midrule
        CITRISVAE & 0shot-1 & 0.80 \scriptsize{$\pm$ 0.01} & 0.22 \scriptsize{$\pm$ 0.00} & 0.88 \scriptsize{$\pm$ 0.00} & 0.31 \scriptsize{$\pm$ 0.00}\\ 
& ft-1 & 0.92 \scriptsize{$\pm$ 0.01} & 0.03 \scriptsize{$\pm$ 0.01} & 0.95 \scriptsize{$\pm$ 0.01} & 0.13 \scriptsize{$\pm$ 0.03}\\ 
& 0shot-2 & 0.81 \scriptsize{$\pm$ 0.00} & 0.17 \scriptsize{$\pm$ 0.00} & 0.82 \scriptsize{$\pm$ 0.00} & 0.20 \scriptsize{$\pm$ 0.00}\\ 
& ft-2 & 0.77 \scriptsize{$\pm$ 0.01} & 0.10 \scriptsize{$\pm$ 0.02} & 0.79 \scriptsize{$\pm$ 0.02} & 0.18 \scriptsize{$\pm$ 0.02}\\ 
& DECAF (Ours) & \textbf{0.98} \scriptsize{$\pm$ 0.00} & \textbf{0.00} \scriptsize{$\pm$ 0.00} & \textbf{0.99} \scriptsize{$\pm$ 0.00} & \textbf{0.03} \scriptsize{$\pm$ 0.00}\\ 
\midrule 
LEAP & 0shot-1 & 0.60 \scriptsize{$\pm$ 0.01} & 0.42 \scriptsize{$\pm$ 0.00} & 0.63 \scriptsize{$\pm$ 0.00} & 0.49 \scriptsize{$\pm$ 0.00}\\ 
& ft-1 & 0.60 \scriptsize{$\pm$ 0.01} & 0.40 \scriptsize{$\pm$ 0.00} & 0.63 \scriptsize{$\pm$ 0.00} & 0.47 \scriptsize{$\pm$ 0.01}\\ 
& 0shot-2 & 0.98 \scriptsize{$\pm$ 0.00} & 0.02 \scriptsize{$\pm$ 0.00} & 0.99 \scriptsize{$\pm$ 0.00} & 0.08 \scriptsize{$\pm$ 0.00}\\ 
& ft-2 & \textbf{1.00} \scriptsize{$\pm$ 0.00} & \textbf{0.00} \scriptsize{$\pm$ 0.00} & \textbf{1.00} \scriptsize{$\pm$ 0.00} & \textbf{0.04} \scriptsize{$\pm$ 0.01}\\ 
& DECAF (Ours) & 0.76 \scriptsize{$\pm$ 0.00} & 0.30 \scriptsize{$\pm$ 0.00} & 0.78 \scriptsize{$\pm$ 0.00} & 0.34 \scriptsize{$\pm$ 0.00}\\ 
\midrule 
DMSVAE & 0shot-1 & 0.76 \scriptsize{$\pm$ 0.00} & 0.21 \scriptsize{$\pm$ 0.00} & 0.83 \scriptsize{$\pm$ 0.00} & 0.34 \scriptsize{$\pm$ 0.00}\\ 
& ft-1 & 0.79 \scriptsize{$\pm$ 0.01} & 0.19 \scriptsize{$\pm$ 0.01} & 0.83 \scriptsize{$\pm$ 0.01} & 0.33 \scriptsize{$\pm$ 0.02}\\ 
& 0shot-2 & 0.78 \scriptsize{$\pm$ 0.00} & 0.13 \scriptsize{$\pm$ 0.00} & 0.85 \scriptsize{$\pm$ 0.00} & 0.28 \scriptsize{$\pm$ 0.00}\\ 
& ft-2 & 0.76 \scriptsize{$\pm$ 0.02} & \textbf{0.10} \scriptsize{$\pm$ 0.03} & 0.80 \scriptsize{$\pm$ 0.02} & \textbf{0.26} \scriptsize{$\pm$ 0.03}\\ 
& DECAF (Ours) & \textbf{0.80} \scriptsize{$\pm$ 0.00} & 0.13 \scriptsize{$\pm$ 0.01} & \textbf{0.86} \scriptsize{$\pm$ 0.00} & 0.27 \scriptsize{$\pm$ 0.01}\\ 
\midrule 
iVAE & 0shot-1 & 0.99 \scriptsize{$\pm$ 0.00} & 0.01 \scriptsize{$\pm$ 0.00} & 1.00 \scriptsize{$\pm$ 0.00} & 0.06 \scriptsize{$\pm$ 0.00}\\ 
& ft-1 & 0.98 \scriptsize{$\pm$ 0.00} & 0.01 \scriptsize{$\pm$ 0.01} & 0.99 \scriptsize{$\pm$ 0.00} & 0.08 \scriptsize{$\pm$ 0.02}\\ 
& 0shot-2 & 0.96 \scriptsize{$\pm$ 0.00} & 0.00 \scriptsize{$\pm$ 0.00} & 0.97 \scriptsize{$\pm$ 0.00} & 0.05 \scriptsize{$\pm$ 0.00}\\ 
& ft-2 & 0.98 \scriptsize{$\pm$ 0.00} & 0.02 \scriptsize{$\pm$ 0.01} & 0.99 \scriptsize{$\pm$ 0.00} & 0.10 \scriptsize{$\pm$ 0.01}\\ 
& DECAF (Ours) & \textbf{1.00} \scriptsize{$\pm$ 0.00} & \textbf{0.00} \scriptsize{$\pm$ 0.00} & \textbf{1.00} \scriptsize{$\pm$ 0.00} & \textbf{0.02} \scriptsize{$\pm$ 0.00}\\

        \bottomrule
    \end{tabular}
    }
\end{table*}

\begin{table*}[t!]
    \centering
    \caption{Temporal Causal3DIdent, CA-jHUE+ROT-HUE$\rightarrow$CA-HUE (1K samples) with sources \texttt{CA-jHUE} and \texttt{ROT-HUE}.}
    \label{tab:app-c3d-co-ca}
    \resizebox{0.8\textwidth}{!}{
    \begin{tabular}{llcccc}
        \toprule
        Approach & Adaptation & $R^2$ diag $\uparrow$ & $R^2$ off-diag $\downarrow$ & Spearman diag $\uparrow$ & Spearman off-diag $\downarrow$\\ 
        \midrule
CITRISVAE  
& 0shot-1 & 0.74 \scriptsize{$\pm$ 0.06} & 0.25 \scriptsize{$\pm$ 0.03} & 0.72 \scriptsize{$\pm$ 0.05} & 0.28 \scriptsize{$\pm$ 0.04}\\ 
& ft-1 & 0.75 \scriptsize{$\pm$ 0.03} & 0.21 \scriptsize{$\pm$ 0.02} & 0.71 \scriptsize{$\pm$ 0.03} & 0.26 \scriptsize{$\pm$ 0.02}\\
& 0shot-2 & 0.83 \scriptsize{$\pm$ 0.01} & 0.09 \scriptsize{$\pm$ 0.00} & 0.84 \scriptsize{$\pm$ 0.02} & 0.21 \scriptsize{$\pm$ 0.01}\\ 
& ft-2 & 0.80 \scriptsize{$\pm$ 0.02} & 0.10 \scriptsize{$\pm$ 0.01} & 0.78 \scriptsize{$\pm$ 0.02} & 0.19 \scriptsize{$\pm$ 0.02}\\
& DECAF (Ours) & \textbf{0.88} \scriptsize{$\pm$ 0.01} & \textbf{0.04} \scriptsize{$\pm$ 0.01} & \textbf{0.88} \scriptsize{$\pm$ 0.01} & \textbf{0.12} \scriptsize{$\pm$ 0.01}\\ 
\midrule 
LEAP  & 0shot-1 & 0.75 \scriptsize{$\pm$ 0.02} & 0.17 \scriptsize{$\pm$ 0.02} & 0.74 \scriptsize{$\pm$ 0.02} & 0.23 \scriptsize{$\pm$ 0.01}\\ 
& ft-1 & 0.72 \scriptsize{$\pm$ 0.01} & 0.15 \scriptsize{$\pm$ 0.01} & 0.71 \scriptsize{$\pm$ 0.02} & 0.22 \scriptsize{$\pm$ 0.01}\\
& 0shot-2 & 0.76 \scriptsize{$\pm$ 0.00} & 0.15 \scriptsize{$\pm$ 0.01} & \textbf{0.78} \scriptsize{$\pm$ 0.00} & 0.27 \scriptsize{$\pm$ 0.01}\\ 
& ft-2 & 0.74 \scriptsize{$\pm$ 0.01} & \textbf{0.12} \scriptsize{$\pm$ 0.01} & 0.75 \scriptsize{$\pm$ 0.01} & 0.24 \scriptsize{$\pm$ 0.01}\\
& DECAF (Ours) & \textbf{0.78} \scriptsize{$\pm$ 0.00} & \textbf{0.12} \scriptsize{$\pm$ 0.01} & \textbf{0.78} \scriptsize{$\pm$ 0.00} & \textbf{0.21} \scriptsize{$\pm$ 0.01}\\ 
\midrule 
DMSVAE  
& 0shot-1 & 0.64 \scriptsize{$\pm$ 0.02} & 0.25 \scriptsize{$\pm$ 0.02} & 0.62 \scriptsize{$\pm$ 0.02} & 0.29 \scriptsize{$\pm$ 0.02}\\ 
& ft-1 & 0.61 \scriptsize{$\pm$ 0.02} & \textbf{0.19} \scriptsize{$\pm$ 0.01} & 0.60 \scriptsize{$\pm$ 0.01} & \textbf{0.27} \scriptsize{$\pm$ 0.01}\\
& 0shot-2 &\textbf{ 0.67} \scriptsize{$\pm$ 0.03} & 0.21 \scriptsize{$\pm$ 0.04} & \textbf{0.66} \scriptsize{$\pm$ 0.04} & 0.28 \scriptsize{$\pm$ 0.01}\\ 
& ft-2 & 0.63 \scriptsize{$\pm$ 0.03} & \textbf{0.19} \scriptsize{$\pm$ 0.04} & 0.62 \scriptsize{$\pm$ 0.04} & \textbf{0.27} \scriptsize{$\pm$ 0.01}\\
& DECAF (Ours) & \textbf{0.67} \scriptsize{$\pm$ 0.03} & 0.21 \scriptsize{$\pm$ 0.02} & \textbf{0.66} \scriptsize{$\pm$ 0.04} & 0.28 \scriptsize{$\pm$ 0.02}\\ 
\midrule 
iVAE  
& 0shot-1 & 0.68 \scriptsize{$\pm$ 0.02} & 0.24 \scriptsize{$\pm$ 0.01} & 0.64 \scriptsize{$\pm$ 0.02} & 0.24 \scriptsize{$\pm$ 0.01}\\ 
& ft-1 & 0.64 \scriptsize{$\pm$ 0.01} & 0.22 \scriptsize{$\pm$ 0.00} & 0.62 \scriptsize{$\pm$ 0.01} & 0.26 \scriptsize{$\pm$ 0.00}\\
& 0shot-2 & 0.72 \scriptsize{$\pm$ 0.02} & 0.16 \scriptsize{$\pm$ 0.02} & \textbf{0.73} \scriptsize{$\pm$ 0.03} & 0.24 \scriptsize{$\pm$ 0.01}\\ 
& ft-2 & 0.71 \scriptsize{$\pm$ 0.02} & \textbf{0.14} \scriptsize{$\pm$ 0.02} & 0.71 \scriptsize{$\pm$ 0.02} & 0.26 \scriptsize{$\pm$ 0.01}\\
& DECAF (Ours) & \textbf{0.74} \scriptsize{$\pm$ 0.02} & 0.15 \scriptsize{$\pm$ 0.03} & 0.72 \scriptsize{$\pm$ 0.03} & \textbf{0.20} \scriptsize{$\pm$ 0.02}\\ 

        \bottomrule
    \end{tabular}
    }
\end{table*}

\subsection{Ablation analysis}
Table~\ref{tab:app-c3d-noD} reports additional ablation experiments when adapting without detecting changed factors. While Table~\ref{tab:app-nf} compares DECAF with CITRISNF~\cite{lippe2022citris} employing a similar adaptation strategy, Table~\ref{tab:app-c3d-angles} evaluates the performance of the model on Temporal Causal3DIdent change from Cartesian to rotated axis, when changing the degrees of rotation.
\begin{table*}[t!]
    \centering
    \caption{Spearman CC (higher best, $\uparrow$) of inferred latents to the ground truth changed variables when adapting the representations. We report \textit{mean} $\pm$\textit{std} over 3 runs. In the table, \textit{target} denotes the model trained directly on large target data (250K) and \textit{DECAF w/o detection} ablates the adaptation without detection of changed factors.}
    \label{tab:app-c3d-noD}
    \resizebox{0.9\textwidth}{!}{
    \begin{tabular}{lccccc}
    \toprule
        Approach & Target samples& CITRISVAE & LEAP & DMSVAE  & iVAE\\ 
    \midrule

    \multicolumn{6}{c}{\cellcolor{gray!25}{\textbf{Voronoi Benchmark REG $\rightarrow$ CH}}}\\
    Target & 250K & 0.96 \scriptsize{$\pm$0.02} &0.80  \scriptsize{$\pm$0.22} & 0.96  \scriptsize{$\pm$0.01} & 0.95  \scriptsize{$\pm$0.04}\\
    \hdashline
    DECAF w/o detection & 750 & 0.84  \scriptsize{$\pm$0.08} & 0.86  \scriptsize{$\pm$0.07} & 0.64  \scriptsize{$\pm$0.16} & 0.83  \scriptsize{$\pm$0.11}\\
    DECAF & 750 & 0.67  \scriptsize{$\pm$0.29} & 0.70  \scriptsize{$\pm$0.01} & 0.68  \scriptsize{$\pm$0.17} &0.75 \scriptsize{$\pm$0.16}\\
    \bottomrule
    \multicolumn{6}{c}{\cellcolor{gray!25}{\textbf{InterventionalPong CA $\rightarrow$ PO}}}\\
        Target & 250K & 0.87\scriptsize{$\pm$0.09} & 0.49 \scriptsize{$\pm$0.03} & 0.78 \scriptsize{$\pm$0.09} & 0.66\scriptsize{$\pm$0.17}\\
        \hdashline
        DECAF w/o detection & 5K & 0.44 \scriptsize{$\pm$0.04} & 0.46\scriptsize{$\pm$0.05} & 0.42\scriptsize{$\pm$0.02} & 0.35\scriptsize{$\pm$0.03}\\
        DECAF & 5K & 0.88\scriptsize{$\pm$0.06} & 0.82\scriptsize{$\pm$0.01} & 0.71\scriptsize{$\pm$0.00} & 0.69\scriptsize{$\pm$0.21}\\
        \bottomrule
        \multicolumn{6}{c}{\cellcolor{gray!25}{\textbf{Temporal Causal3DIdent CA $\rightarrow$ ROT}}}\\
        Target & 250K& 0.99  \scriptsize{$\pm$0.00} & 0.92\scriptsize{$\pm$0.00} & 0.69  \scriptsize{$\pm$0.02} & 0.91  \scriptsize{$\pm$0.00}\\
        \hdashline
        DECAF w/o detection & 1K & 0.35  \scriptsize{$\pm$0.11} & 0.37  \scriptsize{$\pm$0.17} & 0.25  \scriptsize{$\pm$0.12} & 0.41  \scriptsize{$\pm$0.23}\\
        DECAF & 1K & 0.88  \scriptsize{$\pm$0.00} & 0.92  \scriptsize{$\pm$0.00} & 0.72  \scriptsize{$\pm$0.10} & 0.92 \scriptsize{$\pm$0.03}\\
        \bottomrule
    \end{tabular}
    }
\end{table*}

\begin{table*}[t!]
    \centering
    \caption{Spearman CC (higher best, $\uparrow$) of inferred latents to the ground truth changed variables when adapting the representations. We report \textit{mean} $\pm$\textit{std} over 3 runs. We compare DECAF when applied to CITRISVAE with CITRISNF that employs a normalizing flow for identification of causal factors.}
    \label{tab:app-nf}
    \resizebox{0.9\textwidth}{!}{
    \begin{tabular}{lccc}
        \toprule
        Approach & Voronoi Benchmark & InterventionalPong & Temporal Causal3DIdent\\
        & REG $\rightarrow$ CH &CA $\rightarrow$ PO &CA $\rightarrow$ ROT \\
        \midrule
        CITRISNF  & 0.42  \scriptsize{$\pm$0.16} & 0.73  \scriptsize{$\pm$0.07} & 0.18 \scriptsize{$\pm$0.23} \\
        CITRISVAE-DECAF & 0.67  \scriptsize{$\pm$0.29} & 0.88\scriptsize{$\pm$0.06} & 0.88  \scriptsize{$\pm$0.00}\\
        \bottomrule
    \end{tabular}
    }
\end{table*}

\begin{table*}[t!]
    \centering
    \caption{Spearman CC (higher best, $\uparrow$) of inferred latents to the ground truth position variables when adapting the representations in Temporal Causal3DIdent. We report \textit{mean} $\pm$\textit{std} over 3 runs. We consider the change from Cartesian to rotated axis, when changing the degrees of rotation. Approach-* indicates the approach when adapting to the setting with the specified * degree of rotation.}
    \label{tab:app-c3d-angles}
    \resizebox{0.65\textwidth}{!}{
    \begin{tabular}{lcccc}
        \toprule
        Approach & CITRISVAE & LEAP & DMSVAE & iVAE\\
        \midrule
ft-30 & 0.96
\scriptsize{$\pm$0.0} &	0.89
\scriptsize{$\pm$0.0} & 0.82
\scriptsize{$\pm$0.02} & 0.80
\scriptsize{$\pm$0.01}\\
DECAF-30 & 0.88
\scriptsize{$\pm$0.0}& 0.92
\scriptsize{$\pm$0.0}& 0.72
\scriptsize{$\pm$0.1}& 0.92
\scriptsize{$\pm$0.03}\\
\hdashline
ft-40 & 0.89
\scriptsize{$\pm$0.0}& 0.86
\scriptsize{$\pm$0.0}& 0.82
\scriptsize{$\pm$0.02}& 0.79
\scriptsize{$\pm$0.01}\\
DECAF-40 & 0.9
\scriptsize{$\pm$0.01}&	0.91
\scriptsize{$\pm$0.02}& 0.78
\scriptsize{$\pm$0.11}& 0.89
\scriptsize{$\pm$0.02}\\
        \bottomrule
    \end{tabular}
    }
\end{table*}

\end{document}